\documentclass[runningheads]{llncs}

\usepackage{eccv}

\usepackage{eccvabbrv}

\usepackage{graphicx}
\usepackage{booktabs}

\usepackage{array}
\usepackage{multirow}
\usepackage{colortbl}
\usepackage{siunitx}
\usepackage{bigstrut}
\usepackage{float}
\usepackage{booktabs}       
\usepackage{amsfonts}       
\usepackage{nicefrac}       
\usepackage{xcolor}         
\usepackage{adjustbox}
\usepackage{multirow}
\usepackage{siunitx}
\usepackage{tikz}
\usepackage{gensymb}
\usepackage{amsmath}
\usepackage{graphicx}
\usepackage{tabularray}

\usepackage{calculator}

\usepackage{longtable}
\usepackage{adjustbox}

\usepackage{pifont}
\newcommand{\cmark}{\ding{51}}%
\newcommand{\xmark}{\ding{55}}

\usepackage{dsfont}
\usepackage{etoolbox}
\usepackage{color}

\newif\ifshowedits

\newcommand{\addeditor}[3]{%
  \definecolor{#1color}{rgb}{#3}
  \expandafter\newcommand\csname #1\endcsname[1]{%
  \ifshowedits
    {\color{#1color} ##1}%
  \else
    {##1}%
  \fi
  }%
  \expandafter\newcommand\csname #1rmk\endcsname[1]{%
  \ifshowedits
    {\color{#1color} {\bf [#2: ##1]}}
  \fi
  }%
  \expandafter\newcommand\csname #1rpl\endcsname[2]{%
  \ifshowedits
    {\color{#1color} ##1 \sout{##2}}
  \else
    {##1}
  \fi
  }%
}

\newcommand{\createtextvar}[1]{
  \expandafter\newcommand\csname #1\endcsname{%
  {\text{#1}}
}%
}
\usepackage[bigfiles]{pdfbase}
\ExplSyntaxOn
\NewDocumentCommand\embedvideo{smm}{
  \group_begin:
  \leavevmode
  \tl_if_exist:cTF{file_\file_mdfive_hash:n{#3}}{
    \tl_set_eq:Nc\video{file_\file_mdfive_hash:n{#3}}
  }{
    \IfFileExists{#3}{}{\GenericError{}{File~`#3'~not~found}{}{}}
    \pbs_pdfobj:nnn{}{fstream}{{}{#3}}
    \pbs_pdfobj:nnn{}{dict}{
      /Type/Filespec/F~(#3)/UF~(#3)
      /EF~<</F~\pbs_pdflastobj:>>
    }
    \tl_set:Nx\video{\pbs_pdflastobj:}
    \tl_gset_eq:cN{file_\file_mdfive_hash:n{#3}}\video
  }
  \pbs_pdfobj:nnn{}{dict}{
    /Type/RichMediaInstance/Subtype/Video
    /Asset~\video
    /Params~<</FlashVars (
      source=#3&
      skin=SkinOverAllNoFullNoCaption.swf&
      skinAutoHide=true&
      skinBackgroundColor=0x5F5F5F&
      skinBackgroundAlpha=0.75
    )>>
  }
  \pbs_pdfobj:nnn{}{dict}{
    /Type/RichMediaConfiguration/Subtype/Video
    /Instances~[\pbs_pdflastobj:]
  }
  \pbs_pdfobj:nnn{}{dict}{
    /Type/RichMediaContent
    /Assets~<<
      /Names~[(#3)~\video]
    >>
    /Configurations~[\pbs_pdflastobj:]
  }
  \tl_set:Nx\rmcontent{\pbs_pdflastobj:}
  \pbs_pdfobj:nnn{}{dict}{
    /Activation~<<
      /Condition/\IfBooleanTF{#1}{PV}{XA}
      /Presentation~<</Style/Embedded>>
    >>
    /Deactivation~<</Condition/PI>>
  }
  \hbox_set:Nn\l_tmpa_box{#2}
  \tl_set:Nx\l_box_wd_tl{\dim_use:N\box_wd:N\l_tmpa_box}
  \tl_set:Nx\l_box_ht_tl{\dim_use:N\box_ht:N\l_tmpa_box}
  \tl_set:Nx\l_box_dp_tl{\dim_use:N\box_dp:N\l_tmpa_box}
  \pbs_pdfxform:nnnnn{1}{1}{}{}{\l_tmpa_box}
  \pbs_pdfannot:nnnn{\l_box_wd_tl}{\l_box_ht_tl}{\l_box_dp_tl}{
    /Subtype/RichMedia
    /BS~<</W~0/S/S>>
    /Contents~(embedded~video~file:#3)
    /NM~(rma:#3)
    /AP~<</N~\pbs_pdflastxform:>>
    /RichMediaSettings~\pbs_pdflastobj:
    /RichMediaContent~\rmcontent
  }
  \phantom{#2}
  \group_end:
}
\ExplSyntaxOff

\usepackage{graphicx}

\newcommand{\moretextwithfigures}{
\renewcommand{\topfraction}{1}
\renewcommand{\dbltopfraction}{1}
\renewcommand{\bottomfraction}{1}
\renewcommand{\textfraction}{.0}
\renewcommand{\floatpagefraction}{1}
\renewcommand{\dblfloatpagefraction}{1}
}

\newcommand{\mycomment}[1]{}

\newcommand{\calL}{{\cal L}}

\newcommand{\ba}{{\bf a}}

\newcommand{\bd}{{\bf d}}

\newcommand{\bbf}{{\bf f}}  
\newcommand{\bff}{{\bf f}}  

\newcommand{\bp}{{\bf p}}

\newcommand{\br}{{\bf r}}

\newcommand{\IR}{{\mathds{R}}}

\newcommand{\vcomment}[1]{}

\addeditor{vincent}{VL}{0.0, 0.5, 0.0}
\addeditor{arslan}{AA}{0.0, 0.0, 0.8}
\addeditor{corentin}{CS}{0.5, 0.15, 0.15}
\addeditor{tom}{TR}{1.0, 0.0, 1.0}

\moretextwithfigures

\let\subparagraph\paragraph
\usepackage{titlesec}
\titlespacing*{\section}
{0pt}                
{12pt plus 4pt minus 2pt} 
{8pt plus 2pt minus 2pt}  
\titlespacing*{\subsection}
{0pt}{8pt}{4pt}
\titlespacing*{\paragraph}
{0pt}{4pt}{2pt}

\newcommand{\tsb}[1]{{$\pm$#1}}

\usepackage[accsupp]{axessibility}  

\usepackage{hyperref}

\usepackage{orcidlink}

\newcommand{\beginsupplement}{%
\setcounter{table}{0}
\renewcommand{\thetable}{S\arabic{table}}%
\setcounter{figure}{0}
\renewcommand{\thefigure}{S\arabic{figure}}%
\setcounter{section}{0}
\renewcommand{\thesection}{\Alph{section}}%
}

\begin{document}


\title{sim2art: Accurate Articulated Object Modeling\\
from a Single Casual Video using\\
Synthetic Training Data Only} 

\titlerunning{sim2art}

\author{Arslan Artykov \and
Tom Ravaud \and
Corentin Sautier \and Vincent Lepetit}

\authorrunning{A.Artykov et al.}

\institute{LIGM, École Nationale des Ponts et Chaussées, IP Paris, CNRS, France }

\maketitle

\begin{center}
  \centering
  \includegraphics[width=\linewidth]{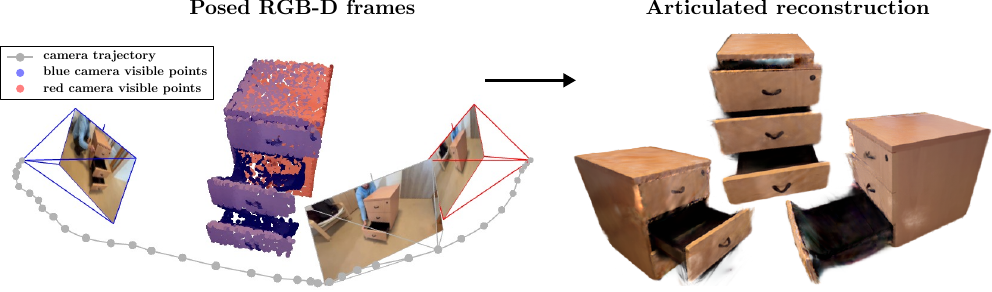}
  \captionof{figure}{ 
    We introduce a method to jointly estimate kinematic models and dynamic part labels for articulated objects captured from a casual video with  significant camera ego-motion. Such motions are challenging as the appearances of the objects vary drastically with some parts appearing and others disappearing, and our approach is significantly more accurate and robust than previous methods. It is then easy to render the object under novel views and configurations not seen in the input video.
  }
  \label{fig:opening}
\end{center}

\begin{abstract}
Understanding articulated objects from monocular video is a crucial yet challenging task in robotics and digital twin creation. Existing methods often rely on complex multi-view setups, high-fidelity object scans, or fragile long-term point tracks that frequently fail in casual real-world captures. In this paper, we present sim2art, a data-driven framework that recovers the 3D part segmentation and joint parameters of articulated objects from a single monocular video captured by a freely moving camera.
Our core insight is a robust representation based on per-frame surface point sampling, which we augment with short-term scene flow and DINOv3 semantic features. Unlike previous works that depend on error-prone long-term correspondences, our representation is easy to obtain and exhibits a negligible difference between simulation and reality without requiring domain adaptation. Also, by construction, our method relies on single-viewpoint visibility, ensuring that the geometric representation remains consistent across synthetic and real data despite noise and occlusions. Leveraging a suitable Transformer-based architecture, sim2art is trained exclusively on synthetic data yet generalizes strongly to real-world sequences.
To address the lack of standardized benchmarks in the field, we introduce two datasets featuring a significantly higher diversity of object categories and instances than prior work. Our evaluations show that sim2art effectively handles large camera motions and complex articulations, outperforming state-of-the-art optimization-based and tracking-dependent methods. sim2art offers a scalable solution that can be easily extended to new object categories without the need for cumbersome real-world annotations.
Project webpage: https://aartykov.github.io/sim2art/


\end{abstract}

\section{Introduction}
\label{sec:intro}

Articulated object understanding is a fundamental problem in robotics and digital twin creation. However, transitioning from controlled lab environments to casual, real-world captures remains an open challenge. Current state-of-the-art approaches typically fall into two categories, both with significant drawbacks: they either demand unwieldy setups, such as multi-view captures and full-object scans~\cite{liu2023paris,weng2024neural,kerr2024rsrd}, or they rely on single RGB-D frames~\cite{jiang2022opd,kawana2024detection}, which inherently lack the temporal evidence required to resolve complex 3D articulations accurately.

The emergence of monocular video-based methods~\cite{liu2023building,artykov-iccvw25-articulated} offers a more accessible path, yet these frameworks are often tethered to long-term point tracking. As we demonstrate in our experiments, despite recent progress in motion estimation, extracting reliable long-range tracks from casual video remains difficult especially in case of occlusions and significant camera movements.

In this paper, we introduce sim2art, a data-driven framework designed to recover high-fidelity articulated 3D models from casual monocular video. Unlike prior art, sim2art sidesteps the pitfalls of point correspondences, optionally augmented with short-term scene flow and DINOv3 semantic features~\cite{simeoni2025dinov3}. Our core insight is a robust representation of the input video based on per-frame surface point sampling. Crucially, our representation generalizes well from synthetic data to real data without requiring domain adaptation. This allows us to train exclusively on synthetic data, bypassing the prohibitive cost of annotating real-world articulated joints.

We will make our source code and datasets publicly available.

\begin{table*}
\centering
\caption{\textbf{Comparison of datasets and evaluation protocols.} We compare our 4art-synth and 4art-real datasets~(collectively referred as 4art here) against those used by state-of-the-art methods. Articulate-Anything does not state the numbers of instances and categories they consider.}
\label{tab:dataset_comparison}
\renewcommand{\arraystretch}{1.2}
\resizebox{\linewidth}{!}{

\begin{tabular}{l@{\;\;\;}c@{\;\;\;}c@{\;\;\;}c@{\;\;\;}c@{\;\;\;}c@{\;\;\;} c@{}}
\toprule
    & \multirow{2}{*}{Input Modality} & \#Instances/\#Categories & \# Instances/\#Categories  & Part Segmentation & \multirow{2}{*}{Camera Motion} &\multirow{2}{*}{Multi-Part Objects}\\
\textit{Method} & & synthetic data & annotated real data &real data & \\
\midrule

GAMMA \cite{yu2024gamma} & Single Point Clouds & 562/7 &2/2 & \xmark & Static &\cmark \\

Video2Articulation \cite{peng2025itaco} & Scans + Videos & 284/11 &0/0 & \xmark & Minimal &\xmark \\

Reart \cite{liu2023building} & Point Cloud Videos & 18/6 &0/0 & \xmark & Static  &\cmark \\

Articulate-Anything \cite{le2024articulate} & Monocular Videos & - &0/0 & \xmark & Static &\xmark \\

Artipoint \cite{werby2025articulated} & Monocular Videos & 0/0  & 30/10  & \xmark & Minimal &\cmark \\

\midrule
\textbf{4art (ours)} & Monocular Videos & 501/\textbf{14} &5/5 & \cmark & \textbf{Large} &\textbf{\cmark} \\
\bottomrule
\end{tabular}
}
\end{table*}








\section{Related Work}
\label{sec:related_work}

Modeling articulated objects is a fundamental problem in computer vision, with applications in digital twins and robotics. It has gained significant traction recently, and many methods have already been  proposed~\cite{mu2021asdf,li2020category,captra,jiang2022ditto,liu2023paris,weng2024neural,sun2023opdmulti,kawana2024detection,jiang2022opd,heppert2023carto,geng2023partmanip,geng2023gapartnet,liu2023building,yu2024gamma,wang2019shape2motion,shi2021self,yan2020rpm,shi2022p,jain2021screwnet,liu2023semi,wu2025reartgs}. However, handling real scenes captured from a handheld device, remains a rarely tackled, challenging, and largely unsolved problem. 

\paragraph{Data-driven articulated object modeling.}
Like us, several approaches already rely on machine learning to infer the structure of articulated objects from known categories. Early works, like A-SDF~\cite{mu2021asdf} extends implicit representations such as DeepSDF~\cite{park2019deepsdf} to model category-level articulations by embedding joint angles within a learned shape code. CAPTRA~\cite{captra} introduces a unified framework for tracking articulated motion from point cloud sequences. ANCSH~\cite{li2020category} builds on the Normalized Object Coordinate Space designed for rigid objects~\cite{wang2019normalized} and generalizes it to represent canonical configurations of articulated objects. However, these methods require training and testing separate models for each object category, hindering their applicability and generalization capabilities.

Ditto~\cite{jiang2022ditto} predicts both the geometry and relative state change using two 3D point clouds captured by multiple viewpoints; it is also limited to a single part. 

The method in~\cite{ota-hsaur} also reasons about articulated parts but follows a very different strategy: it predicts potential movable components and physically interacts with them to validate these hypotheses. In contrast, our approach derives hypotheses directly from observed motions.

The key difference between these earlier learning-based approaches and ours is that we take as input a video sequence, which provides information that is rich but not straightforward to exploit correctly. Moreover, our representation of the video makes our method generalize well from synthetic videos to real ones.

\paragraph{Articulated object modeling from a point cloud sequence.}
Some methods consider a temporal sequence of points.
Reart~\cite{liu2023building} and \cite{chao2025part} are optimization-based methods, which look for a tree of piecewise rigid bodies, and generalize to various categories. We compare to Reart~\cite{liu2023building} in the Results section; unfortunately, the code for \cite{chao2025part} is not available yet.
Shi et al.~\cite{shi2021self} consider ``complete'' point clouds. Such point clouds capture the entire geometry of the object at each time step, while we limit our point clouds to points that are visible from a single camera to make our method more practical. 
With P$^3$-Net~\cite{shi2022p}, they use recurrent networks to predict parts and motion parameters, while our architecture is based on more modern and efficient transformer blocks with linear assignment for predicting an arbitrary number of joint parameters with high accuracy.
Zhong et al.~\cite{zhong2023multi} learns to perform rigid body segmentation in an unsupervised way, however they do not consider predicting joint parameters such as axes of rotation. 

\paragraph{Articulation object reconstruction from multi-view input.}

Instead of relying on learning-based priors, \cite{noguchi-cvpr22-watchitmove} leverages geometric cues from multiple video sequences of a moving object to infer articulation. To reduce the need for multiple sequences, PARIS~\cite{liu2023paris} and DigitalTwin~(DTA)~\cite{weng2024neural} simplify the setup by requiring only two sets of images, each capturing the object in a fixed configuration. These methods then identify and recover the articulated parts that change between the two sets. Wu et al.~\cite{wu2025reartgs} focus on recovering the object geometry using Gaussian Splatting~\cite{kerbl20233d}, but do not aim at predicting the joint parameters nor segmenting the object parts.
Methods like~\cite{peng2025itaco, ai2026Articulation, kerr2024rsrd} estimate joint parameters using object scans and demonstration videos; however, \vincent{this paradigm requires a prior stage for capturing the scan by taking multi-view images, while our method does not require such stage.}

\paragraph{Articulated object reconstruction from single-view input.}
Several methods predict the joints from a single point cloud~\cite{yu2024gamma,liu2023semi,wang2019shape2motion,geng2023gapartnet,geng2023partmanip,yan2020rpm}. 
Those methods often predict both part segmentation and joint parameters, using a point-based neural network, except for RPM-Net~\cite{yan2020rpm} which predicts a future motion from a single frame. While we take a video sequence as input, we compare to GAMMA~\cite{yu2024gamma}, the most recent of these methods, for reference.

\paragraph{Articulated object modeling from a single video input.}
Like us, REACTO~\cite{song2024reacto}, ~\cite{artykov-iccvw25-articulated}, Articulate-Anything~\cite{le2024articulate}, and Artipoint~\cite{werby2025articulated} consider a single video as input. REACTO reconstructs surface geometry but does not estimate joint parameters, whereas Articulate-Anything predicts joint parameters for single-part objects but does not reconstruct geometry—instead retrieving a mesh from an existing database. \arslan{Artipoint utilizes hand-centric priors and persistent 2D trajectories to track keypoints, providing the geometric foundation for extracting articulated joint parameters. We also compare to Articulate-Anything and Artipoint, and show that our method significantly outperforms them.}

\section{Method}

\begin{figure*}[t]
    \centering 
    \includegraphics[width=1\textwidth]{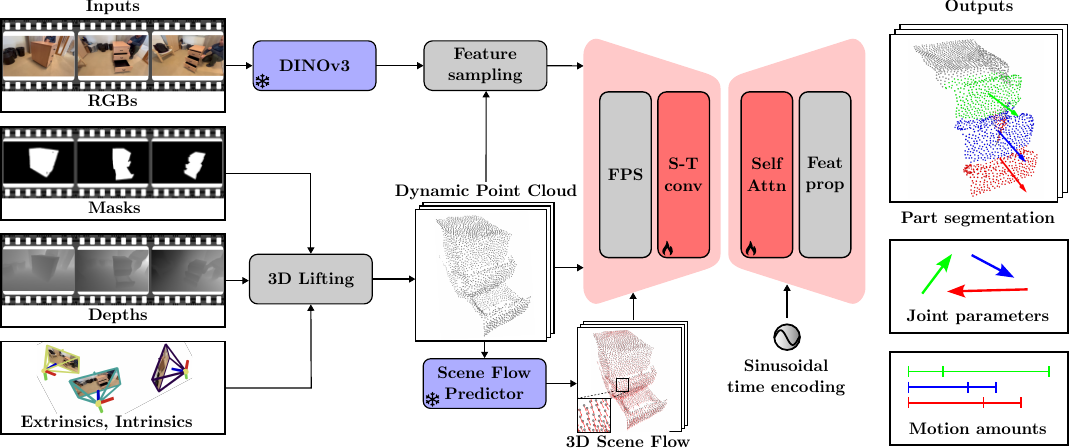}
    \caption{\textbf{Method Overview.} Our method takes as input a sequence of images, from which we get the masks of the objects, the depth maps, and the camera parameters. We sample 2D  points over the masks, lift them to 3D, and augment them with their scene flows and DINOv3 features. From this input, we predict the parts, joint parameters for each part, and amounts of motion for each part and each time step.}
    \label{fig:overview} 
\end{figure*}

Our goal is to segment the target object into articulated parts and predict its joint parameters from a casually captured video sequence. To that end, we use as input 3D points randomly sampled on the target object's surface in each frame. We train our model purely with synthetic data and generalize to real-world objects during inference time without any additional finetuning. For inference on real-world data, we rely on ViPE~\cite{huang2025vipe} to obtain depth maps and camera parameters, however, during training we use only simulated data. We describe our approach in more details below.

\subsection{Problem Formulation}

Given a monocular video sequence made of $T$ images, we build a sequence of point clouds $P_t = \{p_t^i\}_{i=1}^{N_p}$, where each point $p_t^i \in \mathbb{R}^3$ is expressed in the coordinate system of the first camera. Each point $p_t^i$ is obtained by sampling a random pixel $m_t^i$ on the object mask for frame $t$ and lifting it to 3D using the depth map for frame $t$. We jointly predict a part label for each point $p_t^i$ and the joint parameters for each part. In practice, for real sequences, we use the open-source tool ViPE to acquire camera parameters with depth maps and SAM2~\cite{ravi2024sam2} to calculate the background masks from the real-world data. 
ViPE performs surprisingly well, but can still make mistakes. However, our approach is robust to reconstruction artifacts from ViPE: For instance, in the Eyeglasses example of Figure~\ref{fig:real_qualitative}, ViPE returns noisy camera poses, yet our method works correctly.
Similarly, SAM2 works remarkably well. We use $N_p=2048$ in all our experiments and the point coordinates are normalized into  range $[-1, 1]$.

Compared to alternative representations like point tracks, this representation of the input video is significantly more robust and trivial to obtain. It also generalizes well from synthetic data to real data without requiring any domain adaptation: Computing it for training on synthetic data involves sampling synthetic depth maps; obtaining it for inference on real data involves sampling real depth maps.  These real depth maps can be noisy, but our strong performance shows that our method is robust to this noise. Another difference can arise between synthetic point clouds and real point clouds in which data can be missing due to occlusions, for example. However, since the representation is derived exclusively from points visible from a single viewpoint, it remains consistent across both synthetic and real-world data.

\subsection{Point Cloud Sequence Feature Extraction}

Using point sampling only for our representation of the input videos performs already remarkably well. Optionally, we attach to the points their scene flow, which we obtain by tracking the points over a single time step. This adds motion information without introducing large errors like long tracks would do. We also consider the point semantics as captured by DINOv3~\cite{simeoni2025dinov3}. The domain gaps for scene flow and DINOv3 also appear to be negligible, and we can incorporate them to the model input while using synthetic data for training.

More exactly, for each point $p_t^i$, we 
estimate its scene flow $f_t^i \in \IR^3$, i.e., its 3D translation between time steps $t$ and $t+1$. For real-world data, we use GMSF~\cite{zhang2023gmsf} to calculate per-point scene flow. 
We use the DINOv3 model to extract per-frame feature maps $\Psi_t \in \IR^{H\times W\times 1024}$ from each frame of the video sequence, where  $H\times W$ is the resolution of the feature map.  Per-point features are then sampled from $\Psi_t$ using bilinear grid sampling and projected by a small MLP to a 3-dimensional space, yielding the final image features $\phi_t^i$  for each point $p_t^i$.
As our experiments show, these features are not required but provide a performance boost.

\subsection{Encoder}

The encoder part of our architecture is inspired by the one introduced by \cite{fan2021pstnet} for point cloud video processing tasks such as 3D action recognition. However, we made critical changes to properly introduce the scene flow information and the DINO features as additional input, which provide a performance boost. Similarly, we show how to output the joint parameters for each object parts.

Like \cite{fan2021pstnet}, we first subsample $T \times N_k$ keypoints obtained by applying farthest point sampling~(FPS)~\cite{qi-nips17-pointnet++} to each input point cloud $P_t$.  We can then compute a feature vector $\bbf(k_t^i)$ for each keypoint $k_t^i$:
\begin{equation}
    \bbf(k_t^i) = \sum_{(p, t') \in G(k_t^i)} [(p-k_t^i)^\top, t' - t, \bar{v}, \bar{\phi}, \bar{t}]^\top \> ,
\label{eq:conv_feats}
\end{equation}

where $G(k_t^i)$ is the set of points $p$ in the neighborhood of $k_t^i$ and their corresponding time step $t'$ in the spatio-temporal neighborhood of keypoint $k_t^i$. This spatio-temporal neighborhood is defined by an interval of $r_t$ time steps and spatial radius $r_s$. Compared to \cite{fan2021pstnet}, we add at this stage the scene flow information and the DINOv3 features. More exactly, $\bbf(k_t^i)$ depends on $\bar{v}$ and $\bar{\phi}$, the  average scene flow and the average DINOv3 feature in neighborhood $G(k_t^i)$. We also inject the normalized time information $\bar{t} = t/T$, as we observed that it improves the joint prediction. 

We then apply 2 convolutional layers and an MLP to the $\bbf(k_t^i)$ and   obtain $\bbf_e(k_t^i)$ vectors of dimension 512 as output of our encoder.

\subsection{Decoder}

Fan et al.~\cite{fan2022point} uses directly the spatio-temporal features $\bbf_e(k_t^i)$ as input to a self-attention mechanism. We also add time information again by giving as input to our decoder feature vectors of the form
\begin{equation}
\bff'(k_t^i) = [\bff_e(k_t^i); \gamma(t)] \> ,
\end{equation}
for each vector $\bff_e(k_t^i)$ output by the encoder. $\gamma(t) = [\sin(2^k \pi t), \cos(2^k \pi t)]_{k=0}^L$ is a positional encoding for frame index $t$.

We use $L=6$. 
We then perform video-level self-attention on the $\bff'$ vectors.

Finally, we propagate features $\bff''(k_t^i)$ from the keypoints $k_t^i$ to the original 3D points $p_t^i$:
\begin{equation}
\bff''(p_t^i) = \frac{\sum_{k\in G_p(p_t^i)} w(k, p_t^i) \bff''(k)}{\sum_{k\in G_p(p_t^i)} w(k, p_t^i)} 
\end{equation}
where $w(k, p_t^i) = \frac{1}{||k - p_t^i||^2}$ and $G_p(p)$ is the set of nearest neighbors of point $p$.

\subsection{Part Labels and Joint Parameters Prediction}

Computing $\bd(p_t^i) = \text{softmax}(\bff''(p_t^i) . Q^\top)$, where matrix $Q$ is made of learnable queries, gives us a distribution over the parts for each point $p_t^i$ , predicting to which part $p_t^i$ belongs to.

We also extend \cite{fan2022point} to predict the joint parameters for each part. To this end, we aggregate the per-point features $\bff''(p_t^i)$ into per-part features $\bff_\text{part}(m)$, where $m$ is the part index:
\begin{equation}
\bff_\text{part}(m) = \frac{\sum_t \sum_i \bd(p_t^i)_m \bff''(p_t^i)}{\sum_t \sum_i \bd(p_t^i)_m} \> .
\end{equation}

We feed each per-part feature $\bff_\text{part}(m)$ to an MLP with multiple heads, which outputs joint type, translation axis, rotation axis, and pivot point.

We parametrize revolute joints by a rotation axis in the form of a unit vector $\ba_r \in \IR^3$, a pivot point $\bp \in \IR^3$, and a rotation amount $\br_t \in \IR$ for each time step $t$. Prismatic joints are parametrized by a joint axis in the form of a unit vector $\ba_p \in \IR^3$ and a translation amount $\bd_t \in \IR$ for each time step $t$. 

In addition to the revolute and prismatic joint parts, we also consider a 'static' part for the object as some parts do not move.
We predict the joint type and three sets of parameters, corresponding to an axis and a pivot point for the revolute joint, or another axis for the prismatic joint. At inference, only the relevant parameters are used.

\paragraph{Amount of motion prediction.}
We also need to predict the amount of motion~(rotation or translation) for each part and each time step. To do so, we aggregate features \corentin{by calculating} one feature vector for each part at each timestep:
\begin{equation}
\bbf_\text{motion}(m, t) = \frac{\sum_i \bd(p_t^i)_m \bff''(p_t^i)}{\sum_i \bd(p_t^i)_m} \> .
\end{equation}
Then, we feed the aggregated features to a specific decoding MLP head, to predict the per-part rotation/translation amount at each timestep.

\subsection{Loss Functions}

We predict a fixed number $M=20$ of parts, which is generally much greater than the true number of parts in the ground-truth labels. The ordering of the predicted parts has no reason to match the ordering of the ground-truth ones. To solve this, we use the Hungarian algorithm as in \cite{detr} to match the predictions to the ground-truth ones during training. The cost of assignment is computed as a weighted sum given by the binary cross entropy loss and the dice loss between the predicted and ground truth parts:
\begin{equation}
     \calL_\text{partlabels} = w_{\text{BCE}} \calL\>_\text{BCE} + w_{\text{Dice}} \calL\>_\text{Dice} \> ,
\label{eq:labels}
\end{equation}
where $w_{\text{BCE}}  = 2$ and $w_{\text{Dice}} = 1$. We also use this cost as a loss term to learn part segmentation.
To predict the joint types, we use the cross-entropy loss:
\begin{equation}
     \calL_{\text{jointtypes}} = - \frac{1}{M}\sum_{i=1}^{M} \log \frac{\exp(z_{i, y_i})}{\sum_{c=1}^{3} \exp(z_{i, c})} \> ,
\label{eq:jointtypes}
\end{equation}
where $y_i \in \{ 0, 1, 2 \}$ represents the ground-truth joint type for part $i$, in which 0, 1, 2 corresponds to revolute, prismatic, and static joint, respectively. 

For predicting the axis of rotation, we need to predict a 3D line, which we parameterize by a 3D point (usually called 'pivot point' in this context) and a direction vector. The translation axis is parametrized by a direction vector only. To learn predicting the direction vector, we use the geodesic loss on the rotation between the predicted and ground-truth direction:
\begin{equation}
     \calL_{\text{axes}} = \frac{1}{M} \sum_i \text{arccos}(\;\text{clamp} (\hat{a}_i . a_i, -1, 1)\;) \> ,
\label{eq:axes}
\end{equation}
where $a_i$ is the ground truth direction vector for part $i$, and $\hat{a}_i$ the predicted direction vector.
Since pivot points can be predicted anywhere along the axis, we use the point-to-line distance between the predicted pivot point and the line defined by the ``ground truth'' pivot point and the ground truth axis direction:
\begin{equation}
     \calL_{\text{pivots}} = \frac{1}{M} \sum_{i=1}^{M} ||a \times (\hat{c}_i - c_i)|| \> ,
\label{eq:pivots}
\end{equation}
where $\hat{c}_i$ and $c_i$ represent predicted and ground-truth pivot points, respectively. 
For the motion amount prediction, we use the L1 loss:
\begin{equation}
     \calL_\text{motionamounts} = \frac{1}{MT} \sum_{t=1}^T \sum_{i=1}^M |\hat{m}_{i,t} - m_{i,t} | \> ,
\label{eq:BIC}
\end{equation}
where $m_{i,t}$ is the ground truth amount of motion for part $i$ at time $t$, which can be a translation in centimeters or a rotation in degrees, and $\hat{m}_{i,t}$ the corresponding predicted value.
The final loss is simply the sum of these losses, without weights:
\vspace{-0.2cm}
\begin{equation}
\calL =
\lambda_\text{jt} \calL_\text{jointtypes} + 
\lambda_\text{a}\calL_\text{axes} + 
\lambda_\text{p}\calL_\text{pivots} + 
\lambda_\text{pl}\calL_\text{partlabels} + 
\lambda_\text{ma}\calL_\text{motionamounts} \> .
\label{eq:loss}
\end{equation}
In practice, we simply use $\lambda_\text{jt} = \lambda_\text{a} = \lambda_\text{p} = \lambda_\text{pl} = \lambda_\text{ma} = 1$ for all our experiments.

\section{Experiments}


As mentioned earlier, many existing approaches require multi-view captures or object scanning under different configurations. Instead, our method takes as input a single monocular video captured with a freely moving camera. \cite{artykov-iccvw25-articulated} proposed a method that takes the same input, but did not release the code. Comparisons with previous methods is therefore not easy. We compare here our method to GAMMA~\cite{yu2024gamma}, Reart~\cite{liu2023building}, Video2Articulation~\cite{peng2025itaco}, Articulate-Anything~\cite{le2024articulate}, and Artipoint~\cite{werby2025articulated}, which input and output are the closest to ours. To further highlight the consistent accuracy and robustness of our method, we created a custom baseline method, called FeatClust.

GAMMA is a feed-forward method taking as input a point cloud of an articulated object at a single articulation state and predicts per-point part labels along with joint parameters. We retrained GAMMA on our point cloud sequences, and during inference, the final joint parameters are obtained by averaging the per-timestep joint predictions across the sequence. 

Reart, Video2Articulation, Articulate-Anything, and Artipoint are geometry- and optimiza\-tion-based methods. Reart is the most comparable recent method to ours in terms of input and output: They predict part labels for the point set at the chosen canonical timestep along with joint parameters from the input point cloud sequence. Video2Articulation is a very recent method~(3DV'26) and predicts movable part label and joint parameters of a two-part object given a scan and a scan-aligned video demonstration of an object. Articulate-Anything is a state-of-the-art method predicting joint parameters for the object part meshes retrieved from a mesh database given an input video. As Articulate-Anything and Video2Articulation are designed to predict joint parameters for two-part only objects, we restrict its comparison to two-part objects. Finally, Artipoint recovers joint parameters from an RGB-D video using human-object interaction cues. They sample points around hand masks and prompt an off-the-shelf instance segmentation model to segment the object. Then, they track the sampled points with CoTracker3~\cite{karaev2024cotracker3} and calculate the articulation model of the object via factor graph formulation. Since Artipoint requires hand priors, we compare against them using only our real-world data. 

We also consider a simple baseline we call FeatClust. The point of FeatClust is to evaluate whether pretrained features and motion cues alone are sufficient for part segmentation and motion estimation. FeatClust uses the same preprocessing as our method, including DINO features, depth, and scene flow. It clusters the points using k-means in a joint appearance–motion feature space to obtain part segmentation, and estimates the rigid motion for each cluster over time to then obtain the joint parameters. We provide more technical details in the supplementary material.

We implemented our method using PyTorch~\cite{paszke-19-pytorch}. We trained and tested our model on a single A100 GPU, with inference time of 6.25 ms/frame on average.

\subsection{The 4art-synth and 4art-real Datasets}

As mentioned in the introduction, our 4art-synth dataset is made of videos of 501 different objects, which are split into training, validation, and test sets with a ratio of 70\%, 15\%, and 15\%, respectively. RGB-D sequences were rendered from a monocular camera freely moving around each object in a PyBullet environment~\cite{coumans2016pybullet}. 
To better simulate human motion, we slightly tilt the camera around its viewing axis. The camera begins from random poses sampled on the upper hemisphere of the object and follows a circular, arc-like trajectory of varying length and direction to which we add some noise.

Our 4art-real dataset is made of casual real-world videos of a box, laptop, stapler, eyeglasses, and a cabinet with large camera motions. We manually annotated the videos for ground-truth joints and part labels. Details on how this was done are provided in the supplementary material. These videos are significantly more challenging than the real videos typically used for evaluating the previous methods. A human operator is interacting with the articulated object while another person is capturing the video by moving around the object. We then process the videos with ViPE~\cite{huang2025vipe} to obtain depth maps and camera parameters. We obtain the background masks with SAM2~\cite{ravi2024sam2}.

\subsection{Metrics}
We evaluate part segmentation prediction using the widely adopted mean Intersec\-tion-over-Union~(mIoU) metric. The axis-angle prediction is assessed by the sign-agnostic axis-angle error~(in degrees) between the predicted and ground-truth joint axes, following the evaluation protocol of GAMMA and Articulate-Anything. The axis position~(pivot point) error~(in centimeters) is computed as the point-to-line distance between predicted and ground-truth axes. 

Additionally, we report 'part rotation' and 'part translation' errors, the motion's magnitude prediction errors per frame averaged over the sequence. Those errors correspond to the geodesic distance for revolute joints and the Euclidean distance for prismatic joints. We also report joint type classification accuracy~(in percents), corresponding to the proportion of correctly predicted joint types. 

\sisetup{detect-all=true}
\begin{table*}[t]
\centering
\def\mywidth{1.\textwidth}
\caption{
\textbf{Quantitative results on 4art-synth.} `F' indicates that the method fails on the category. `N/A' indicates that the method does not predict the values for the metric. `\texttimes' indicates the method is not applicable to the category. Since Artipoint requires human-object interaction for kinematic modeling, it cannot be evaluated on synthetic data, and we limit its evaluation to the real-world sequences.
We provide the means and standard deviations for sim2art computed for each category over all the videos and all the joints.
}
\resizebox{\mywidth}{!}{
\sisetup{table-auto-round,table-format=.2,table-column-width=1.6cm}

\begin{tabular}{@{}cc|ccccccccccccccc}

\toprule

\multicolumn{1}{c}{\textbf{Metrics}} &
\multicolumn{1}{c|}{\textbf{Method}} &      
\multicolumn{1}{c}{Storage1} & 
\multicolumn{1}{c}{Storage2} & 
\multicolumn{1}{c}{Storage3} & 
\multicolumn{1}{c}{Storage4} & 
\multicolumn{1}{c}{Storage5} & 
\multicolumn{1}{c}{Storage6} & 
\multicolumn{1}{c}{Box} &
\multicolumn{1}{c}{Eyeglasses} &
\multicolumn{1}{c}{Laptop} &
\multicolumn{1}{c}{Stapler} &
\multicolumn{1}{c}{Table1} & 
\multicolumn{1}{c}{Table2} &
\multicolumn{1}{c}{Fan} &
\multicolumn{1}{c}{Scissors} &
  
\multicolumn{1}{c}{Mean} \\

\midrule

\multirow{6}[2]{*}{mIoU $\uparrow$} 
& GAMMA~\cite{yu2024gamma} & 0.27 & F & F & F & F & 0.30 & F & 0.69 & F  & 0.31 & 0.22 & F &0.28 &0.41  & 0.35\\

& Video2Articulation~\cite{peng2025itaco} & \texttimes & \texttimes & 0.57 & \texttimes & \texttimes & \texttimes & F & \texttimes & 0.50  & 0.26 & 0.54 & \texttimes &0.53 &0.54  &0.49 \\

& Reart~\cite{liu2023building} & F & 0.54 &\textbf{0.94} & F &F &F  &0.93 &0.43  &0.87  & 0.82 &0.62 &F &0.49 &0.72  &0.71 \\
                               
& Articulate-Anything~\cite{le2024articulate} & \texttimes & \texttimes & N/A & \texttimes & \texttimes & \texttimes & N/A & \texttimes & N/A & N/A & N/A & \texttimes &N/A  &N/A &N/A \\
                               
& FeatClust &0.16 & 0.25 &0.72 & 0.16   &0.098  & 0.20  &0.67 &0.24 &0.68 &0.70 &\textbf{0.91} &0.26 &0.46 &0.49  &0.43 \\

\rowcolor[rgb]{ .988,  .894,  .839}  \cellcolor[rgb]{1, 1, 1}                               
& sim2art (ours) & \textbf{0.91} & \textbf{0.78} & 0.80 &\textbf{0.95} &\textbf{0.87} &\textbf{0.99} &\textbf{0.99}  &\textbf{0.98}  &\textbf{0.98}   &\textbf{0.99}  &0.79  &\textbf{0.91}  &\textbf{0.54} &\textbf{0.97} &\textbf{0.89}  \\

\midrule

\multirow{6}[2]{*}{\shortstack{Axis\\Ang} (°) $\downarrow$} 
& GAMMA~\cite{yu2024gamma} & 6.86 & F & F & F & F &12.56 & F &10.23 &F  &8.68 &6.69 &F &28.87 &30.57  &14.92 \\

& Video2Articulation~\cite{peng2025itaco} & \texttimes & \texttimes & 18.00 & \texttimes & \texttimes & \texttimes & F & \texttimes & 45.93  & 38.17 & 79.36 & \texttimes &56.06 &34.02  &45.26 \\

& Reart~\cite{liu2023building} &F &65.75 &17.90 &F &F &F &6.48 &55.76 &4.97  &49.71 &74.14 &F &29.74 &14.92  &35.49 \\

& Articulate-Anything~\cite{le2024articulate} & \texttimes & \texttimes &40.00 &  \texttimes &  \texttimes &  \texttimes &\textbf{0.00} & \texttimes &\textbf{0.14}  &\textbf{0.00} &\textbf{0.00} & \texttimes &18.00 &F  &9.69 \\


& FeatClust &55.18 & 58.91 & 12.19 & 33.16   & 41.13  & 67.53  &1.55 &28.03 &1.17 &4.20
    & 0.53 &80.84 &52.11 &16.98 &32.39\\

\rowcolor[rgb]{ .988, .894, .839} \cellcolor[rgb]{1, 1, 1} 
& sim2art (ours) &\textbf{3.21\tsb{7.29}}   &\textbf{3.08\tsb{1.70}}  &\textbf{2.90\tsb{2.43}}  &\textbf{2.53\tsb{1.36}} &\textbf{3.18\tsb{3.95}} &\textbf{4.02\tsb{4.94}} & 8.60\tsb{2.22} &\textbf{3.63\tsb{2.05}} &3.31\tsb{3.22}  &4.57\tsb{3.79} &3.13\tsb{1.73} &\textbf{12.35\tsb{14.86}} &\textbf{10.18\tsb{14.04}} &\textbf{6.20\tsb{5.80}} &\textbf{5.06\tsb{4.96}}  \\

\midrule

\multirow{6}[2]{*}{\shortstack{Axis\\Pos} (cm) $\downarrow$} 

& GAMMA~\cite{yu2024gamma} &  - & F & F & F & F & 37.57 & F &39.83 &F  &29.47 & - & - &32.00 &24.53  &32.68 \\

& Video2Articulation~\cite{peng2025itaco} & \texttimes & \texttimes & 102.46 & \texttimes & \texttimes & \texttimes & F & \texttimes & 112.47  & 175.95 & - & \texttimes &61.56 &61.49  &102.79 \\

& Reart~\cite{liu2023building} & - &121.40 &71.33 &F &F &F &27.57 &134.97 &24.00 &197.89
& - &- &73.61 &82.57  &91.67\\

& Articulate-Anything~\cite{le2024articulate} & \texttimes & \texttimes &28.11 &  \texttimes & \texttimes &  \texttimes & 61.13 & \texttimes &19.94  &18.21 & - & \texttimes &\textbf{1.07}  &F &25.69 \\

& FeatClust &- & 69.18 & 16.59 & 49.75   & 69.54  & 47.26  &\textbf{4.08} &86.40 &4.73 &\textbf{7.72} &- &- &55.19  &54.20 & 42.24\\

\rowcolor[rgb]{ .988, .894, .839} \cellcolor[rgb]{1, 1, 1} & 
sim2art (ours) &-  &\textbf{6.29\tsb{3.22}} &\textbf{6.40\tsb{4.82}} &\textbf{5.65\tsb{3.19}} &\textbf{5.73\tsb{1.53}} &\textbf{7.35\tsb{8.81}} &4.54\tsb{2.21} &\textbf{13.77\tsb{8.35}} &\textbf{4.58\tsb{5.15}}  &10.43\tsb{6.11} &- &- &5.98\tsb{4.90} &\textbf{10.96\tsb{6.95}} &\textbf{7.43\tsb{5.02}} \\

\midrule

\multirow{6}[2]{*}{\shortstack{Part\\Rotation} (°) $\downarrow$} 

& GAMMA~\cite{yu2024gamma} & - & N/A & N/A & N/A & N/A & N/A & N/A & N/A & N/A & N/A &- & - &N/A &N/A  & N/A \\

& Video2Articulation~\cite{peng2025itaco} & \texttimes & \texttimes & 18.79 & \texttimes & \texttimes & \texttimes & F & \texttimes & 13.47  &\textbf{2.05} & - & \texttimes &91.52 &\textbf{8.09}  &26.78 \\

& Reart~\cite{liu2023building} &- &88.97 &11.16 &F &F &F &\textbf{4.64} &60.14 &15.22 &25.92 & - &- &115.43 &34.69  &44.52\\

& Articulate-Anything~\cite{le2024articulate} & \texttimes & \texttimes &  N/A &  \texttimes &  \texttimes  &  \texttimes  &  N/A  & \texttimes  & N/A   & N/A & - & \texttimes &N/A &N/A  & N/A \\
                            
& FeatClust &- & 10.61 & 31.60 & 30.26   &  25.49  & 16.31  &33.07 &15.37 &27.20 &21.99 &- &- &101.96  &12.06 & 29.63\\

\rowcolor[rgb]{ .988,  .894,  .839}  \cellcolor[rgb]{1, 1, 1}
& sim2art (ours) & - &\textbf{1.88\tsb{1.32}} &\textbf{1.97\tsb{1.63}} &\textbf{2.15\tsb{1.18}}   &\textbf{2.26\tsb{1.58}}     &\textbf{2.57\tsb{4.50}} &6.35\tsb{4.37} &\textbf{4.50\tsb{4.05}}  &\textbf{5.73\tsb{5.82}}   &3.92\tsb{2.51}  &-  &-  &\textbf{13.91\tsb{14.68}} &14.11\tsb{17.00} &\textbf{5.40\tsb{5.33}}  \\

\midrule

\multirow{6}[2]{*}{\shortstack{Part\\Translation} (cm) $\downarrow$} 

& GAMMA~\cite{yu2024gamma} & N/A & - & - & N/A & N/A & N/A & - & - & - & - & N/A & N/A &- &-  & N/A \\

& Video2Articulation~\cite{peng2025itaco} & \texttimes & \texttimes & - & \texttimes & \texttimes & \texttimes & - & \texttimes & -  & - & 13.68 & \texttimes &- &-  &13.68 \\

& Reart~\cite{liu2023building} &F &- &- &F &F &F & - & - & -  &- & F & F &- &- &F \\

& Articulate-Anything~\cite{le2024articulate} & \texttimes & \texttimes &  N/A &  \texttimes &  \texttimes  &  \texttimes  &  -  & -  & -  & -  & N/A & \texttimes &- &-  & N/A \\
                            
& FeatClust &11.29 & - & - & 36.86   & 14.03  & 45.36  & - &- &- &- &23.82 &26.05 &- &- &26.24 \\

\rowcolor[rgb]{ .988,  .894,  .839}  \cellcolor[rgb]{1, 1, 1}
& sim2art (ours) &\textbf{5.11\tsb{7.31}}  &-  &-  &\textbf{2.14\tsb{2.17}}   &\textbf{1.48\tsb{2.84}}     &\textbf{2.54\tsb{2.74}}   &-  &-  &-   &-  &\textbf{5.66\tsb{3.18}}  &\textbf{4.85\tsb{6.22}}  &- &- &\textbf{3.63\tsb{4.08}}  \\

\midrule

\multirow{6}[2]{*}{\shortstack{Type\\Accuracy} ($\%$) $\uparrow$}

& GAMMA~\cite{yu2024gamma} &\textbf{100} & F & F &F & F & 90.91 & F &\textbf{100} &F  &\textbf{100} &\textbf{100} &F &\textbf{100} &\textbf{100} &98.70  \\

& Video2Articulation~\cite{peng2025itaco} & \texttimes & \texttimes & \textbf{100} & \texttimes & \texttimes & \texttimes & F & \texttimes & \textbf{100}  & \textbf{100} & \textbf{100} & \texttimes &\textbf{100} &\textbf{100}  &\textbf{100} \\

& Reart~\cite{liu2023building} &F &\textbf{100} &\textbf{100} &F &F &F &\textbf{100} &\textbf{100} &\textbf{100}  &\textbf{100} & 0.00 &F &\textbf{100} &\textbf{100} &88.88 \\
 
& Articulate-Anything~\cite{le2024articulate} & \texttimes & \texttimes &88.89 & \texttimes & \texttimes & \texttimes &\textbf{100} & \texttimes &\textbf{100}  &\textbf{100} &\textbf{100} & \texttimes &85.71 &F &95.77 \\
 
& FeatClust &\textbf{100} & 71.43 & 68.75 & \textbf{100}   &  \textbf{100}  & 75.00  & \textbf{100} &55.55 &87.50 &50.00 & \textbf{100} &\textbf{100} &83.33 &\textbf{100} &85.11 \\

\rowcolor[rgb]{ .988, .894, .839} \cellcolor[rgb]{1, 1, 1}
& sim2art (ours) &98.55  &\textbf{100}  &\textbf{100}  &99.52  &99.85 &\textbf{100} &\textbf{100} &\textbf{100} &\textbf{100}  &\textbf{100} &\textbf{100} &92.53 &\textbf{100} &75.00 &97.32  \\

\bottomrule
\end{tabular}%
}
\label{tab:quant_table}
\end{table*}

\sisetup{detect-all=true}
\begin{table}[t]
\centering
\def\mywidth{0.65\columnwidth}
\caption{
\textbf{Quantitative results on 4art-real.} (*) Reart does not often return the correct number of joints; we consider the predicted  joints with the smallest errors, giving it a significant advantage. Yet, our method is significantly better. 
}
\resizebox{\mywidth}{!}{
\sisetup{table-auto-round,table-format=.2,table-column-width=1.6cm}

\begin{tabular}{@{}cc|cccccc}

\toprule

\multicolumn{1}{c}{\textbf{Metrics}} &
\multicolumn{1}{c|}{\textbf{Method}} &      
\multicolumn{1}{c}{Box} & 
\multicolumn{1}{c}{Laptop} & 
\multicolumn{1}{c}{Stapler} & 
\multicolumn{1}{c}{Eyeglasses} & 
\multicolumn{1}{c}{Drawer} &

\multicolumn{1}{c}{Mean} \\

\midrule

\multirow{6}[2]{*}{mIoU $\uparrow$} 

& Reart~(*) & 0.20 & 0.17 &F & F &0.057 &0.14   \\
                               
& Art.-Anything & N/A & N/A & N/A & \texttimes & \texttimes & N/A \\
                               
& Video2Art. &0.49 &0.53 &0.47 &\texttimes &\texttimes &0.49 \\

& Artipoint~\cite{werby2025articulated} & N/A  &N/A  &N/A  &N/A &N/A &N/A  \\

& FeatClust &0.10  &0.35  &0.54  &F &0.13 &0.28  \\

\rowcolor[rgb]{ .988,  .894,  .839}  \cellcolor[rgb]{1, 1, 1}                      
& sim2art (ours) &\textbf{ 0.91} &\textbf{ 0.94} & \textbf{0.96} &\textbf{0.54} &\textbf{0.81}   &\textbf{0.83} \\

\midrule

\multirow{6}[2]{*}{\shortstack{Axis\\Ang} (°) $\downarrow$} 


& Reart~(*) &\textbf{2.87} &13.63 &F &F &21.71 &12.74 \\

& Art.-Anything & F & 10.30 &12.23 &  \texttimes &  \texttimes &  14.38 \\

& Video2Art. &89.02 &87.31 &82.17 &\texttimes &\texttimes &86.16 \\

& Artipoint~\cite{werby2025articulated} & 7.54  &38.58  &86.00  &40.80 &52.58 &45.10  \\

& FeatClust &4.10  &83.83  &34.05  &F &63.90 &46.47  \\

\rowcolor[rgb]{ .988, .894, .839} \cellcolor[rgb]{1, 1, 1} 
& sim2art (ours) & 13.5   &\textbf{5.50}  &\textbf{6.30}  &\textbf{10.08}  &\textbf{1.77} & \textbf{7.43}  \\

\midrule

\multirow{6}[2]{*}{\shortstack{Axis\\Pos} (cm) $\downarrow$} 


& Reart~(*) &15.12 &11.59 &F &F &- &18.02 \\

& Art.-Anything & F & 14.76 &5.11 &  \texttimes & \texttimes &  6.62  \\

& Video2Art. &135.69 &65.70 &56.06 &\texttimes &\texttimes &85.82 \\

& Artipoint~\cite{werby2025articulated} & 4.07  &52.28  &12.10  &102.71 &- &42.79  \\

& FeatClust &9.96  &39.11  &4.71  &F &- &16.93  \\

\rowcolor[rgb]{ .988, .894, .839} \cellcolor[rgb]{1, 1, 1} 
& sim2art (ours) &\textbf{3.80}  &\textbf{5.02} &\textbf{0.77} &\textbf{2.34} &- &\textbf{2.98}   \\

\midrule

 

                            



\multirow{6}[2]{*}{\shortstack{Type\\Accuracy} ($\%$) $\uparrow$}

 
& Reart~(*) &14.29 &25 &F &F &\textbf{100} &27.86 \\
 
& Art.-Anything & F & \textbf{100} &\textbf{100} & \texttimes & \texttimes & 66.66 \\
 
& Video2Art. &\textbf{100} &\textbf{100} &\textbf{100} &\texttimes &\texttimes &\textbf{100} \\

& Artipoint~\cite{werby2025articulated} & \textbf{100}  &\textbf{100}  &\textbf{100}  &50.00 &33.33 &76.66  \\

& FeatClust &\textbf{100}  &\textbf{100}  &\textbf{100}  &F &\textbf{100} &\textbf{100}  \\

\rowcolor[rgb]{ .988, .894, .839} \cellcolor[rgb]{1, 1, 1}
& sim2art (ours) &\textbf{100}  &\textbf{100}  &\textbf{100}   &\textbf{100} &\textbf{100} &\textbf{100} \\

\bottomrule
\end{tabular}%
}
\label{tab:quant_table_real}
\end{table}

\subsection{Results}

Table~\ref{tab:quant_table} and Figure~\ref{fig:qual_fig_merged} show that our method significantly outperforms existing approaches on synthetic data, while requiring only a single RGB video captured by a freely moving camera. Both GAMMA and Reart fail on several categories, whereas our method achieves state-of-the-art performance across all cases. FeatClust achieves satisfactory results only for geometrically simple objects and requires perfectly accurate scene flow. 

Both Articulate-Anything and GAMMA do not predict part motion. Articulate-Anything and Video2Articulation are limited to one moving part per object.

We also obtain state-of-the-art performance on challenging real-world scenarios, as shown in Table~\ref{tab:quant_table_real} and Figure~\ref{fig:real_qualitative}. Previous methods often over-segment the object, predicting more joints than actually exist. Our approach is robust to reconstruction artifacts from ViPE: For instance, in the eyeglasses example, our method successfully segments the incomplete and deformed object and accurately predicts its joints, whereas the other methods completely fail. As Articulate-Anything and Artipoint do not perform part segmentation, we visualize their outputs using predicted joints alongside retrieved meshes for the former, and joint predictions projected onto the input point cloud for the latter.

\paragraph{Why/when do the previous methods fail?} We emphasize the inherent difficulty of articulation estimation when both the camera and object parts move simultaneously. Reart fits a kinematic model by warping a canonical frame; however, large viewpoint changes and the resulting occlusions significantly degrade the performance of such frame-anchored approach. Similarly, the reliance of Video2Articulation on point correspondences makes them susceptible to errors during substantial camera motion. 
Methods like Articulate-Anything and Artipoint, which depend on long-term 2D tracks and modular off-the-shelf pipelines, exhibit significant fragility when robust trajectories cannot be established.
The FeatClust results indicate that pretrained features and preprocessing alone are insufficient for reliable part segmentation, which results in poor accuracy.

\subsection{Ablation Study}
We conducted several ablation studies to show how various components of our method contribute to overall performance. Table~\ref{tab:ablation} shows that removing the scene flow component~($\bar{v}$ in Eq.~\eqref{eq:conv_feats}) results in significant joint parameters prediction performance drop. Incorporating DINOv3 features and frame indices also clearly help improving the overall performance. 

\subsection{Application}

\tom{
We can easily leverage 2D Gaussian Splatting~\cite{huang20242d} to produce a textured 3D reconstruction of the articulated object from the output of our method, \vincent{as we did for Fig.~\ref{fig:opening}.} To this end, we initialize one splat per point of the dynamic point cloud. The initial  color is obtained from the RGB image, the orientation is given by the point normal computed from the depth map, and the scale is set to the average distance to the nearest neighbors. The predicted joints and segmentation are used to transform the Gaussians from one timestep to the next. We further refine the splat appearance parameters as well as the joint amounts by gradient descent, with an $L_1$ RGB reconstruction objective.
}


\def\qualitWidth{0.20\linewidth}

\newcommand{\epicfail}{
\begin{tikzpicture}[x=\qualitWidth,y=\qualitWidth]
    \useasboundingbox (0,0) rectangle (1,1); 
    \draw[gray, line width=0.1cm] (0.3,0.3) -- (0.7,0.7);
    \draw[gray, line width=0.1cm] (0.3,0.7) -- (0.7,0.3);
\end{tikzpicture}
}

\setlength{\tabcolsep}{0pt}

\begin{figure*}[t]
  \centering

\resizebox{0.6\linewidth}{!}{
\begin{tabular}{c@{$\;$}c@{$\;\;$}cccc}

\rotatebox{90}{\hspace{0.7cm}\vphantom{A}Ground} &
\rotatebox{90}{\hspace{0.8cm}\vphantom{A}Truth} &
\includegraphics[trim={3cm 3cm 3cm 3cm},clip,width=\qualitWidth]{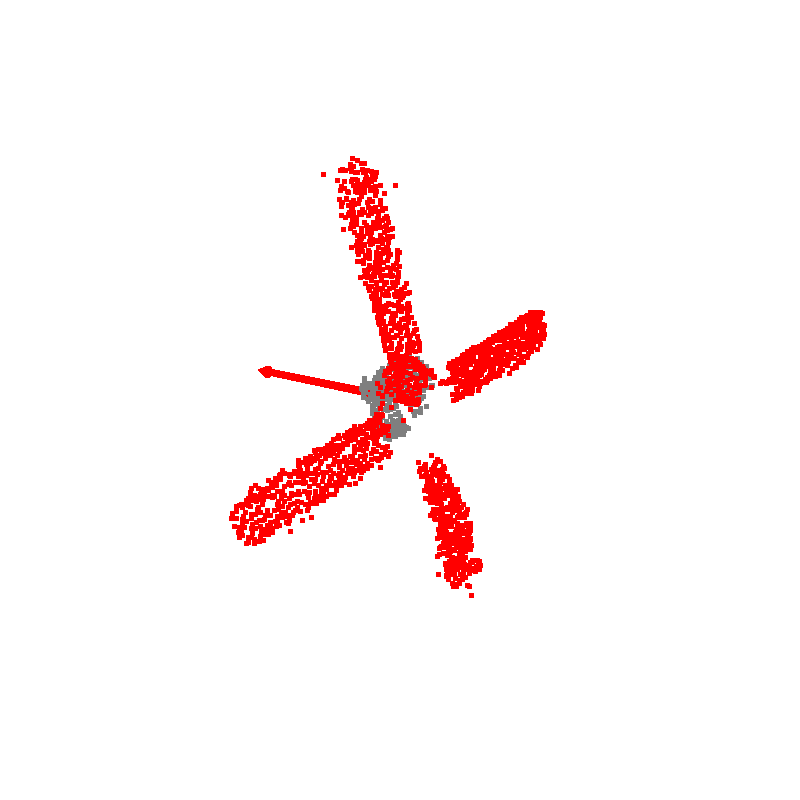} &
\includegraphics[trim={3cm 3cm 3cm 3cm},clip,width=\qualitWidth]{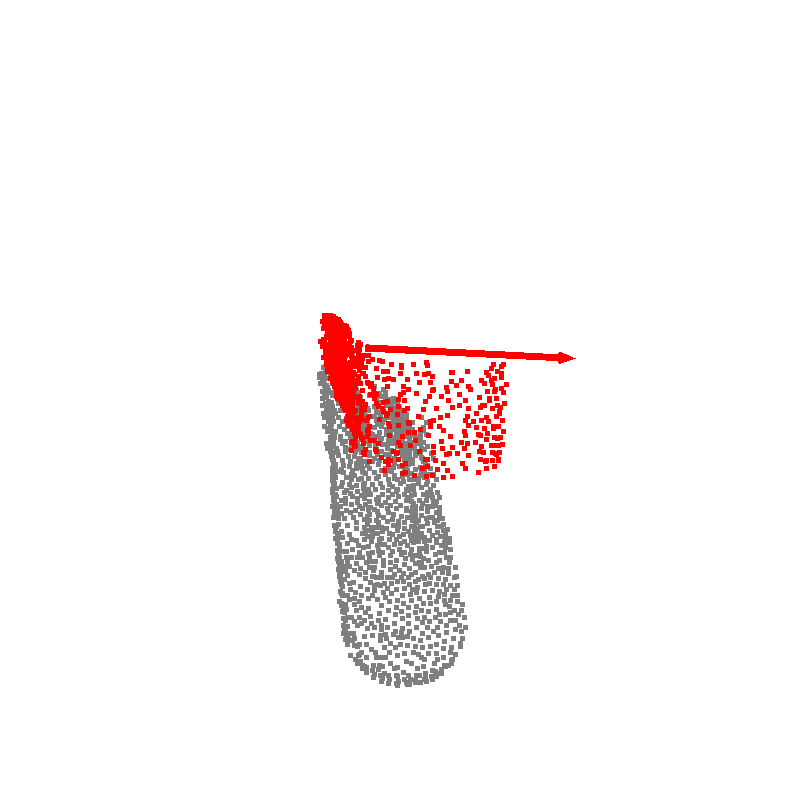} &
\includegraphics[trim={3cm 5cm 3cm 5cm},clip,width=\qualitWidth]{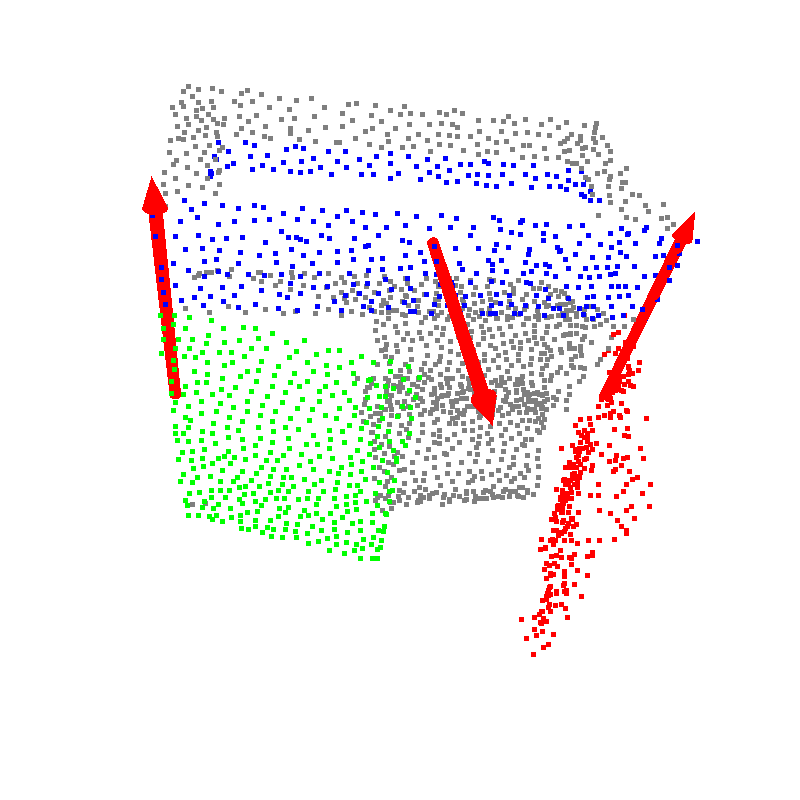} &
\includegraphics[trim={3cm 3cm 3cm 3cm},clip,width=\qualitWidth]{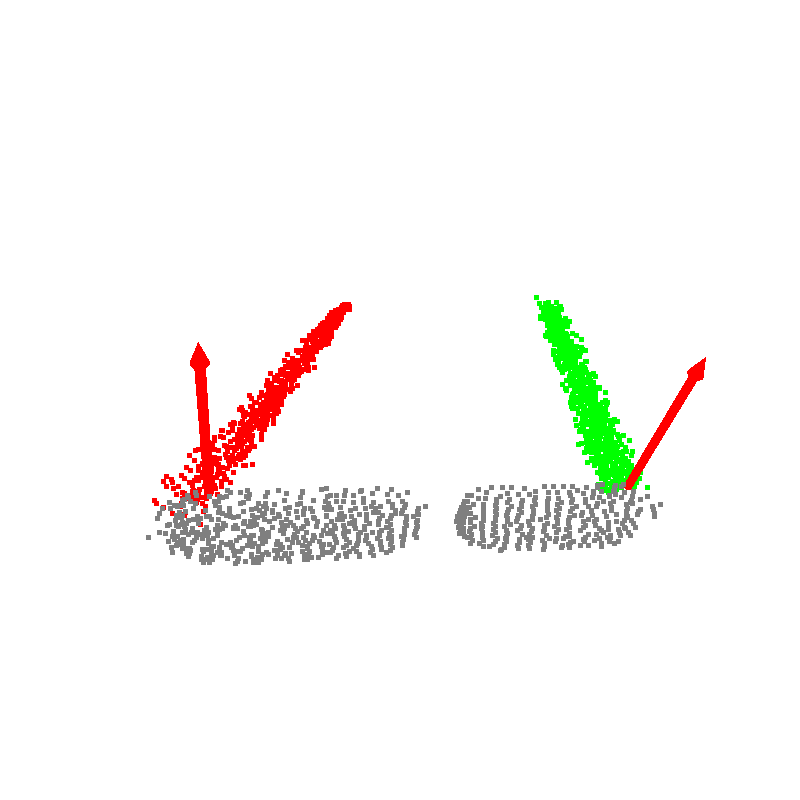}  \\[-0.0cm]

\rotatebox{90}{\hspace{0.7cm}sim2art} &
\rotatebox{90}{\hspace{0.8cm}(ours)} &
\includegraphics[trim={3cm 3cm 3cm 3cm},clip,width=\qualitWidth]{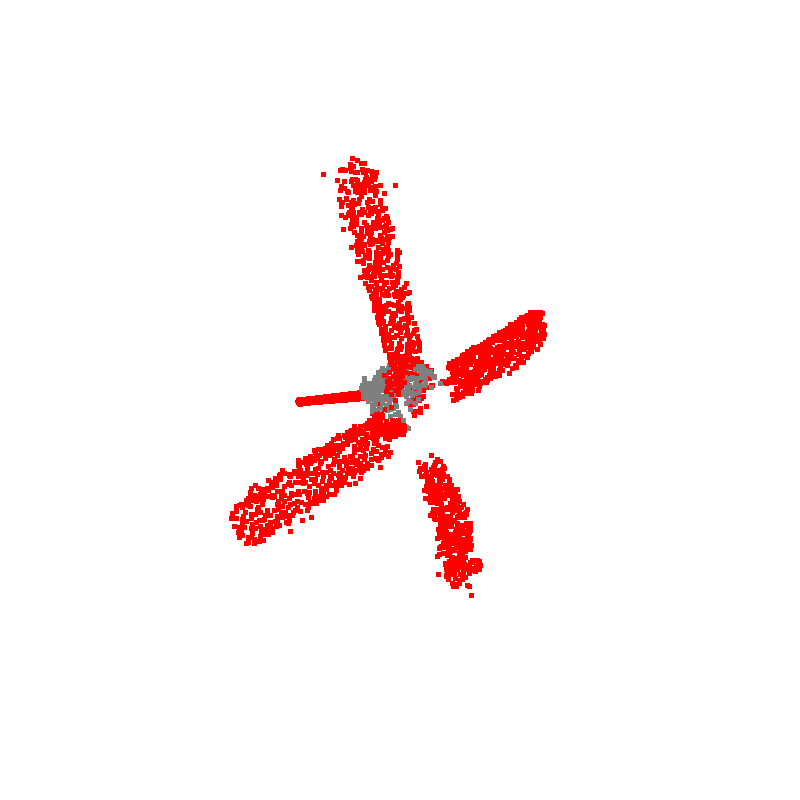}  &
\includegraphics[trim={3cm 3cm 3cm 3cm},clip,width=\qualitWidth]{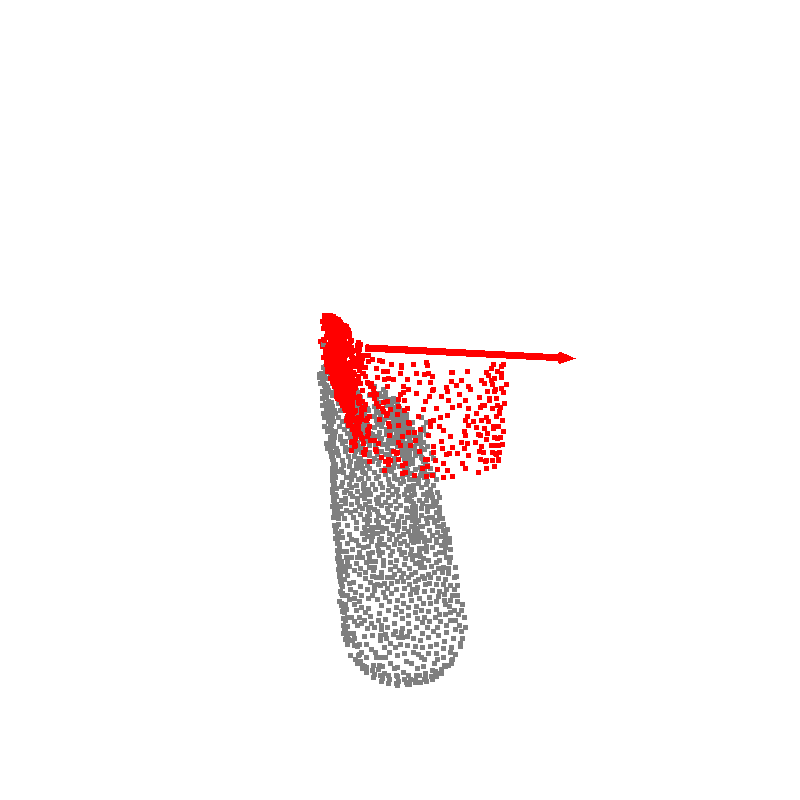} &
\includegraphics[trim={3cm 5.5cm 3cm 5.5cm},clip,width=\qualitWidth]{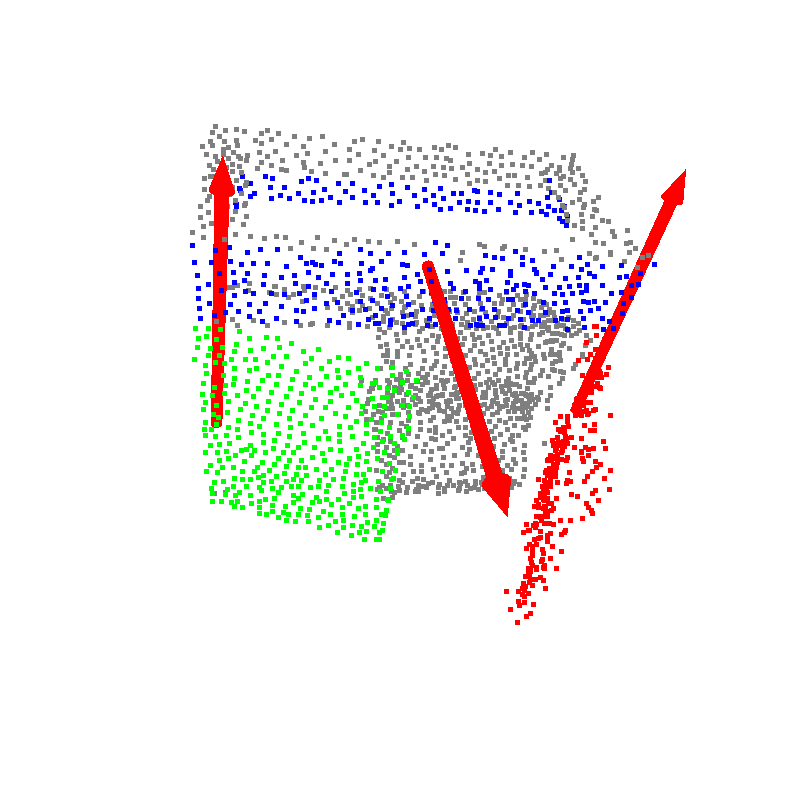} &
\includegraphics[trim={3cm 3cm 3cm 3cm},clip,width=\qualitWidth]{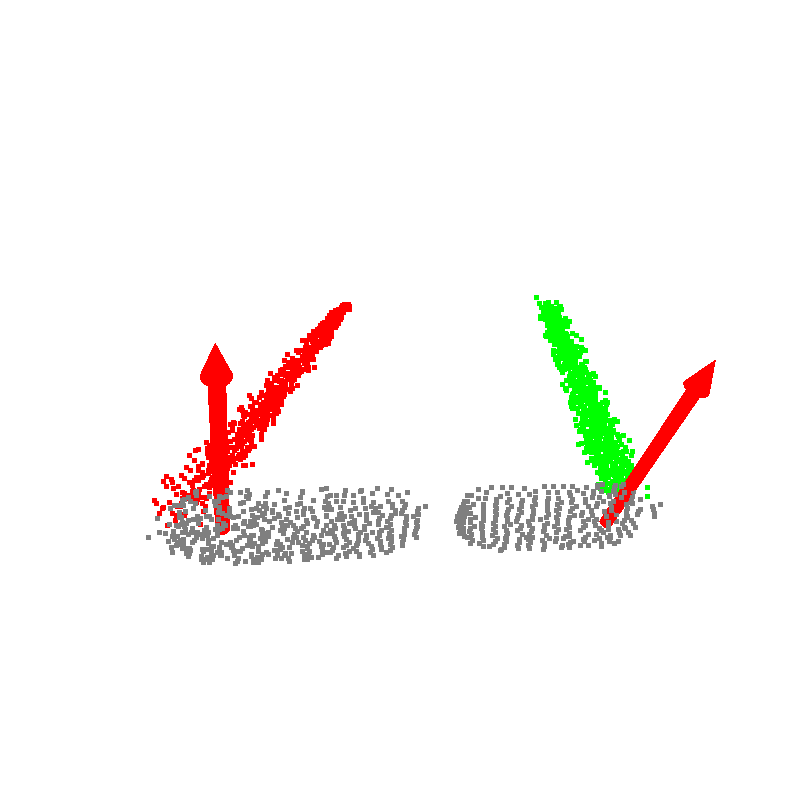} \\[-0.0cm]

\rotatebox{90}{\;\;\;\;\;\;Articulate-} &
\rotatebox{90}{\;\;\;\;\;\;Anything} &
\includegraphics[trim={4cm 5cm 4cm 5cm},clip,width=\qualitWidth]{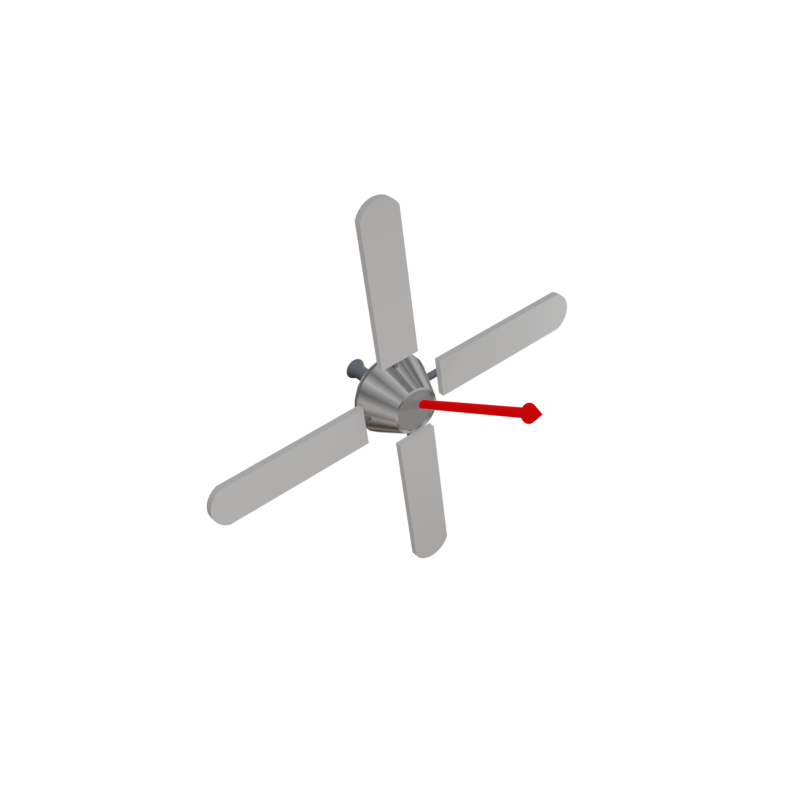} &
\includegraphics[trim={5cm 5cm 5cm 5cm},clip,width=\qualitWidth]{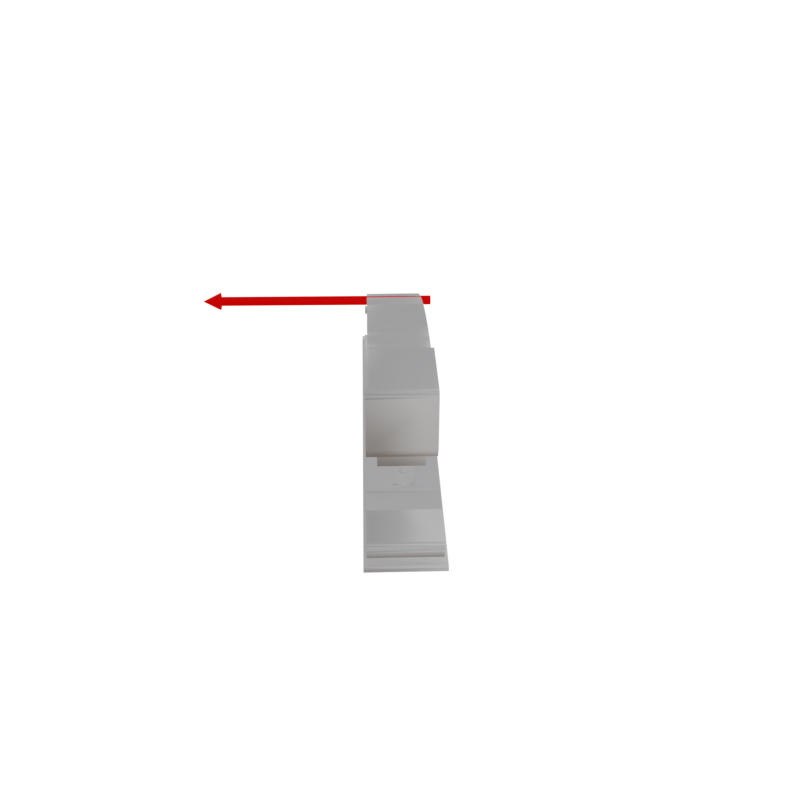} &
\epicfail & \epicfail \\ [-0.5cm]

&
\rotatebox{90}{\hspace{1.0cm}\vphantom{A}Reart} &
\includegraphics[trim={3cm 3cm 3cm 3cm},clip,width=\qualitWidth]{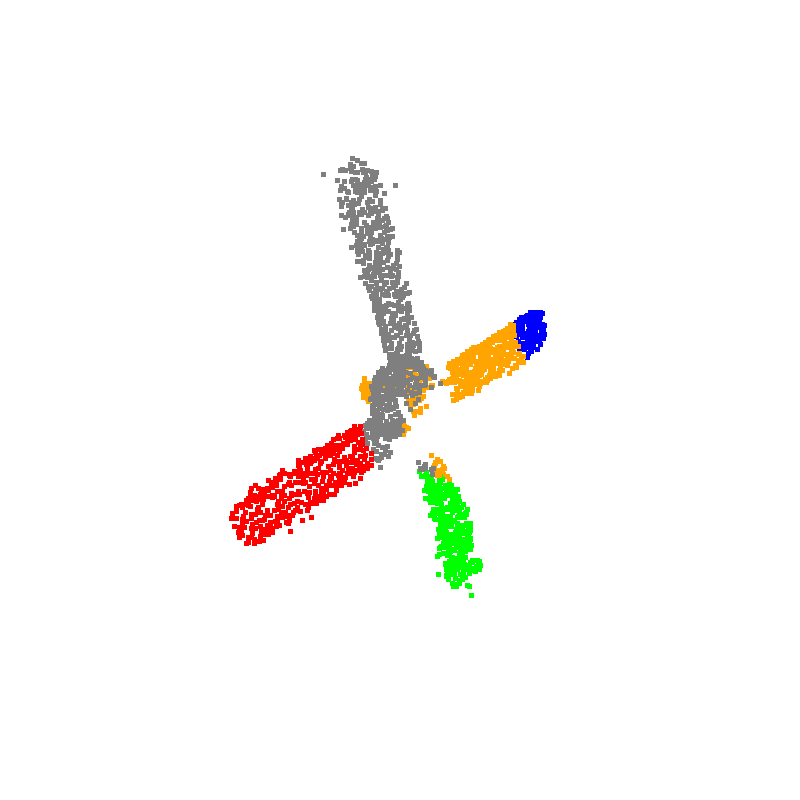}  &
\includegraphics[trim={5cm 3cm 5cm 3cm},clip,width=\qualitWidth]{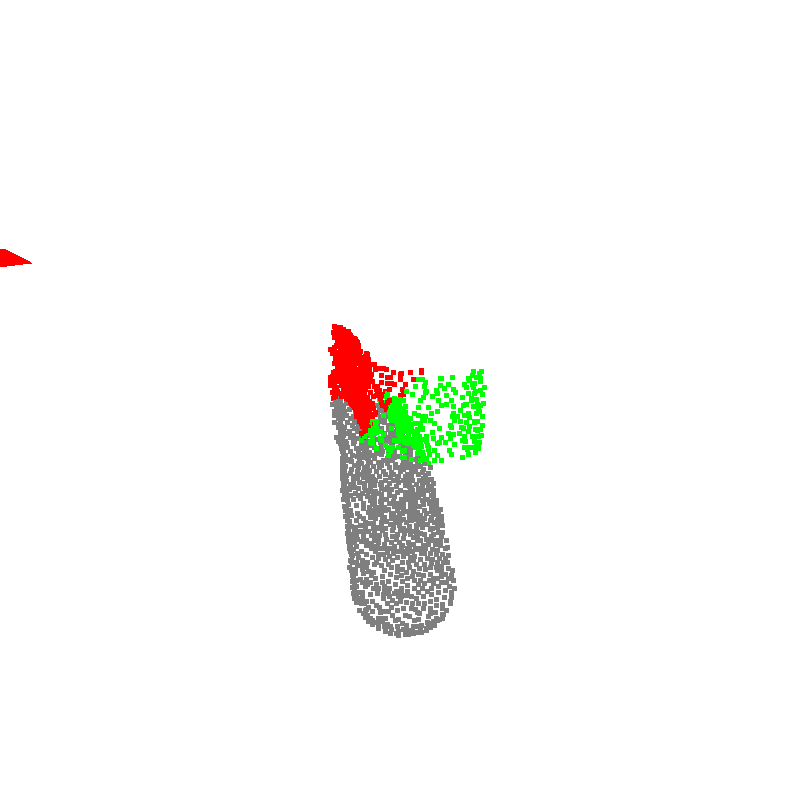} &
\includegraphics[trim={3cm 3cm 3cm 3cm},clip,width=\qualitWidth]{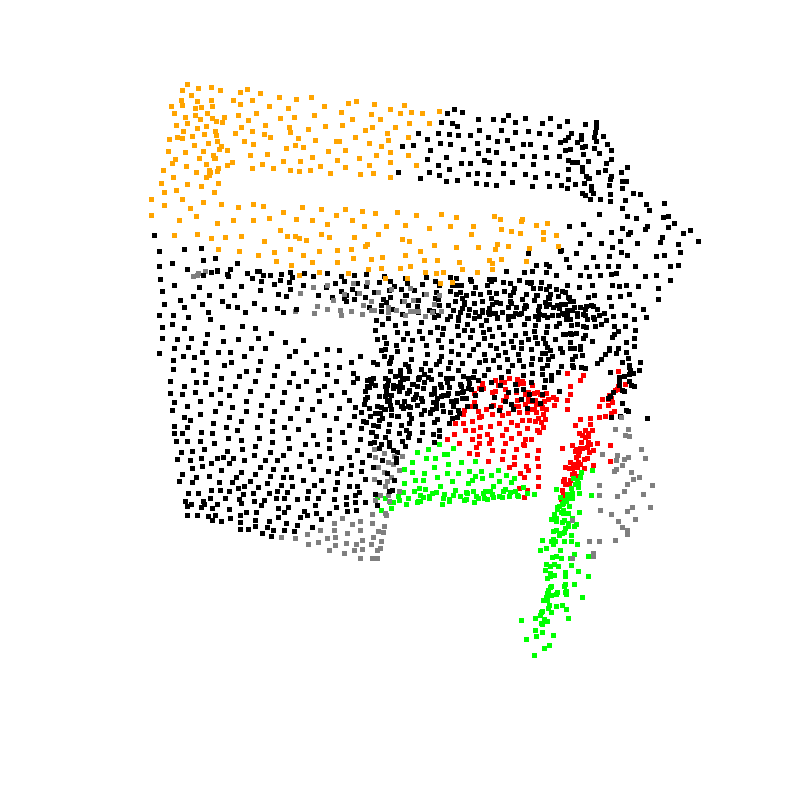} &
\includegraphics[trim={6cm 7cm 6cm 10cm},clip,width=\qualitWidth]{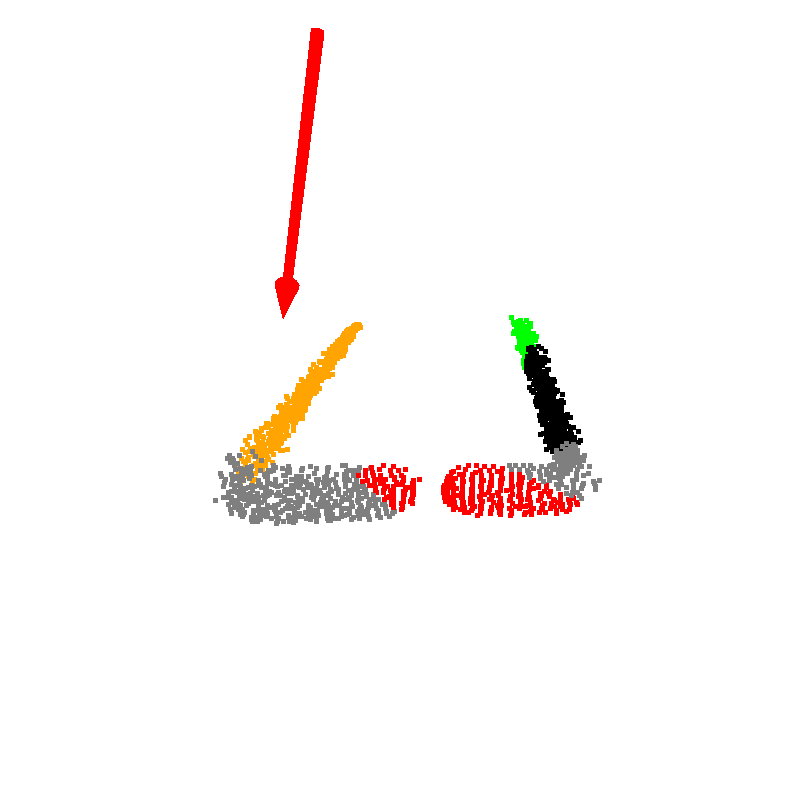}  \\ [-0.5cm]

\rotatebox{90}{\hspace{0.6cm}Video2} &
\rotatebox{90}{\hspace{0.4cm}Articulation} &
\includegraphics[trim={5cm 3cm 5cm 3cm},clip,width=\qualitWidth]{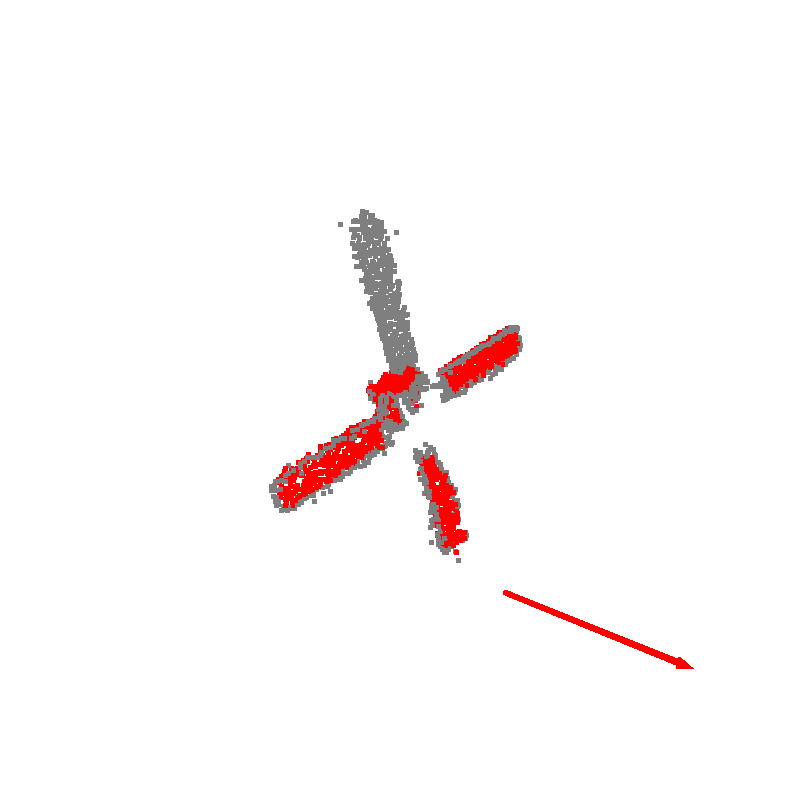}  &
\includegraphics[trim={5cm 3cm 5cm 3cm},clip,width=\qualitWidth]{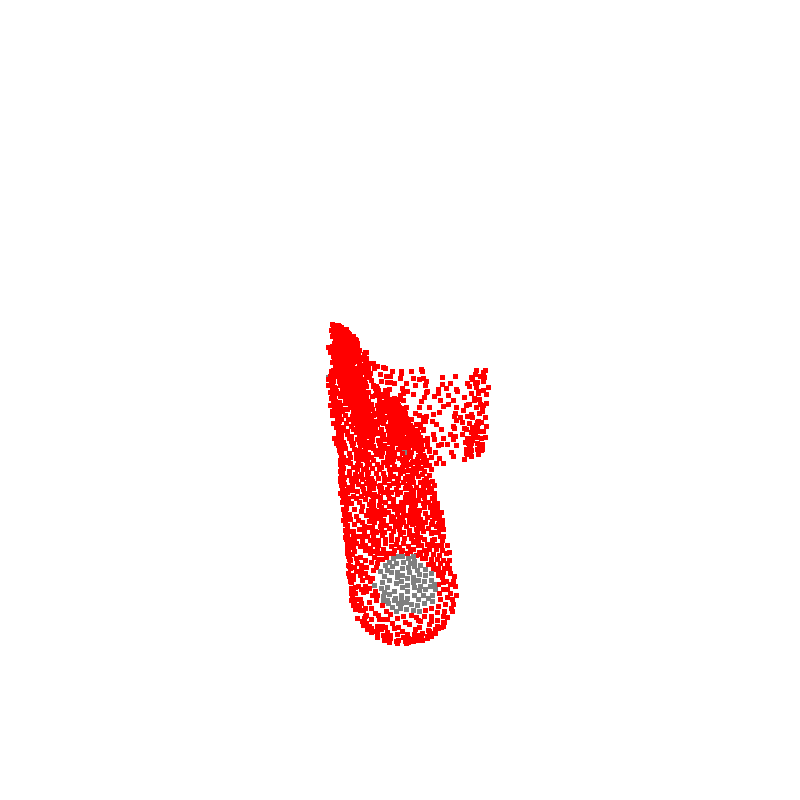} &
\epicfail & \epicfail \\[-0.5cm]

&
\rotatebox{90}{\hspace{0.2cm}\vphantom{A} FeatClust} &
\includegraphics[trim={3cm 3cm 3cm 3cm},clip,width=\qualitWidth]{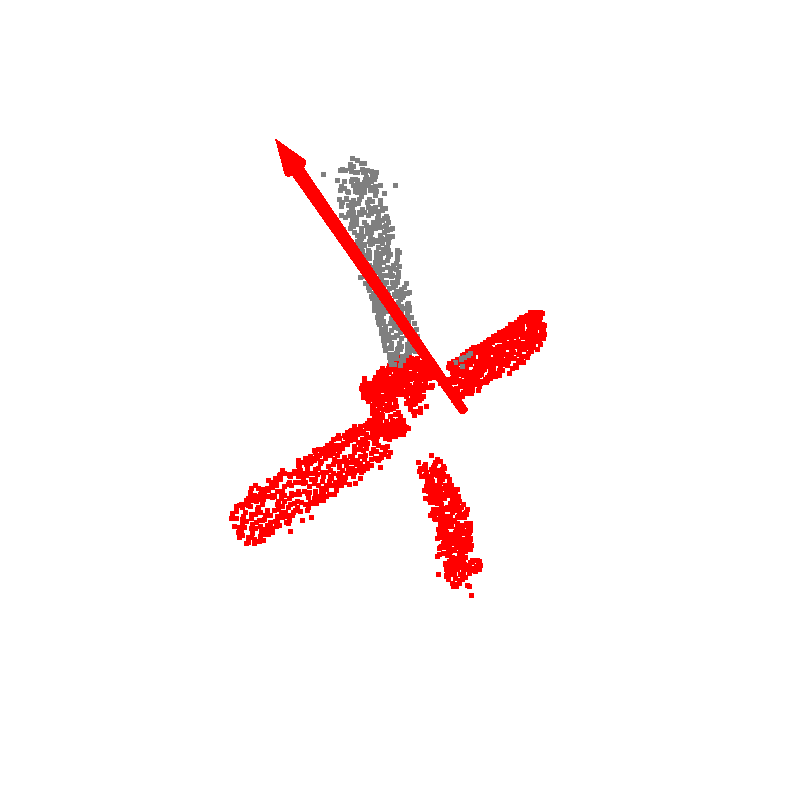}  &
\includegraphics[trim={3cm 3cm 3cm 3cm},clip,width=\qualitWidth]{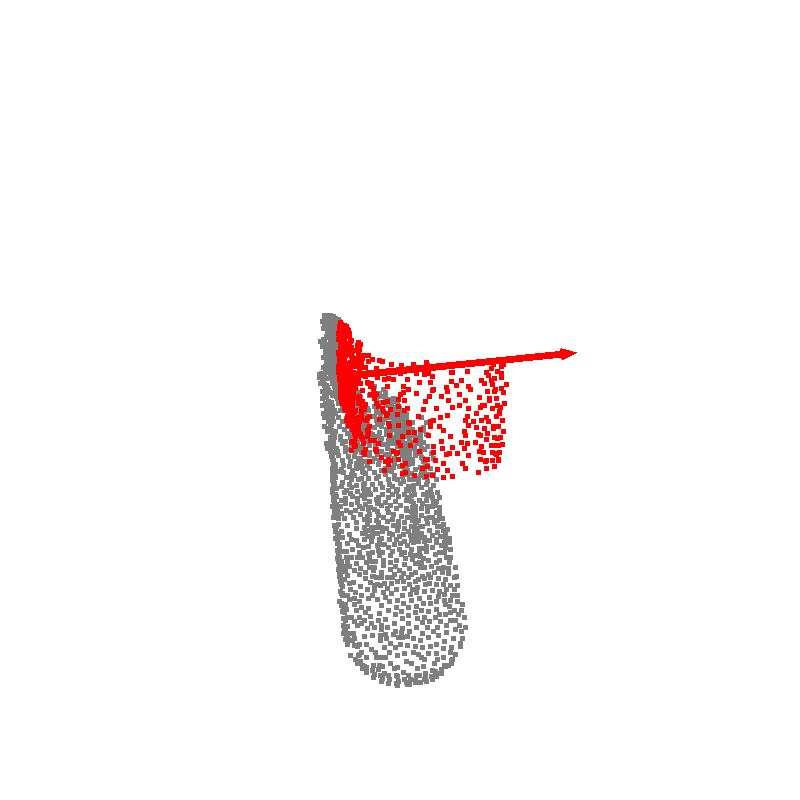} &
\includegraphics[trim={3cm 7cm 3cm 7cm},clip,width=\qualitWidth]{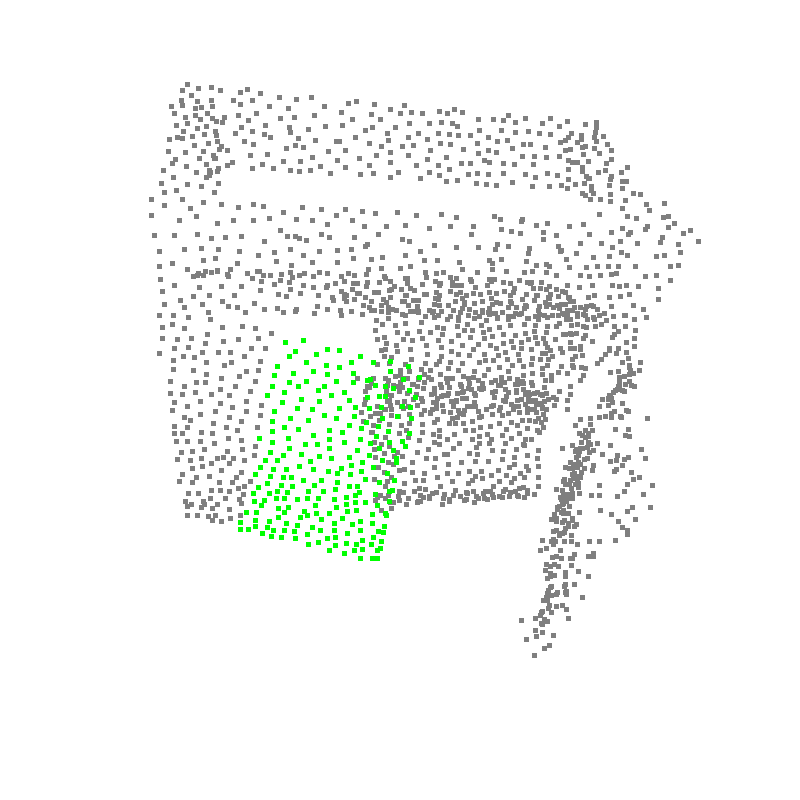} &
\includegraphics[trim={4cm 3cm 4cm 3cm},clip,width=\qualitWidth]{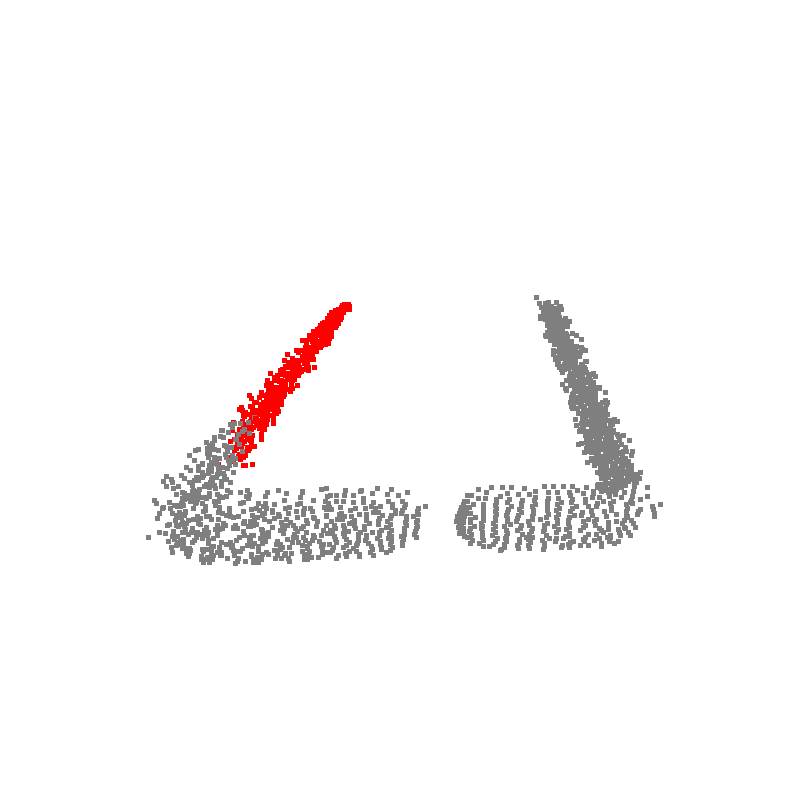}  \\

& & Fan & Stapler & Storage4 & Eyeglasses \\

\end{tabular}
} 
   \caption{\textbf{Qualitative results on randomly selected sequences from 4art-synth.} Red arrows denote predicted joint axes. Note how our method retrieves the part segmentations and the rotation axes much more accurately and robustly than all the other methods. '$\times$' indicates that Articulate-Anything and Video2Articulation are not designed for multi-part objects.  }
   \label{fig:qual_fig_merged}
\end{figure*}

\def\realdatawidth{0.8\linewidth}

\begin{figure*}[t] 
  \centering
  \resizebox{0.85\linewidth}{!}{
    \begin{tblr}{
colspec = {c@{$\;$}cX[c,m] X[c,m] X[c,m] X[c,m] X[c,m]}, 
colsep = 0cm,
rowsep = -1cm
  }
    \rotatebox{90}{Representative} &
    \rotatebox{90}{\hspace{1cm}Frame} &
    \includegraphics[width=\realdatawidth, height=0.8\linewidth]{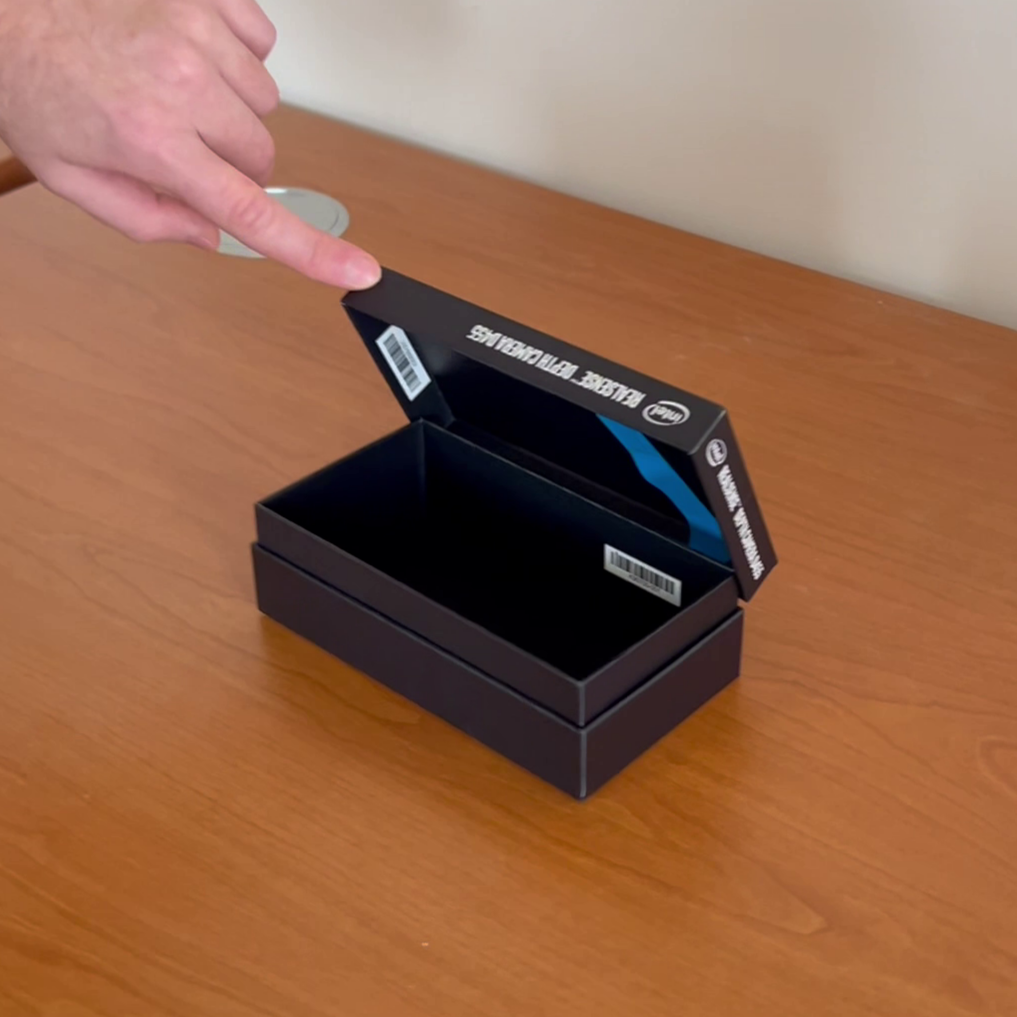} &
    \includegraphics[width=\realdatawidth, height=0.8\linewidth]{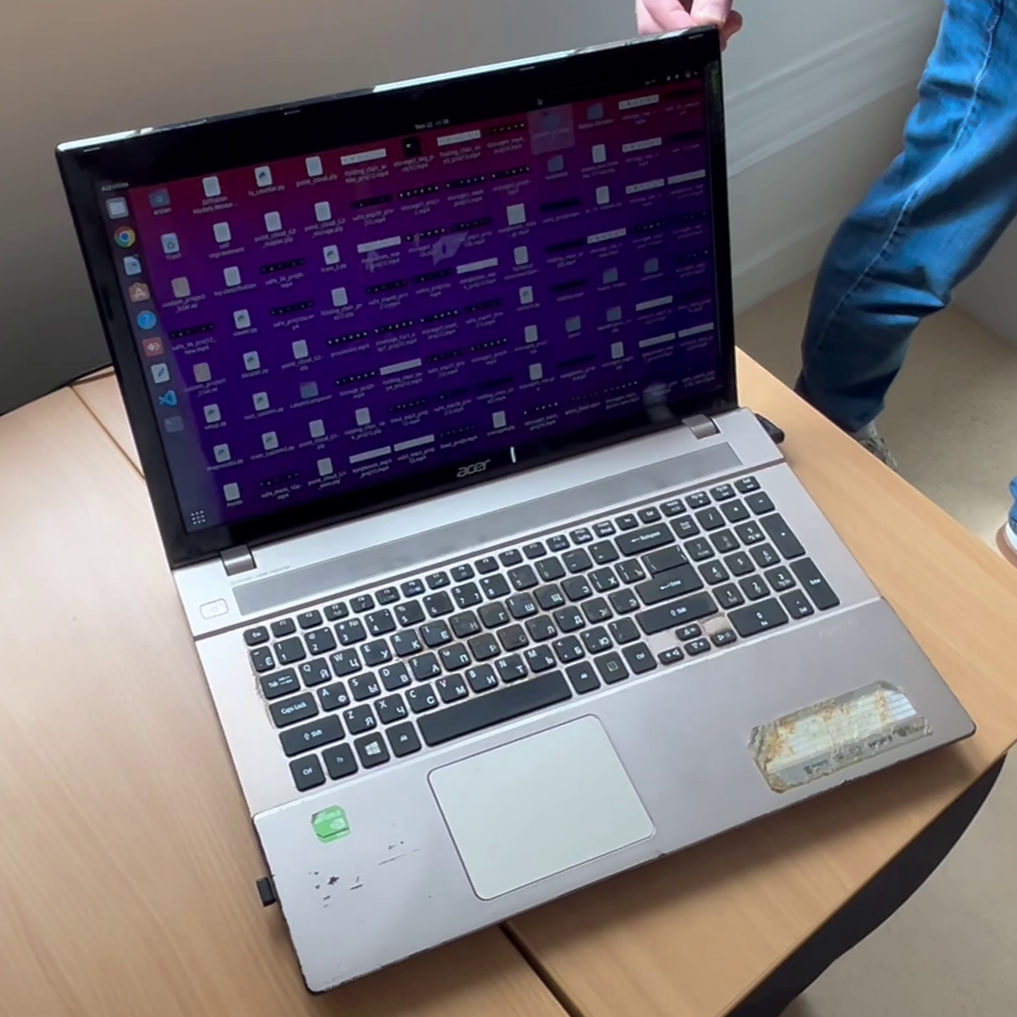} &
    \includegraphics[width=\realdatawidth, height=0.8\linewidth]{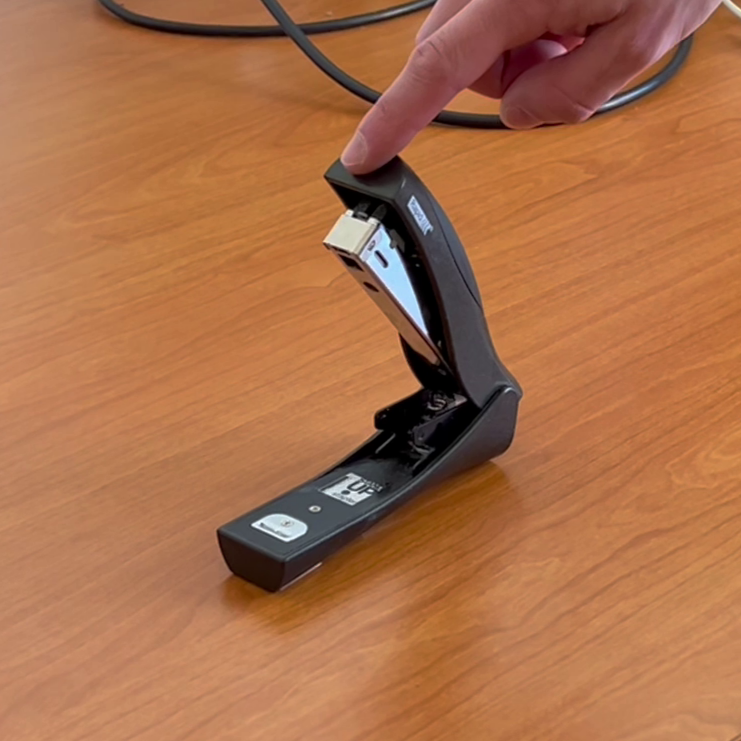} &
    \includegraphics[width=\realdatawidth, height=0.8\linewidth]{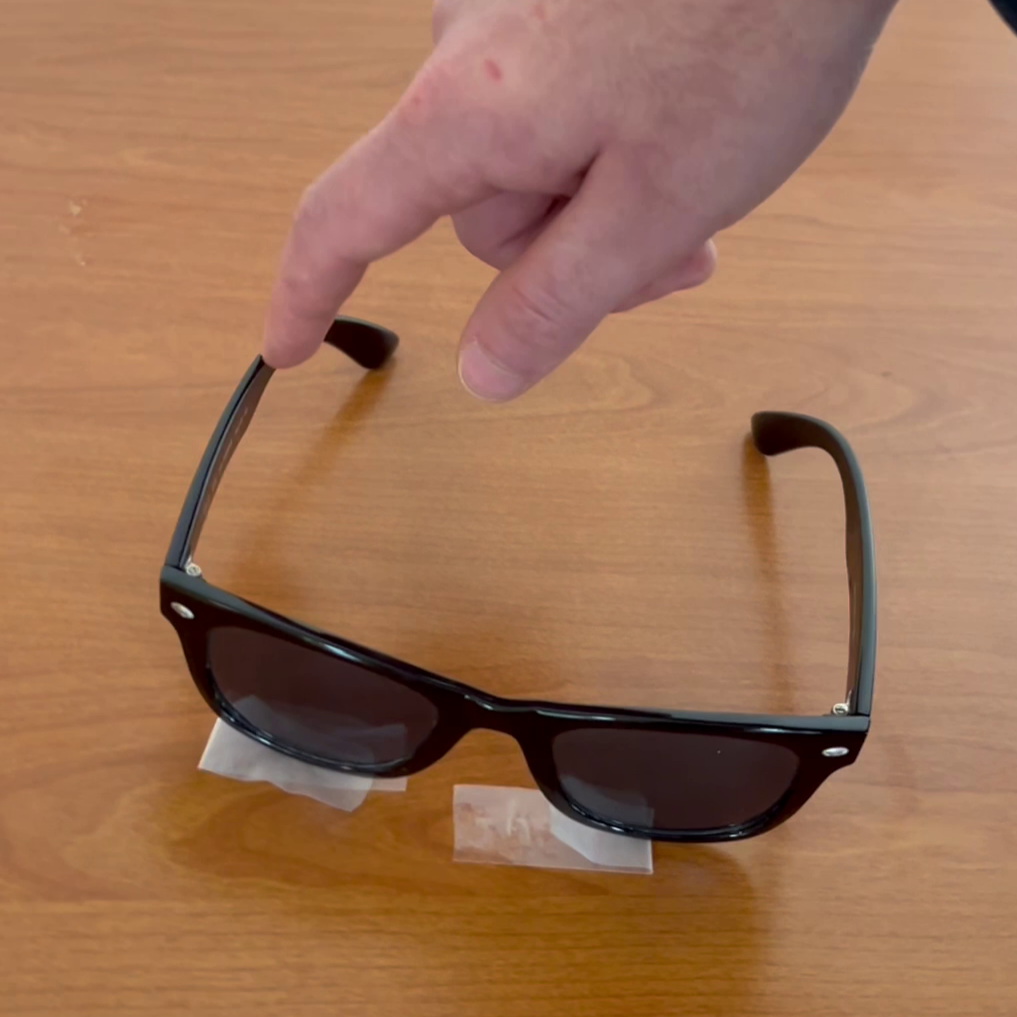} &
    \includegraphics[width=\realdatawidth, height=0.8\linewidth]{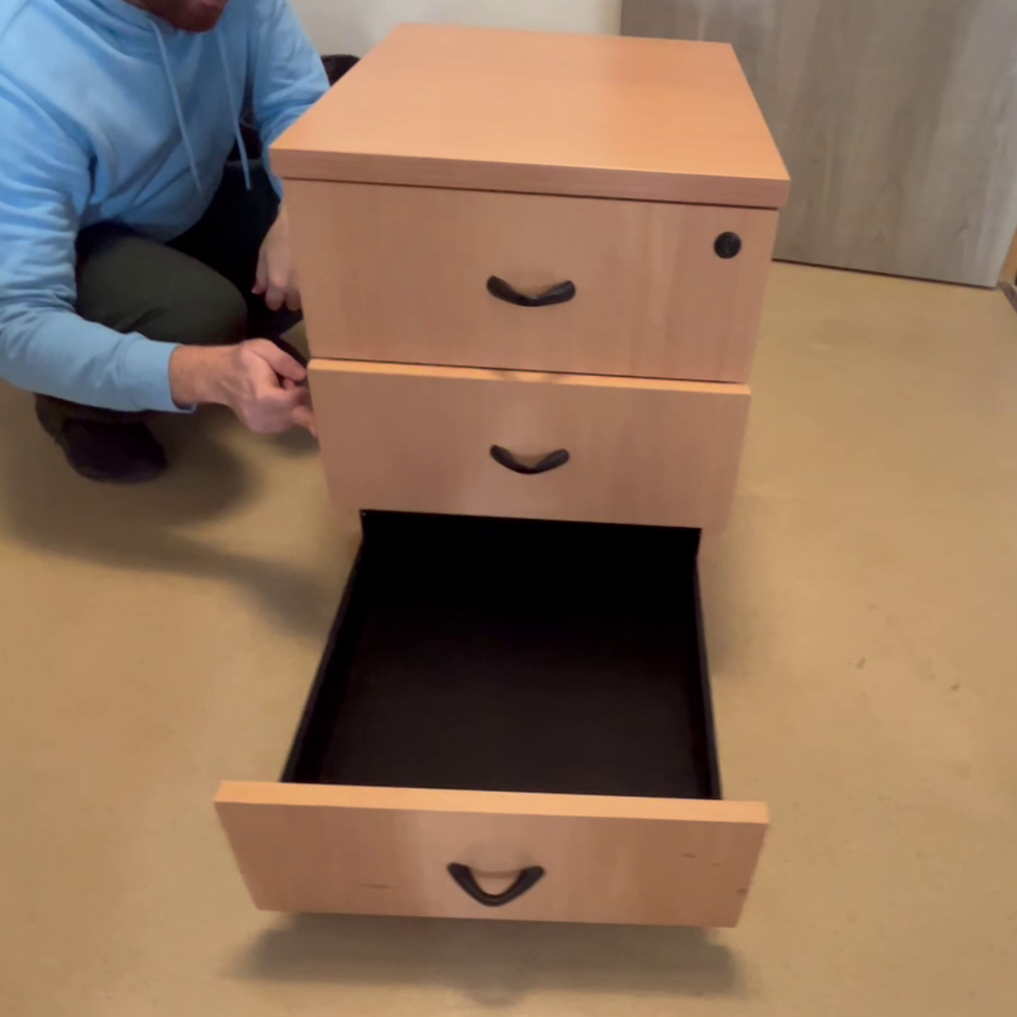} \\
    
    \rotatebox{90}{Ground} &
    \rotatebox{90}{$\;$Truth} &
    \includegraphics[trim={3cm 4cm 3cm 4cm},clip,width=\realdatawidth]{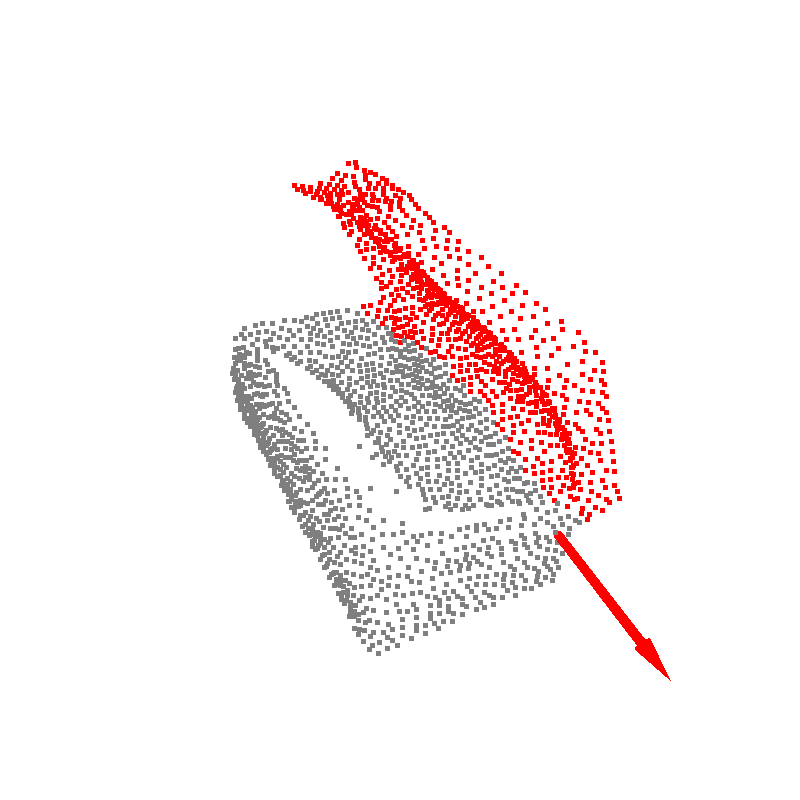} &
    \includegraphics[trim={3cm 5cm 3cm 6cm},clip,width=\realdatawidth]{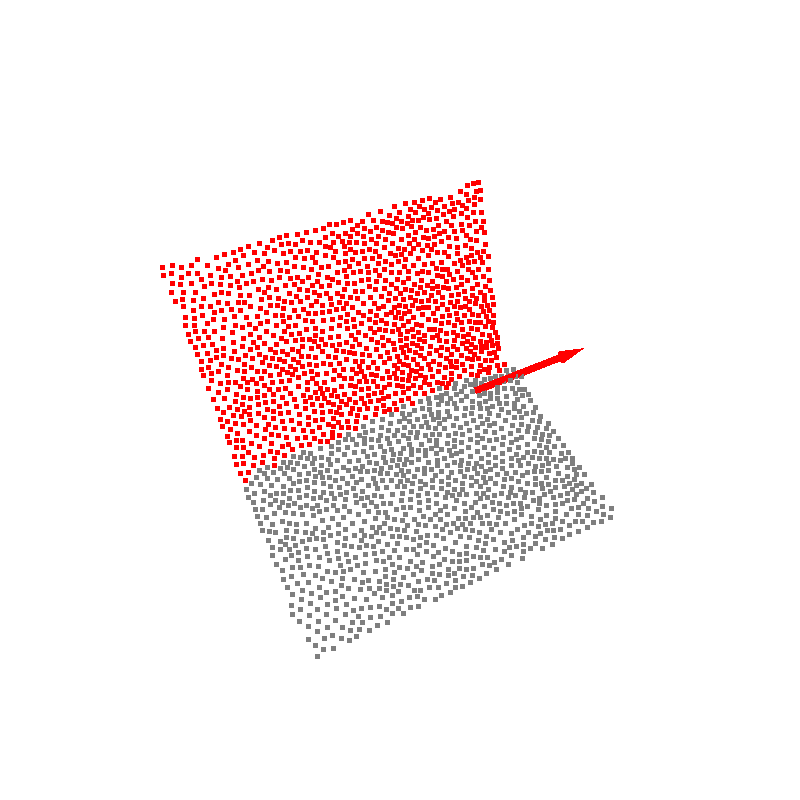} &
    \includegraphics[trim={0cm 5cm 0cm 5cm},clip,width=\realdatawidth]{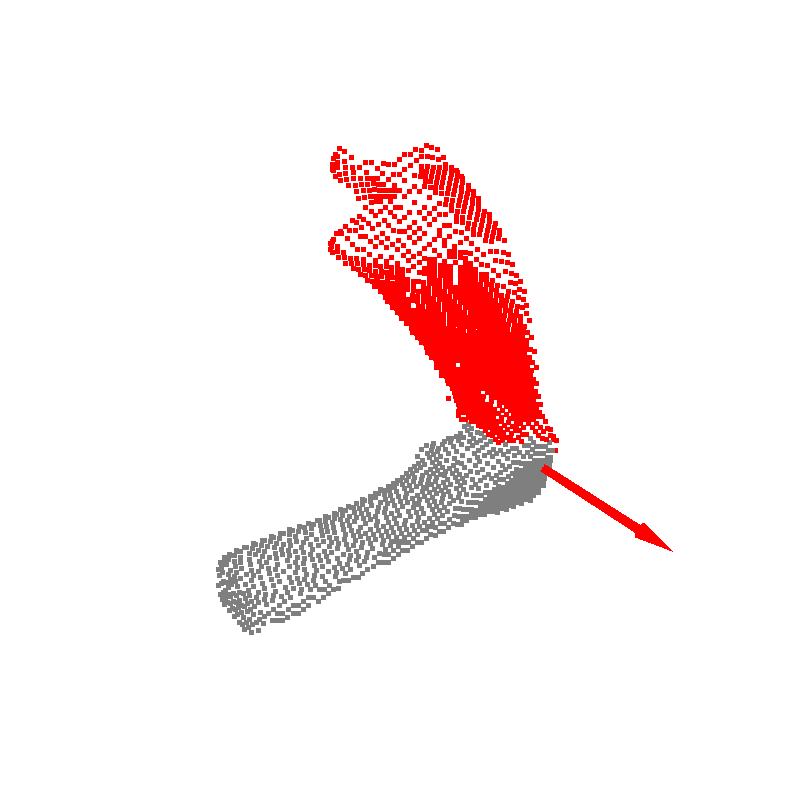} &
    \includegraphics[trim={3cm 6cm 3cm 6cm},clip,width=\realdatawidth]{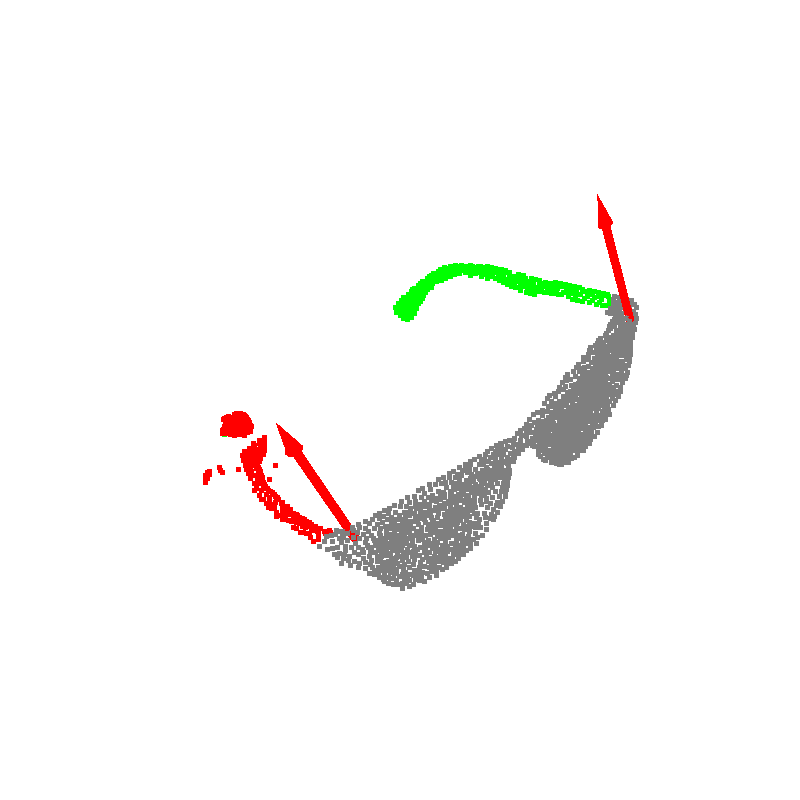} &
    \includegraphics[trim={3cm 5cm 3cm 6cm},clip,width=\realdatawidth]{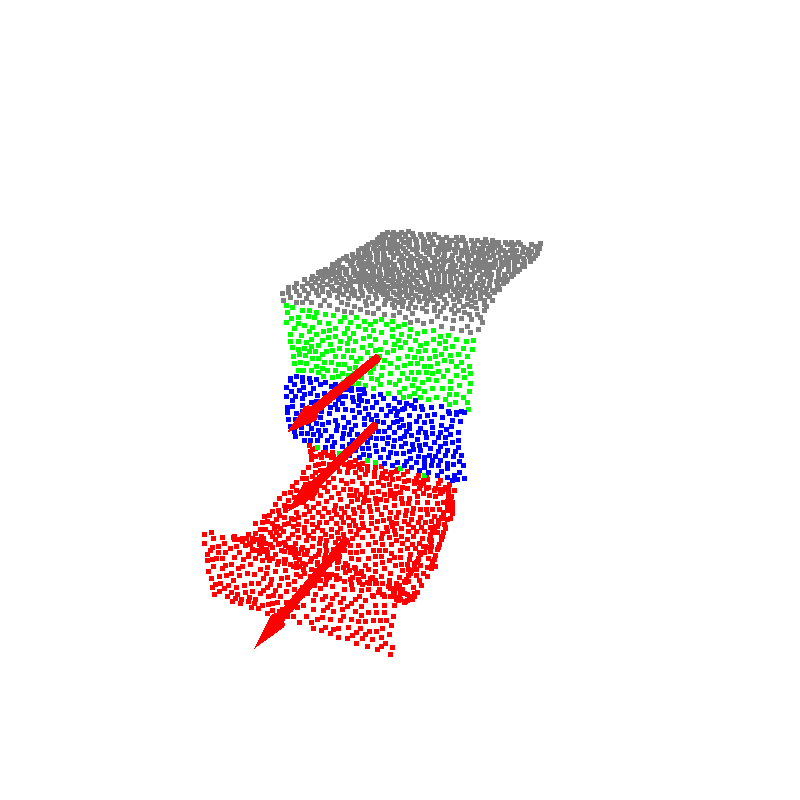} \\

    \rotatebox{90}{\hspace{0.2cm}sim2art} &
    \rotatebox{90}{\hspace{0.3cm}(ours)} &
    \includegraphics[trim={3cm 4cm 3cm 4cm},clip,width=\realdatawidth]{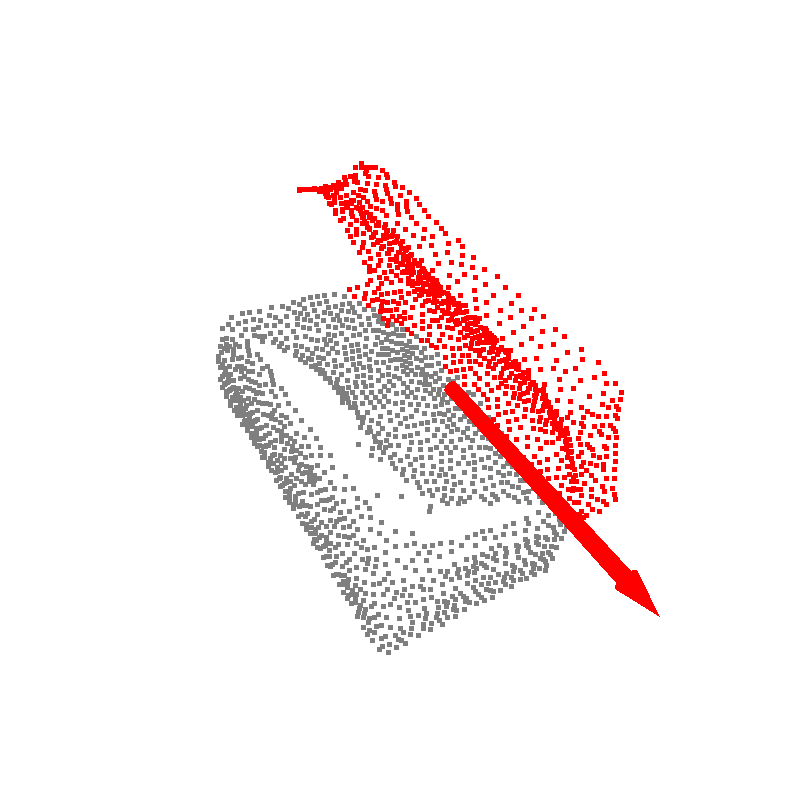} &
    \includegraphics[trim={3cm 5cm 3cm 5cm},clip,width=\realdatawidth]{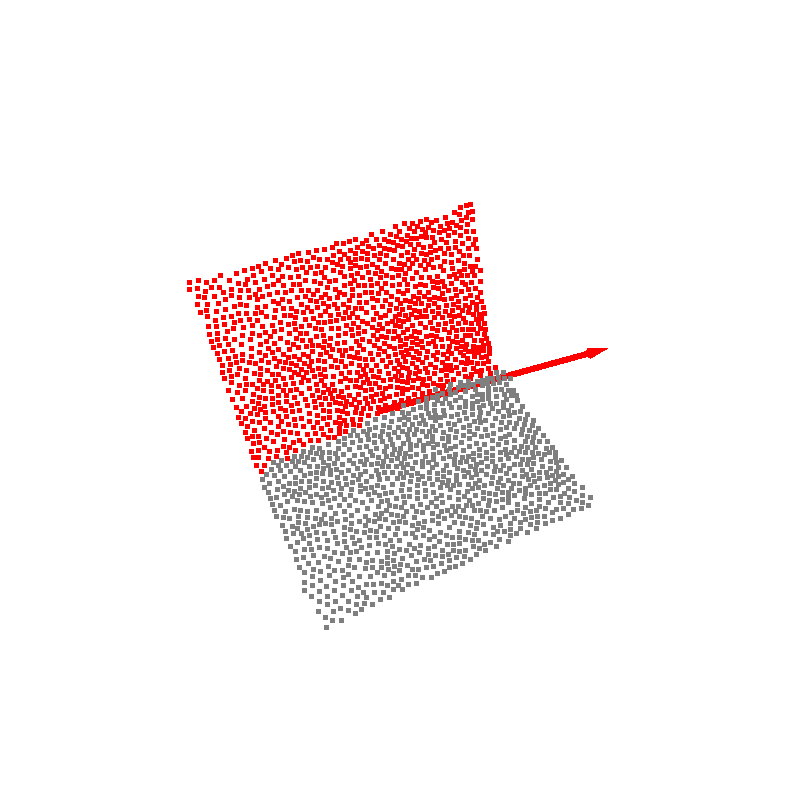} &
    \includegraphics[trim={0cm 6cm 0cm 5cm},clip,width=\realdatawidth]{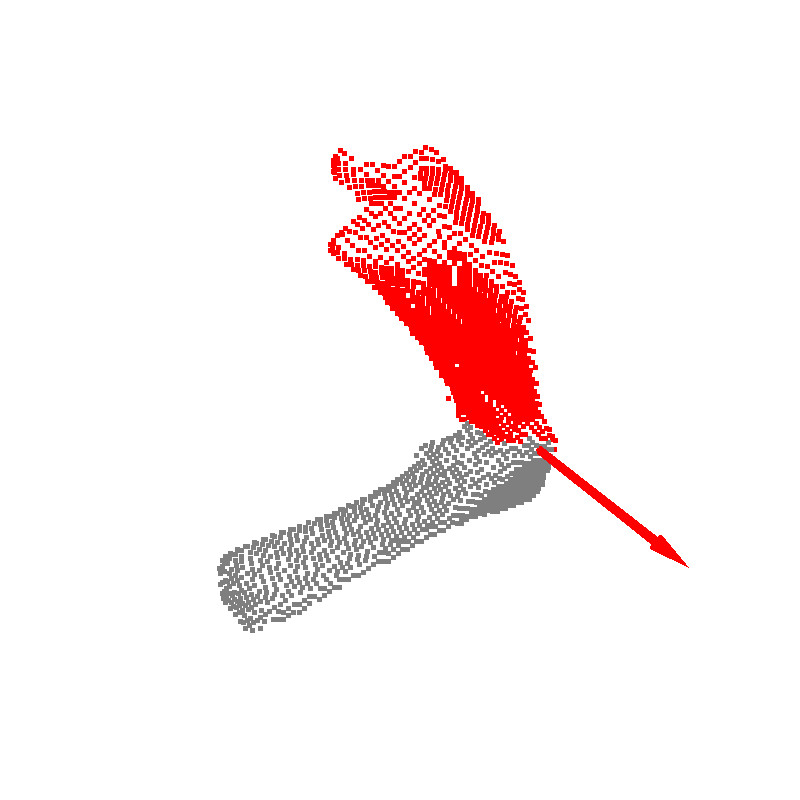} &
    \includegraphics[trim={3cm 7.3cm 3cm 6cm},clip,width=\realdatawidth]{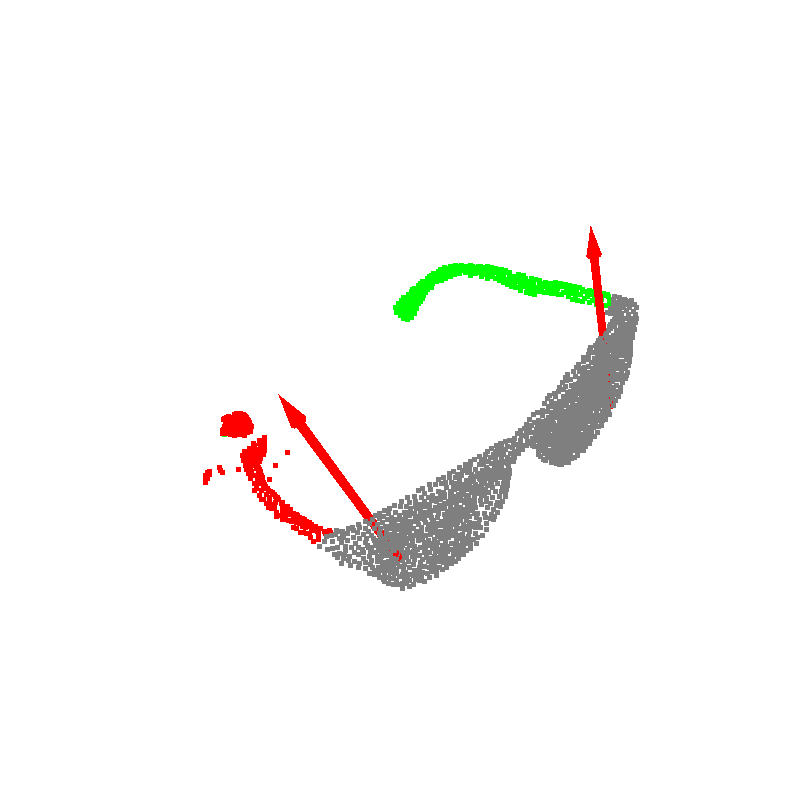} &
    \includegraphics[trim={3cm 5cm 3cm 6cm},clip,width=\realdatawidth]{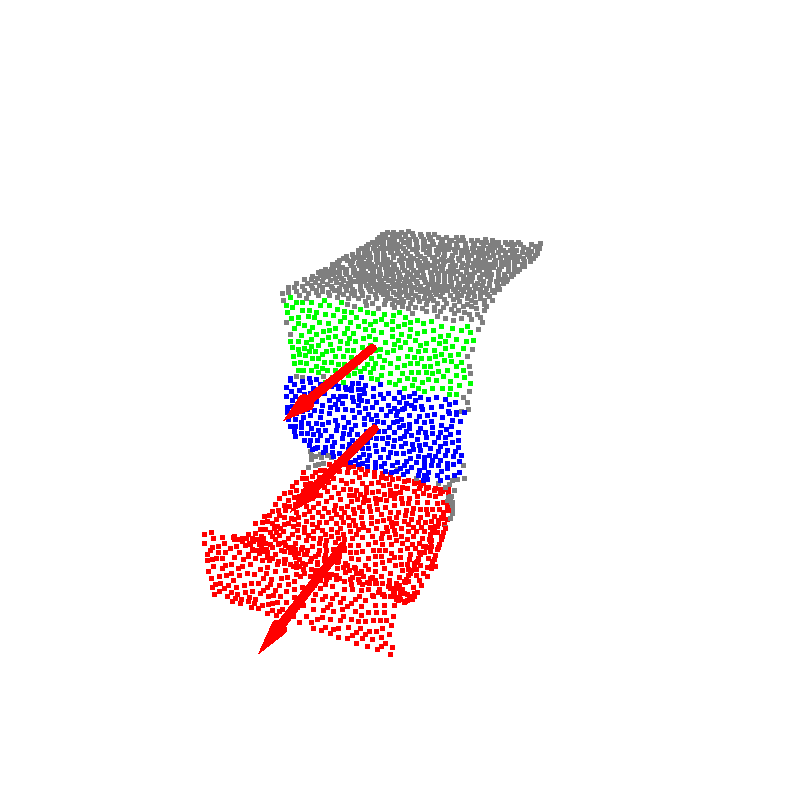}\\
    
    \rotatebox{90}{$\;\;\;\;\;\;$Articulate-} &
    \rotatebox{90}{$\;\;\;\;\;\;\;\;$Anything} &
    \includegraphics[trim={5cm 3cm 6cm 0cm},clip,width=\realdatawidth]{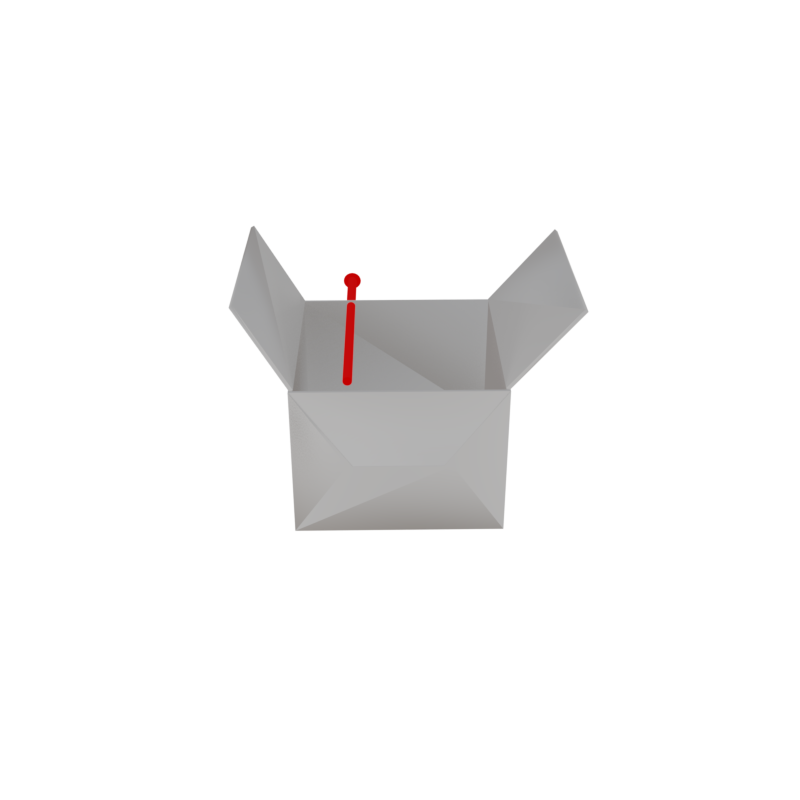} &
    \includegraphics[trim={3cm 2cm 3cm 0cm},clip,width=\realdatawidth]{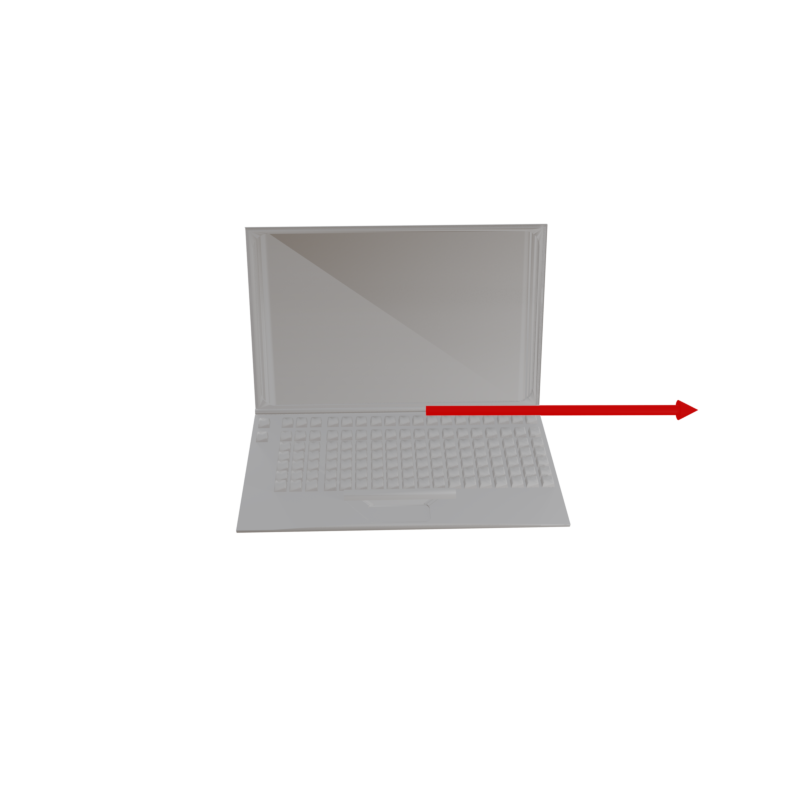} &
    \includegraphics[trim={3cm 2cm 1cm 3cm},clip,width=\realdatawidth]{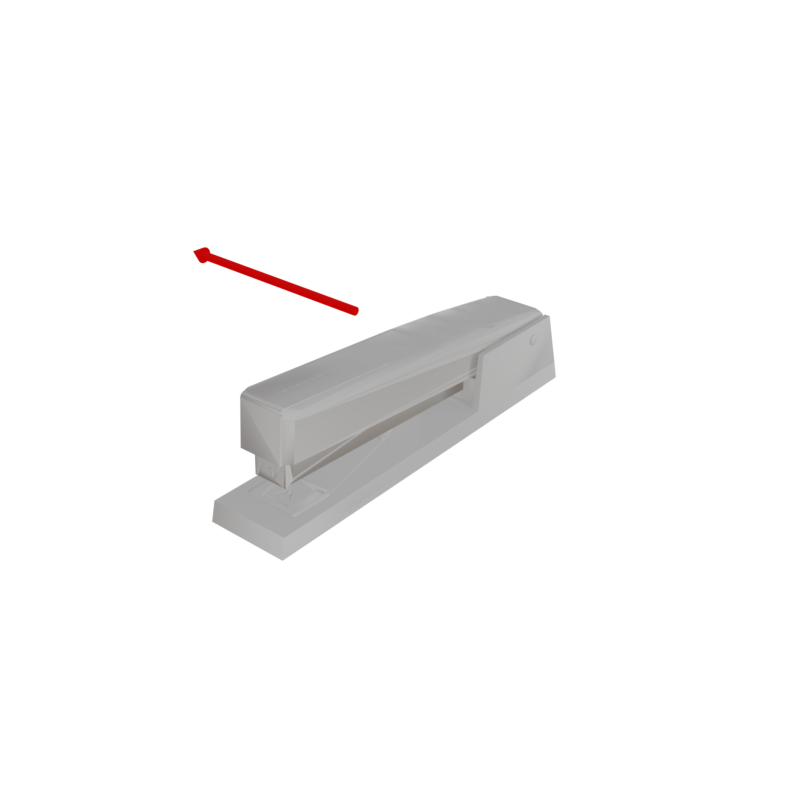} &
    \begin{tikzpicture}[x=\realdatawidth,y=\realdatawidth]
        \draw[gray, line width=0.1cm] (0.1,0.1) -- (0.9,0.9);
        \draw[gray, line width=0.1cm] (0.1,0.9) -- (0.9,0.1);
    \end{tikzpicture} &
    \begin{tikzpicture}[x=\realdatawidth,y=\realdatawidth]
        \draw[gray, line width=0.1cm] (0.1,0.1) -- (0.9,0.9);
        \draw[gray, line width=0.1cm] (0.1,0.9) -- (0.9,0.1);
    \end{tikzpicture} \\
    
    &
    \rotatebox{90}{\hspace{0.5cm}\vphantom{A} Reart} &
    \includegraphics[trim={6cm 5cm 3cm 6cm},clip,width=\realdatawidth]{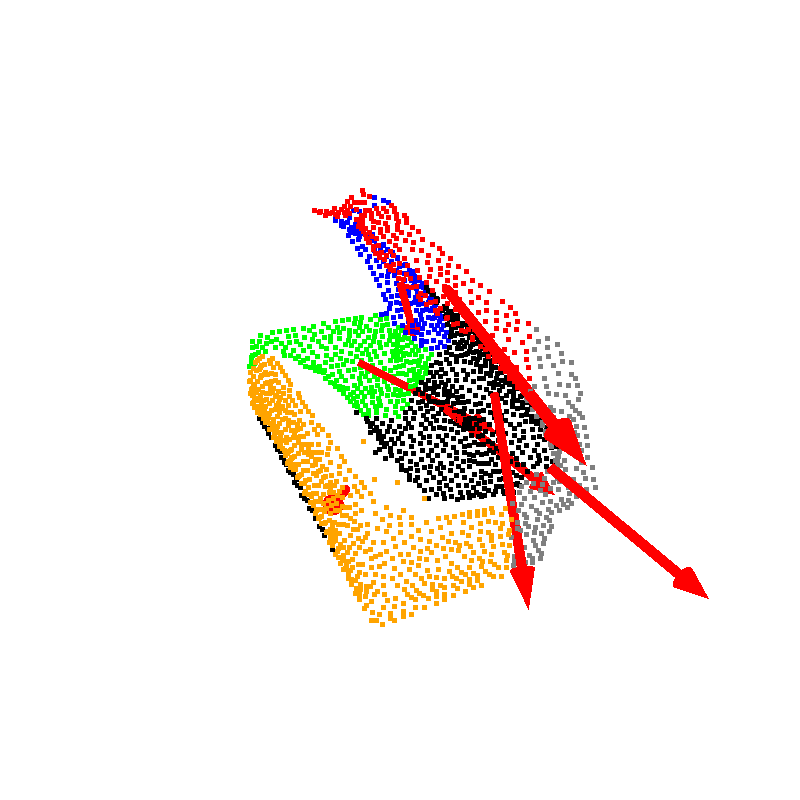} &
    \includegraphics[trim={3cm 4cm 3cm 4cm},clip,width=\realdatawidth]{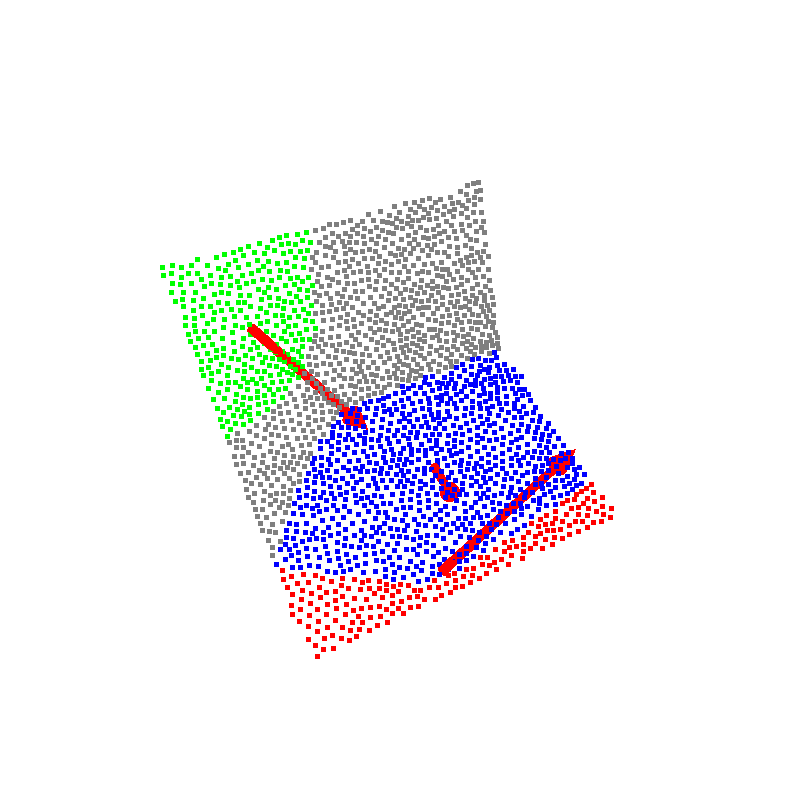} &
    \includegraphics[trim={5cm 6cm 5cm 5cm},clip,width=\realdatawidth]{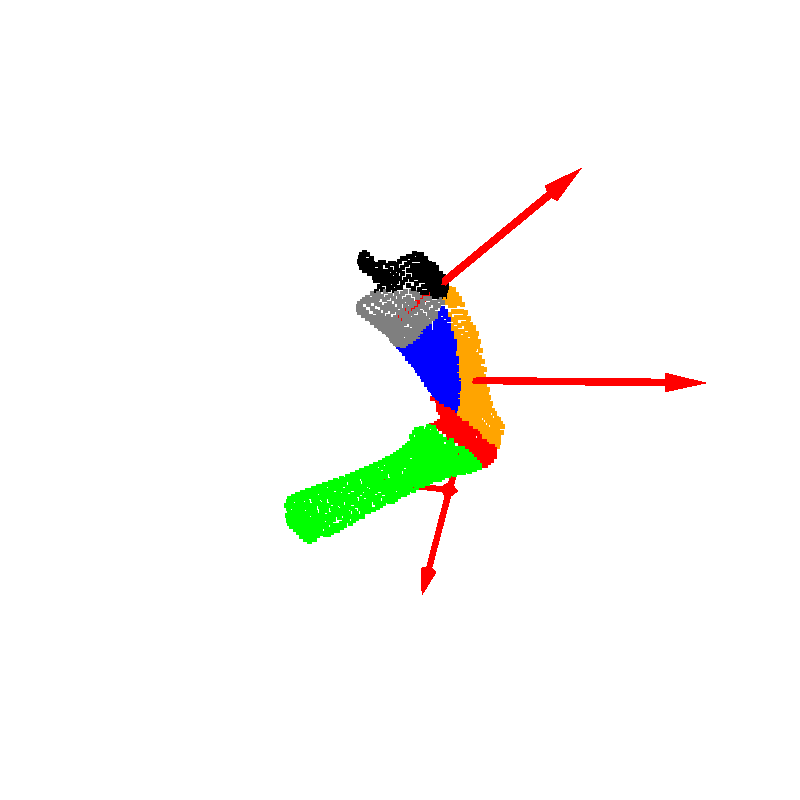} &
    \includegraphics[trim={5cm 7.3cm 5cm 6cm},clip,width=\realdatawidth]{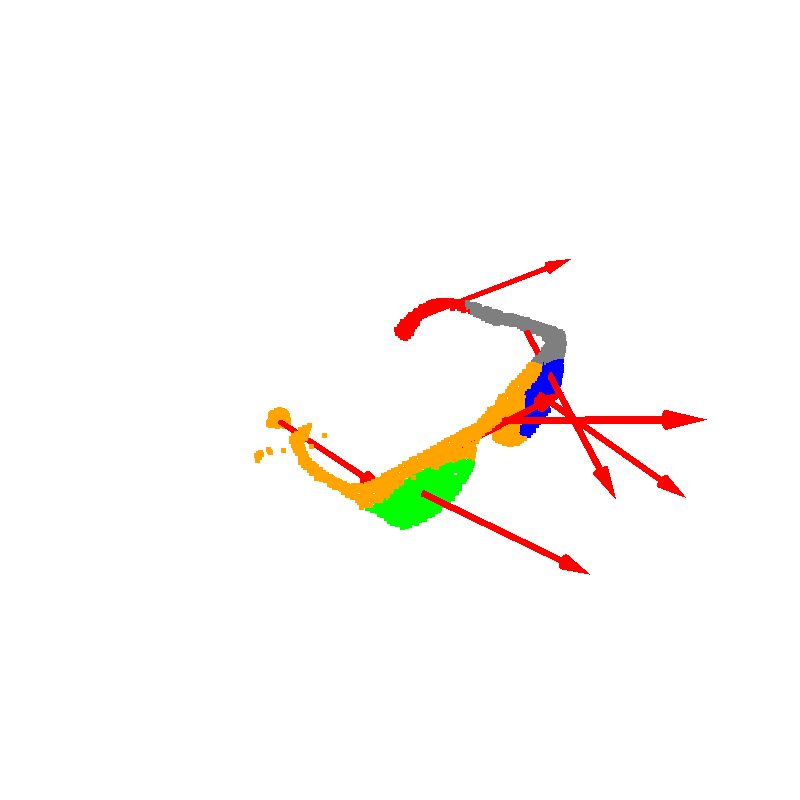} &
    \includegraphics[trim={5cm 5cm 5cm 6cm},clip,width=\realdatawidth]{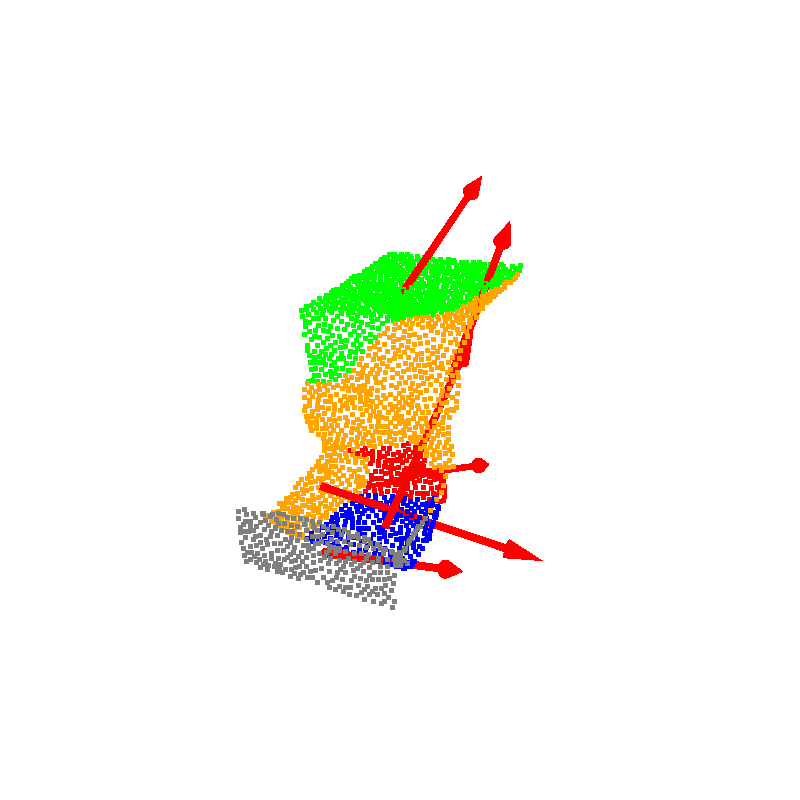} \\
    
    \rotatebox{90}{$\;\;\;\;\;\;\;\;\;\;\;\;$Video2} &
    \rotatebox{90}{$\;\;\;\;\;\;\;\;$Articulation} &
    \includegraphics[trim={3cm 0cm 3cm 0cm},clip,width=\realdatawidth]{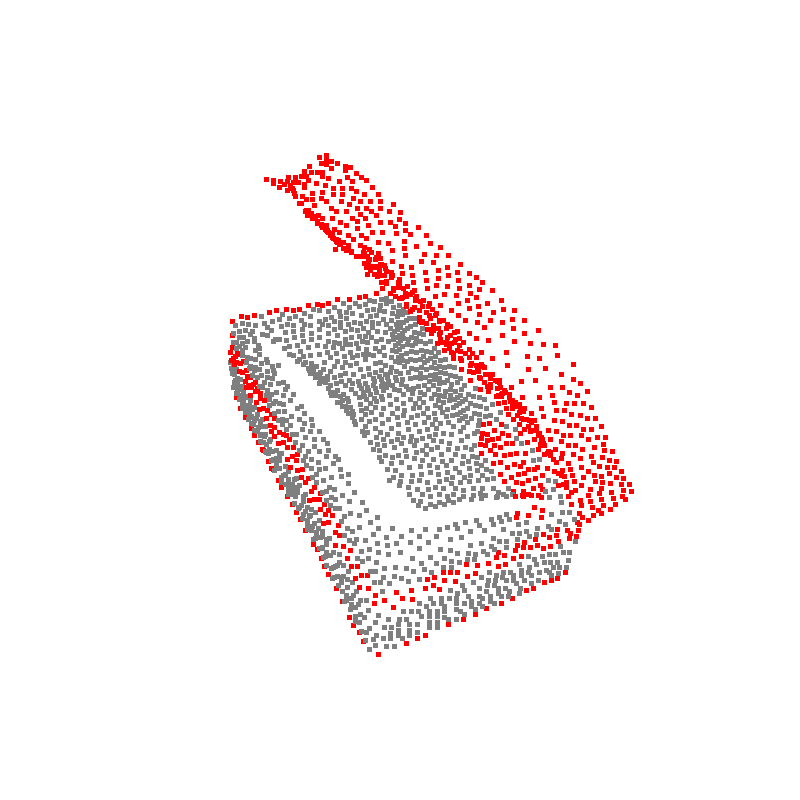} &
    \includegraphics[trim={5cm 5cm 5cm 5cm},clip,width=\realdatawidth]{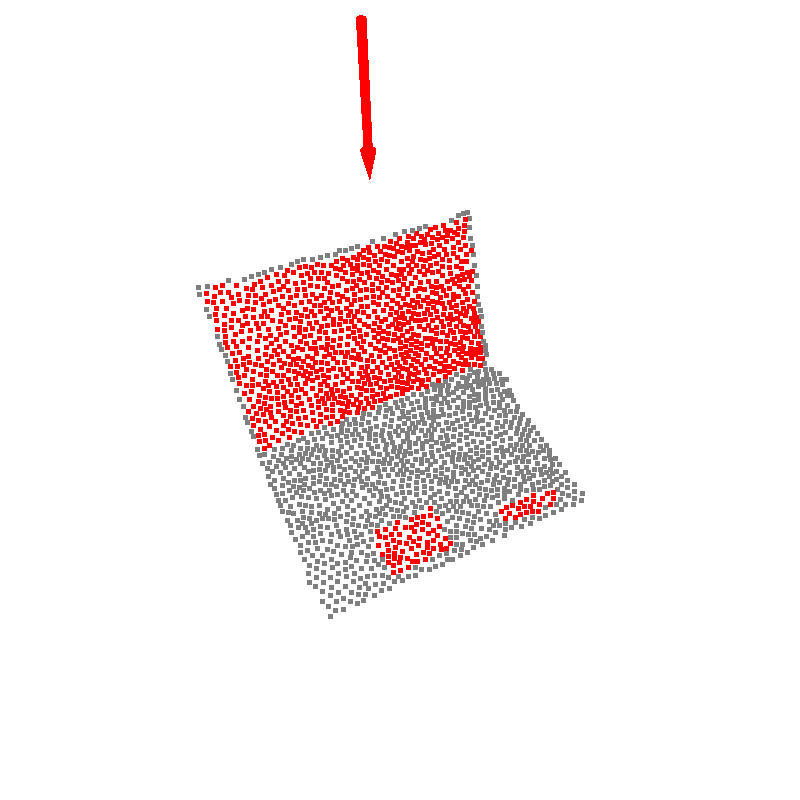} &
    \includegraphics[trim={0cm 3cm 0cm 3cm},clip,width=\realdatawidth]{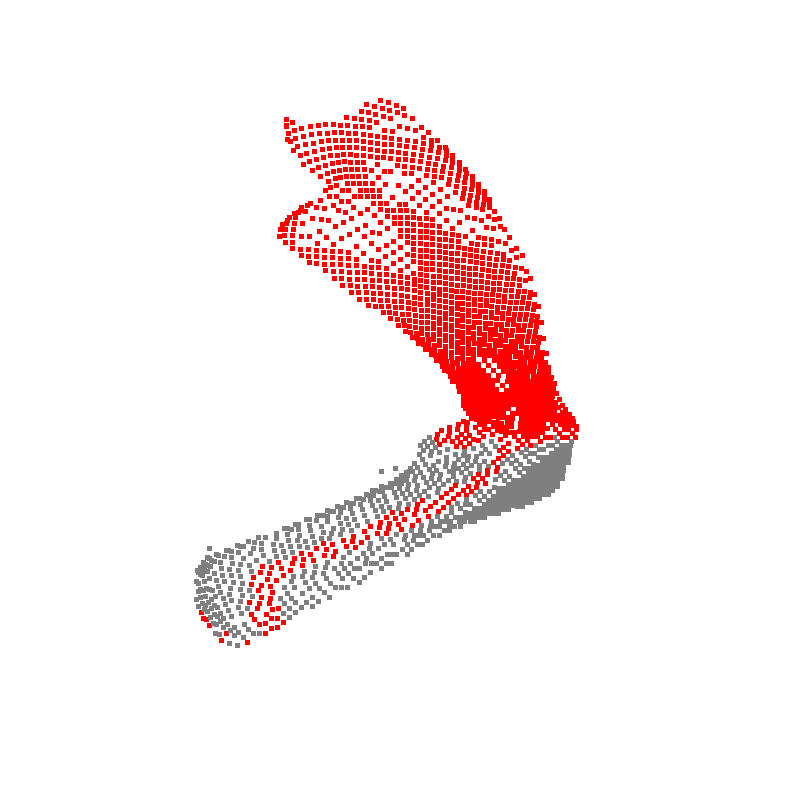} &
    \begin{tikzpicture}[x=\realdatawidth,y=\realdatawidth]
        \draw[gray, line width=0.1cm] (0.1,0.1) -- (0.9,0.9);
        \draw[gray, line width=0.1cm] (0.1,0.9) -- (0.9,0.1);
    \end{tikzpicture} &
    \begin{tikzpicture}[x=\realdatawidth,y=\realdatawidth]
        \draw[gray, line width=0.1cm] (0.1,0.1) -- (0.9,0.9);
        \draw[gray, line width=0.1cm] (0.1,0.9) -- (0.9,0.1);
    \end{tikzpicture} \\
    
    &
    \rotatebox{90}{\hspace{0.3cm}\vphantom{A} FeatClust} &
    \includegraphics[trim={6cm 5cm 0cm 5cm},clip,width=\realdatawidth]{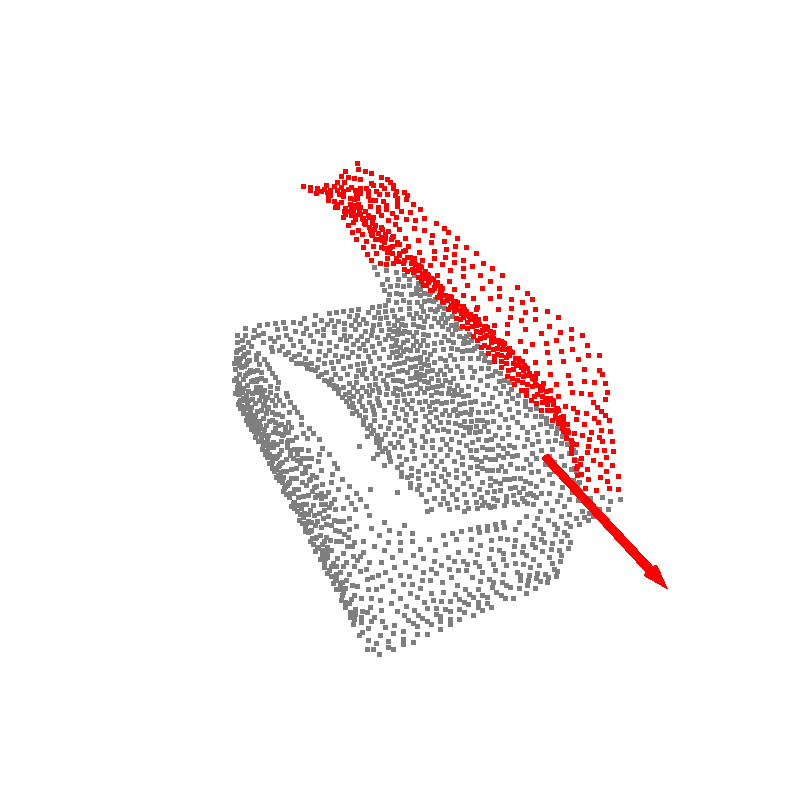} &
    \includegraphics[trim={5cm 6cm 5cm 5cm},clip,width=\realdatawidth]{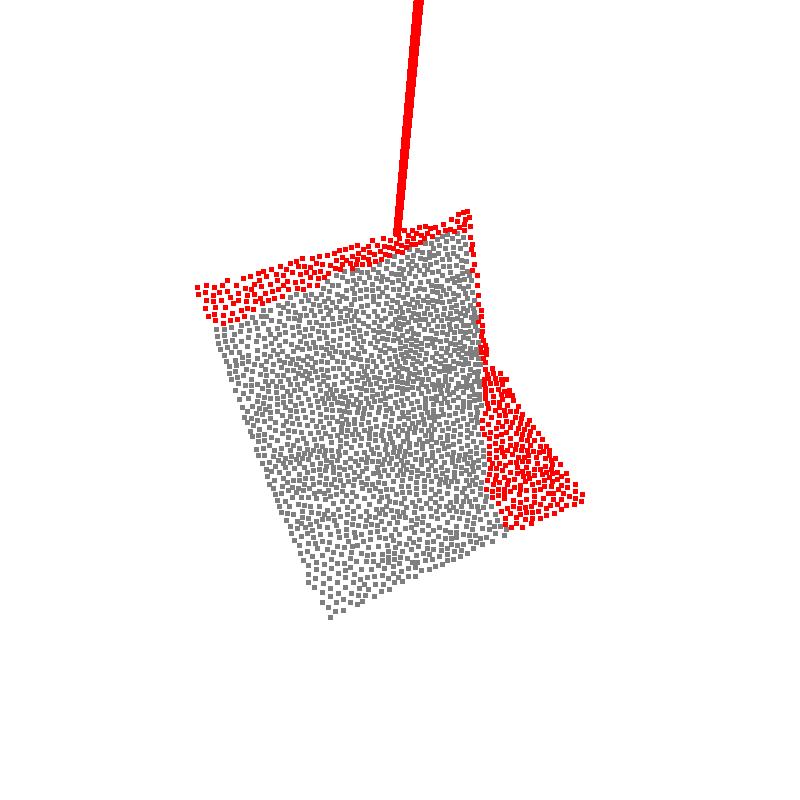} &
    \includegraphics[trim={2cm 4cm 2cm 5cm},clip,width=\realdatawidth]{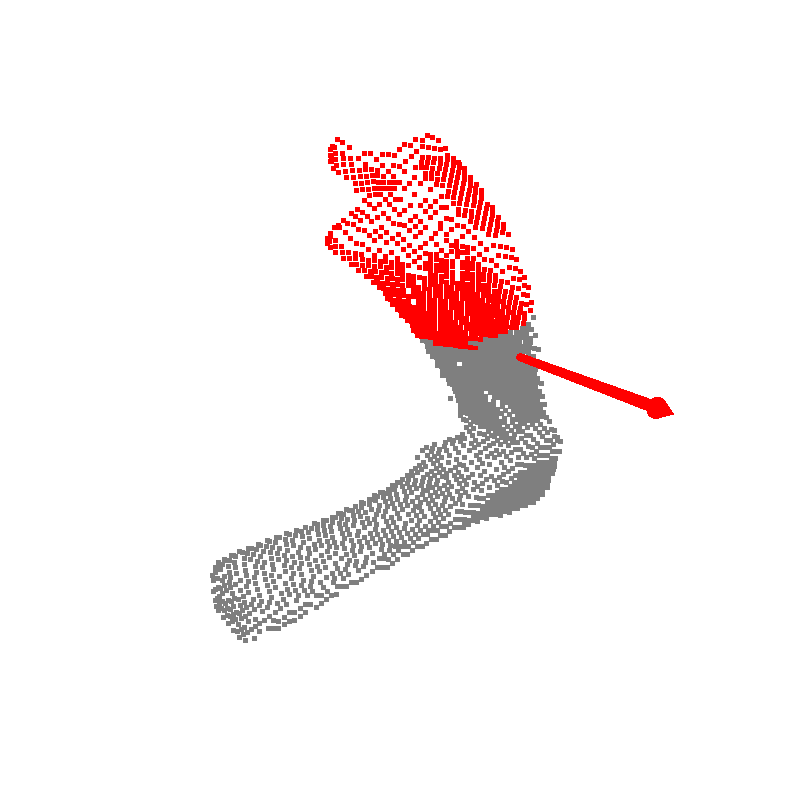} &
    \includegraphics[trim={3cm 7.3cm 3cm 6cm},clip,width=\realdatawidth]{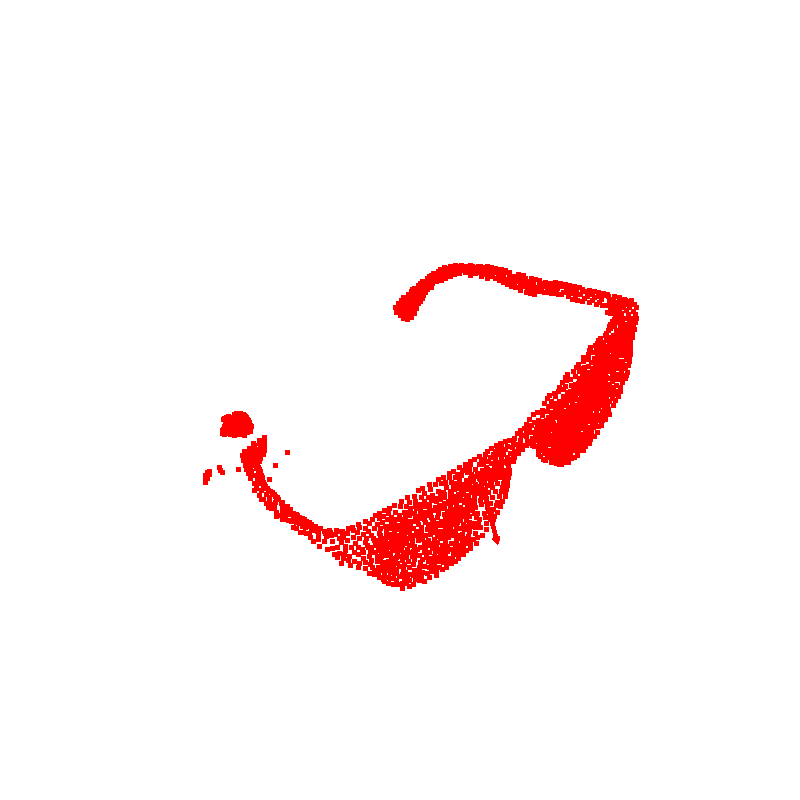} &
    \includegraphics[trim={3cm 5cm 3cm 6cm},clip,width=\realdatawidth]{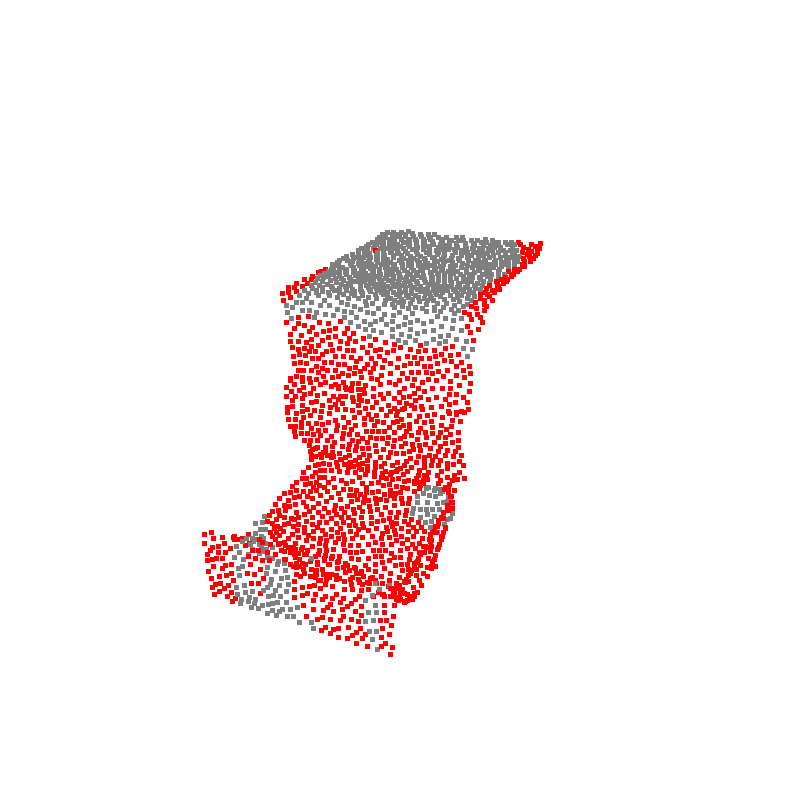} \\

    &
    \rotatebox{90}{\hspace{0cm}\vphantom{A} Artipoint} &
    \includegraphics[trim={6cm 5cm 0cm 5cm},clip,width=\realdatawidth]{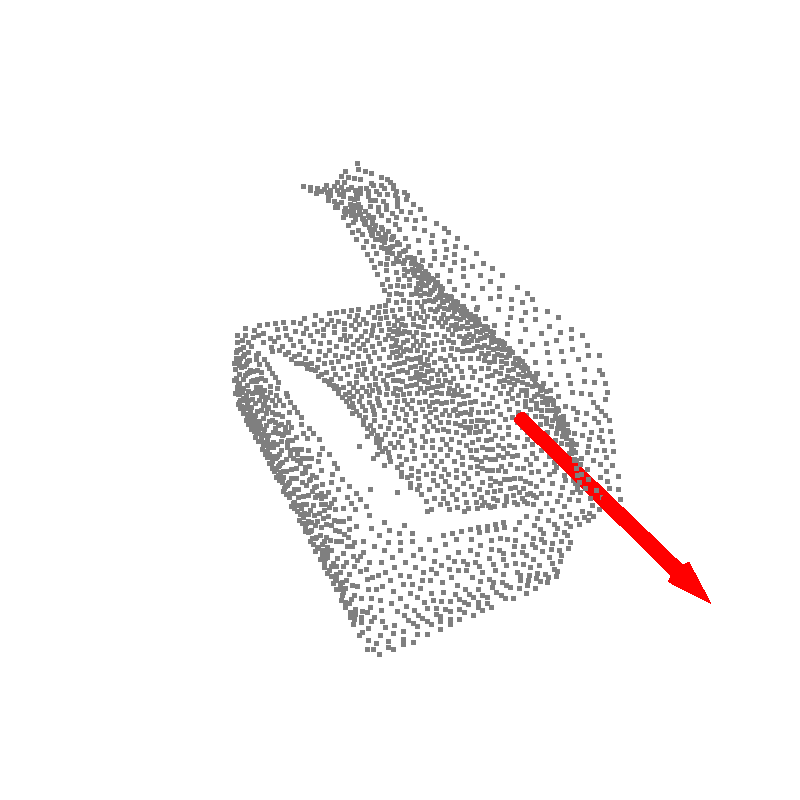} &
    \includegraphics[trim={3cm 5cm 3cm 5cm},clip,width=\realdatawidth]{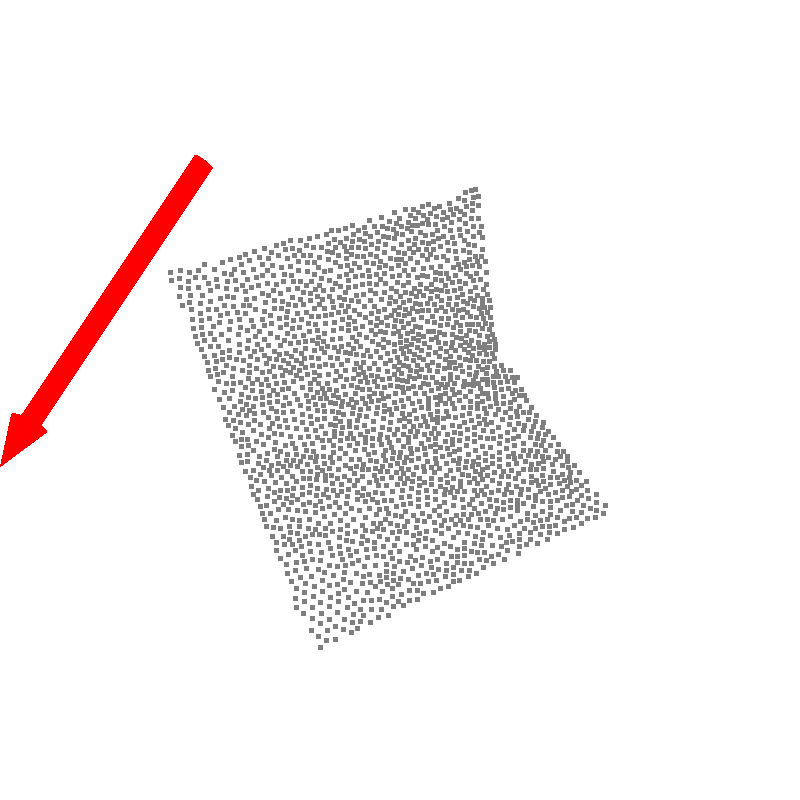} &
    \includegraphics[trim={5cm 6cm 5cm 5cm},clip,width=\realdatawidth]{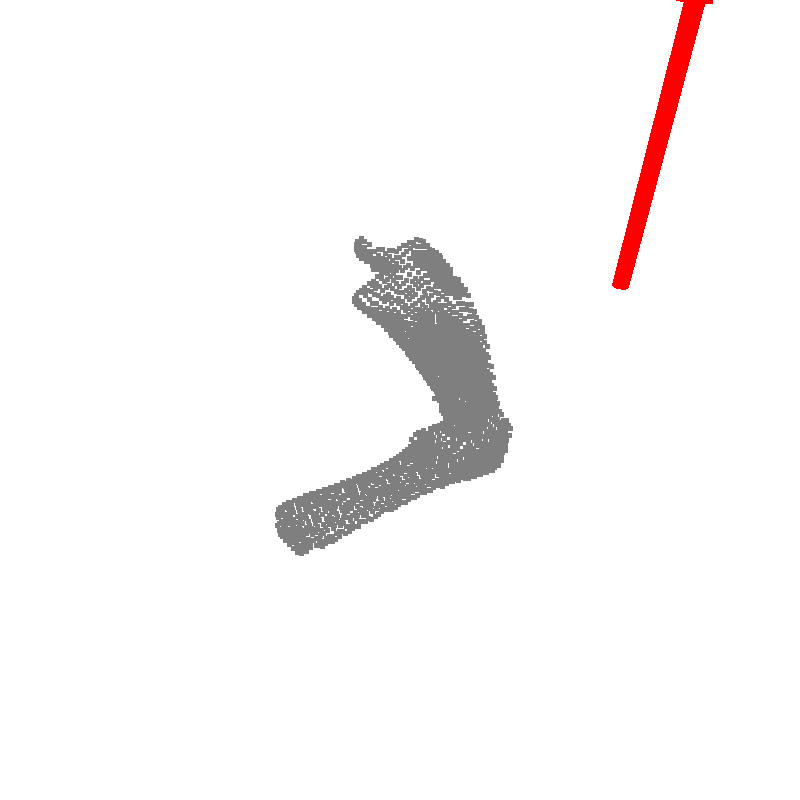} &
    \includegraphics[trim={3cm 7.3cm 3cm 6cm},clip,width=\realdatawidth]{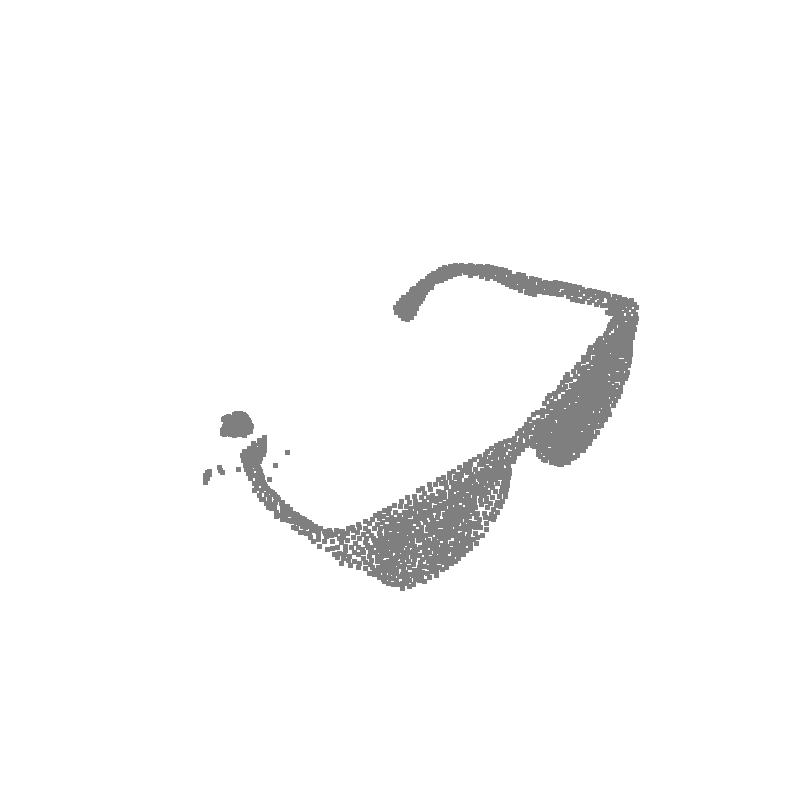} &
    \includegraphics[trim={3cm 5cm 3cm 6cm},clip,width=\realdatawidth]{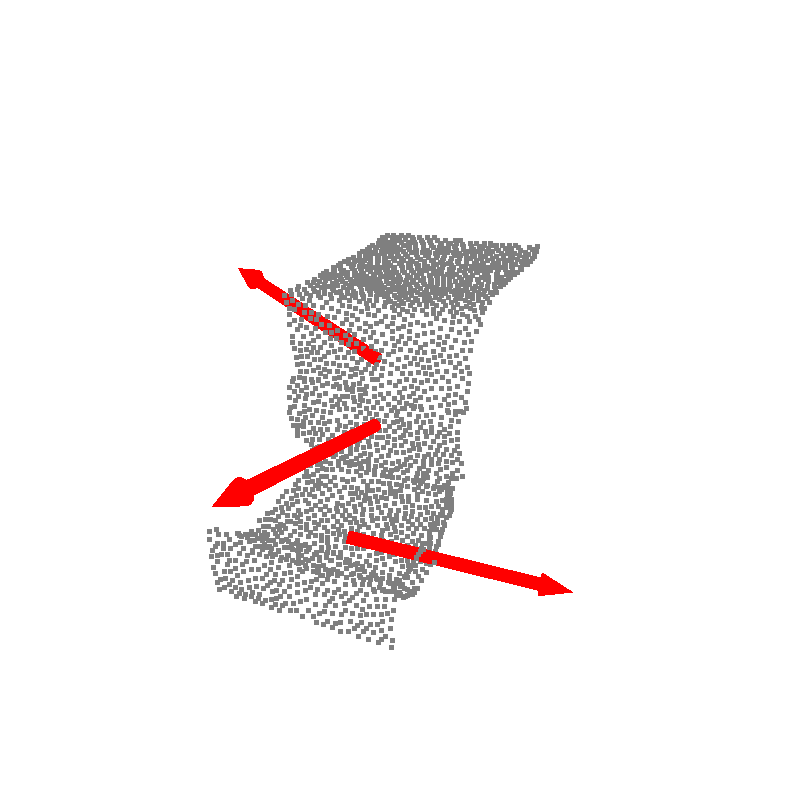} \\
    
    & & Box & Laptop & Stapler & Eyeglasses & Drawer\\
    \end{tblr}
  }
  \caption{\textbf{Qualitative results on sequences from 4art-real.} Red arrows denote the (predicted) joint axes---our results are very close to the ground truth, while the other methods are much more fragile on these challenging sequences. Artipoint does not provide part segmentation, thus only joint axes are shown. $\times$: Articulate-Anything and Video2Articulation are not designed for multi-part objects. \textit{Video results and results on less challenging videos from previous work are provided in the supplementary material. }} 
  \label{fig:real_qualitative}
\end{figure*}
\newcommand{\scale}[2]{\MULTIPLY{#1}{#2}{\myr}\ROUND[2]{\myr}{\myr}\myr}

\begin{table*}[t]
\centering
\setlength{\tabcolsep}{5.2pt}
\caption{
\label{tab:ablation}
\textbf{Ablation experiments.} We report the metrics averaged on categories: Storage2, Storage6, Box, Eyeglasses, Laptop, and Stapler, the categories with the largest numbers of instances. 
Together, the different components of our method contributes to its performance, except for the amount of rotation (Part Rotation) metric, which is slightly better for one configuration. 
}
\begin{adjustbox}{width=0.9\textwidth}
\begin{tabular}{@{}lcccccc@{}}
\toprule
       & mIoU & Axis Ang & Axis Pos & Part Rotation & Part Translation & Type Accuracy \\
\emph{Method} & $\uparrow$ & ($^{\circ}$)~$\downarrow$ & (cm)~$\downarrow$ & ($\degree$)~$\downarrow$ & (cm)~$\downarrow$ & ($\%$)~$\uparrow$ \\
\midrule
w/o flow        & 0.95 & \scale{0.86}{6.72} & \scale{0.77}{12.74} & \scale{0.63}{8.33} & \scale{0.63}{4.56} & 99.84 \\
w/o DINO        & 0.94 & \scale{0.86}{5.83} & \scale{0.77}{11.25} & \scale{0.63}{7.57} & \scale{0.63}{4.59} & 99.89 \\
w/o $\gamma(t)$ & 0.94 & \scale{0.86}{6.25} & \scale{0.77}{12.40} & \scale{0.63}{6.96} & \scale{0.63}{4.83} & 99.97 \\
w/o $\bar{t}$   & 0.93 & \scale{0.86}{5.28} & \scale{0.77}{11.62} & \textbf{\scale{0.63}{5.69}} & \scale{0.63}{4.06} & 99.65  \\
Complete method & \textbf{0.97} & \textbf{\scale{0.86}{5.23}} & \textbf{\scale{0.77}{10.14}} & \scale{0.63}{6.55} & \textbf{\scale{0.63}{3.99}} & \textbf{100}  \\
\bottomrule
\end{tabular}
\end{adjustbox}
\end{table*}


\section{Limitations and Failure Cases}

Our method relies on off-the-shelf 4D reconstruction methods for predicting camera parameters and depth maps. 
In practice, ViPE works very well on our challenging sequences and sim2art is sufficiently robust to handle its occasional mistakes. However, it is possible that 
ViPE's accuracy decreases too much on textureless scenes. Finally, our method struggles to generalize to object categories unseen during training. But since it can rely on synthetic training data only, it is relatively easy to create the required training.

\section{Conclusion}

We presented a method for estimating the joint parameters, part decomposition, and motion magnitudes of an articulated object at every time step from a single input video. Our approach is both robust and accurate across a variety of challenging scenarios. Notably, our method is able to rely exclusively on synthetic data for training. This is particularly valuable, as collecting and annotating real-world data for this problem is extremely time-consuming and often impractical. Synthetic data, by contrast, can be produced in large quantities. As a result, our framework can be scaled easily, and further expanding the synthetic training set should lead to even stronger performance and broader generalization.

\section*{Acknowledgements}
This project was funded in part by the European Union (ERC Advanced Grant explorer Funding ID \#101097259). This work was granted access to the HPC resources of IDRIS under the allocation 2025-AD011014756R1 made by GENCI.

%
%
\bibliographystyle{splncs04}
\bibliography{main}

\beginsupplement
\clearpage
\setcounter{page}{1}

\begin{center}
    \Large \textbf{Supplementary Material: \\ sim2art: Accurate Articulated Object Modeling\\
from a Single Video\\
using Synthetic Training Data Only}
\end{center}
\vspace{0.5cm}


\section{Experimental Results}
\label{sec:experimental_res}

\def\instanceWidth{0.13\linewidth}

\begin{figure*}
  \centering
  \setlength{\tabcolsep}{1.6pt} 
  \begin{tabular}{ccccccc}
    \includegraphics[trim={9cm 4.5cm 13cm 6.3cm},clip,width=\instanceWidth]{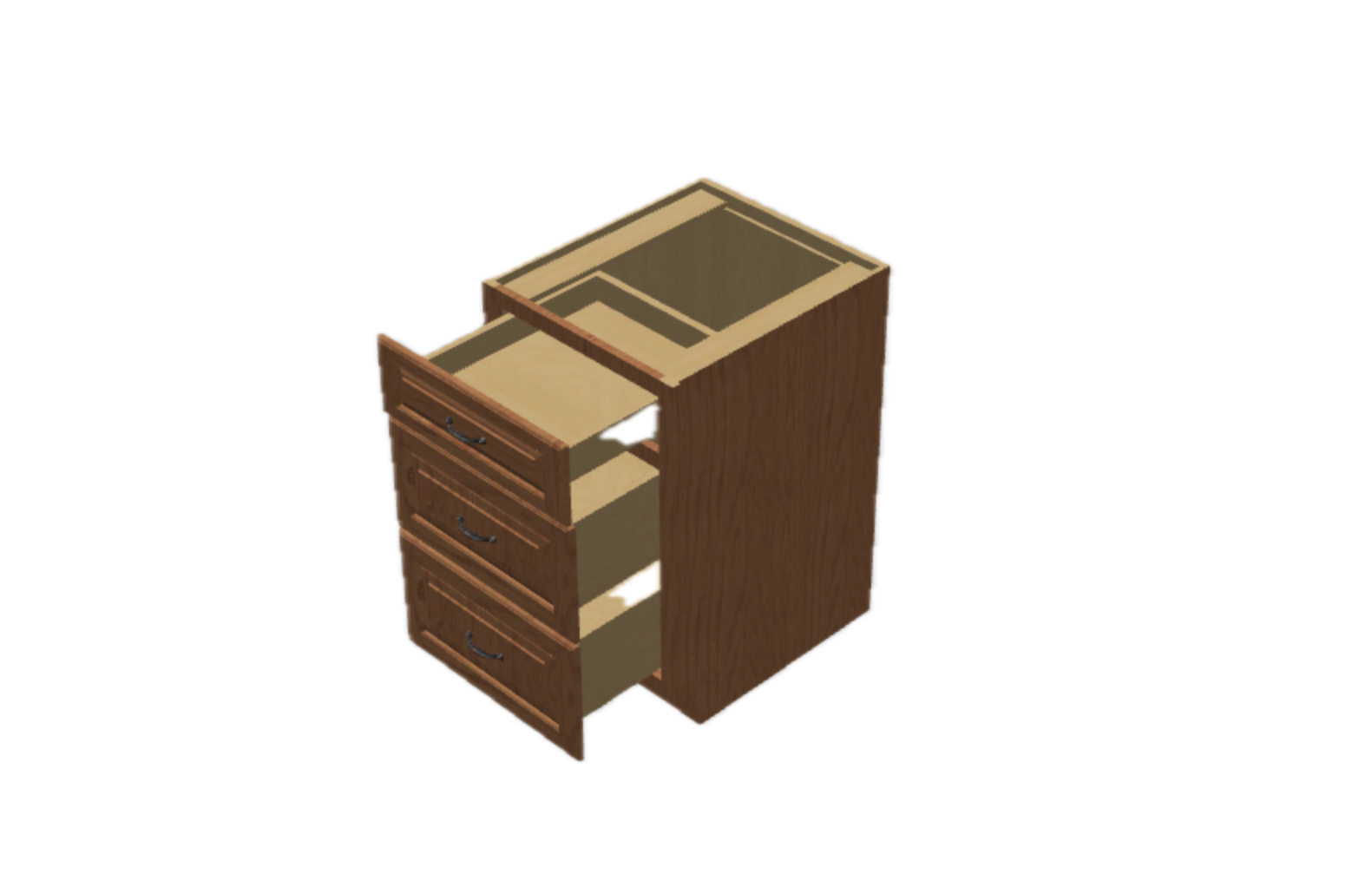} &
    \includegraphics[trim={9cm 5cm 13cm 6.3cm},clip,width=\instanceWidth]{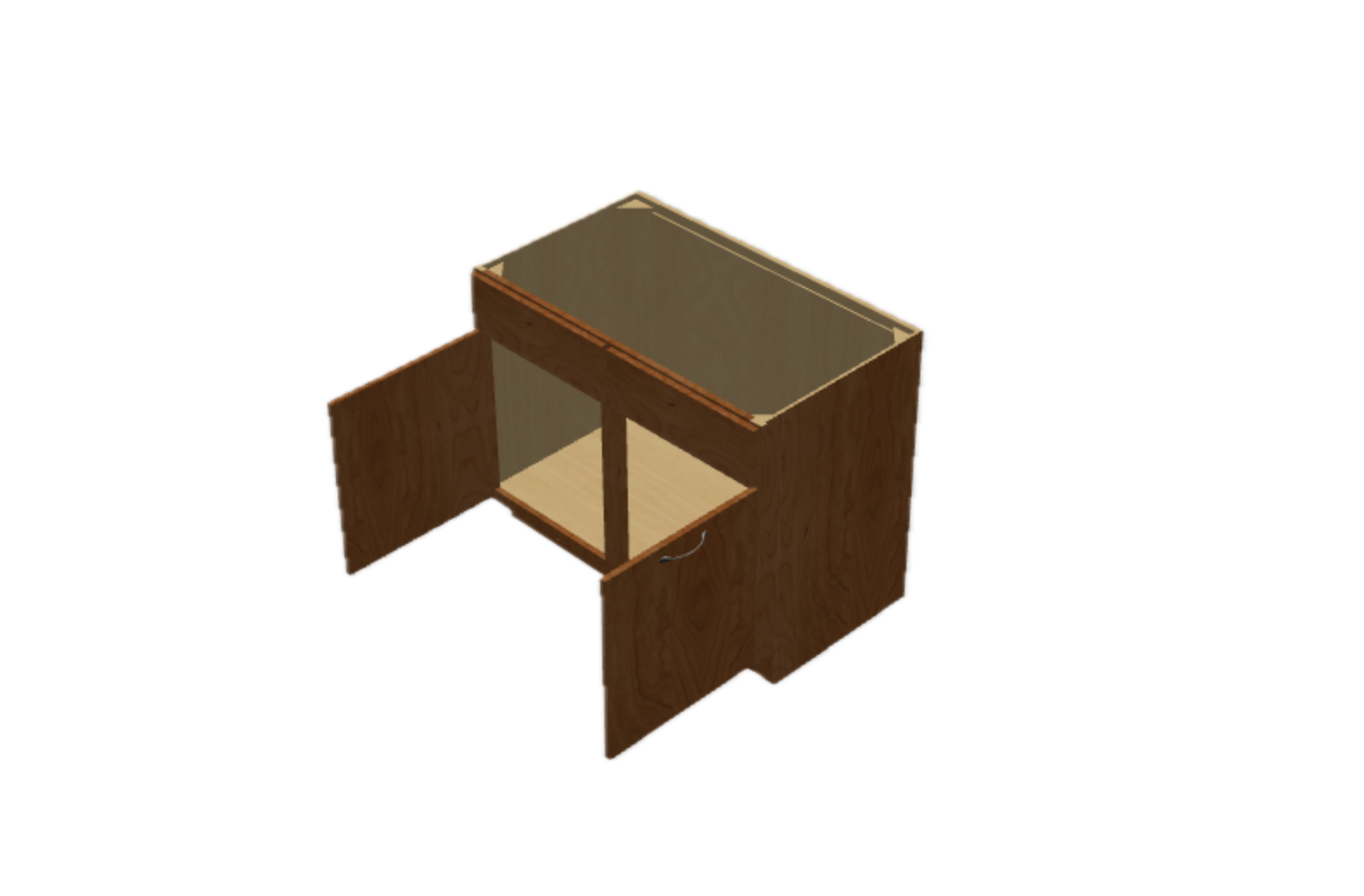} &
    \includegraphics[trim={5cm 5cm 17cm 6.3cm},clip,width=\instanceWidth]{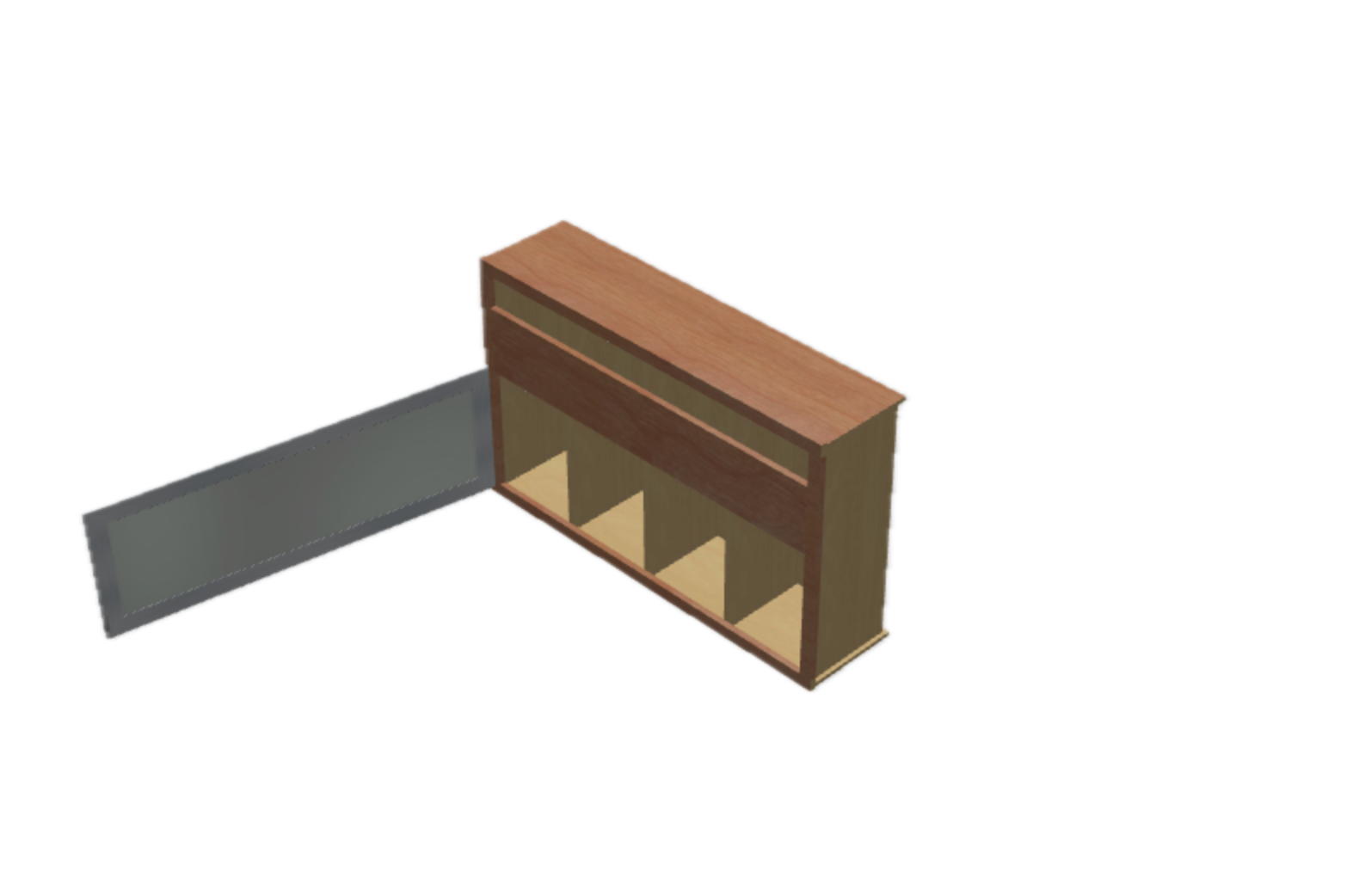} &
    \includegraphics[trim={9cm 4cm 13cm 6.2cm},clip,width=\instanceWidth]{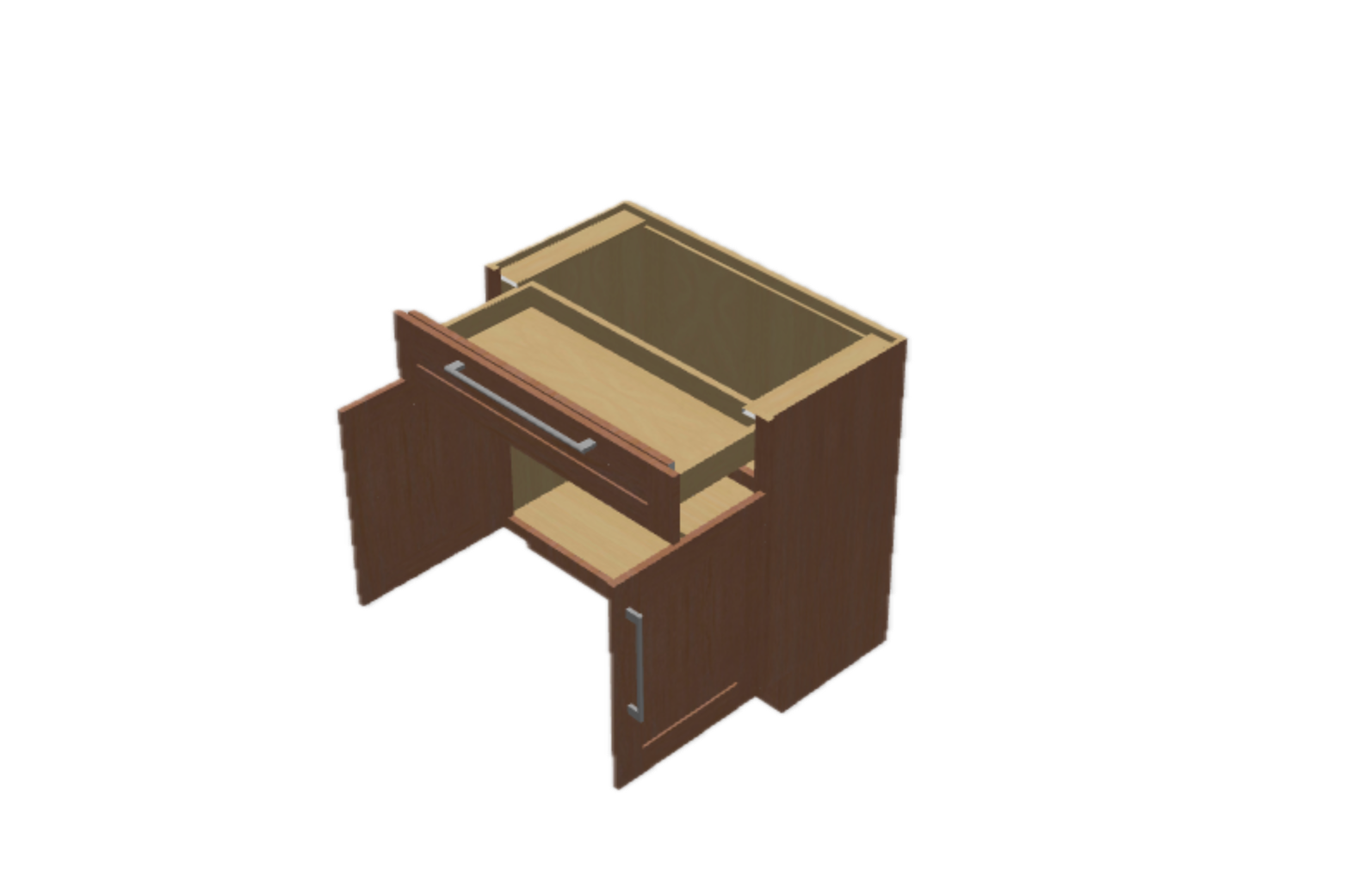} &
    \includegraphics[trim={9cm 4.5cm 13cm 6.9cm},clip,width=\instanceWidth]{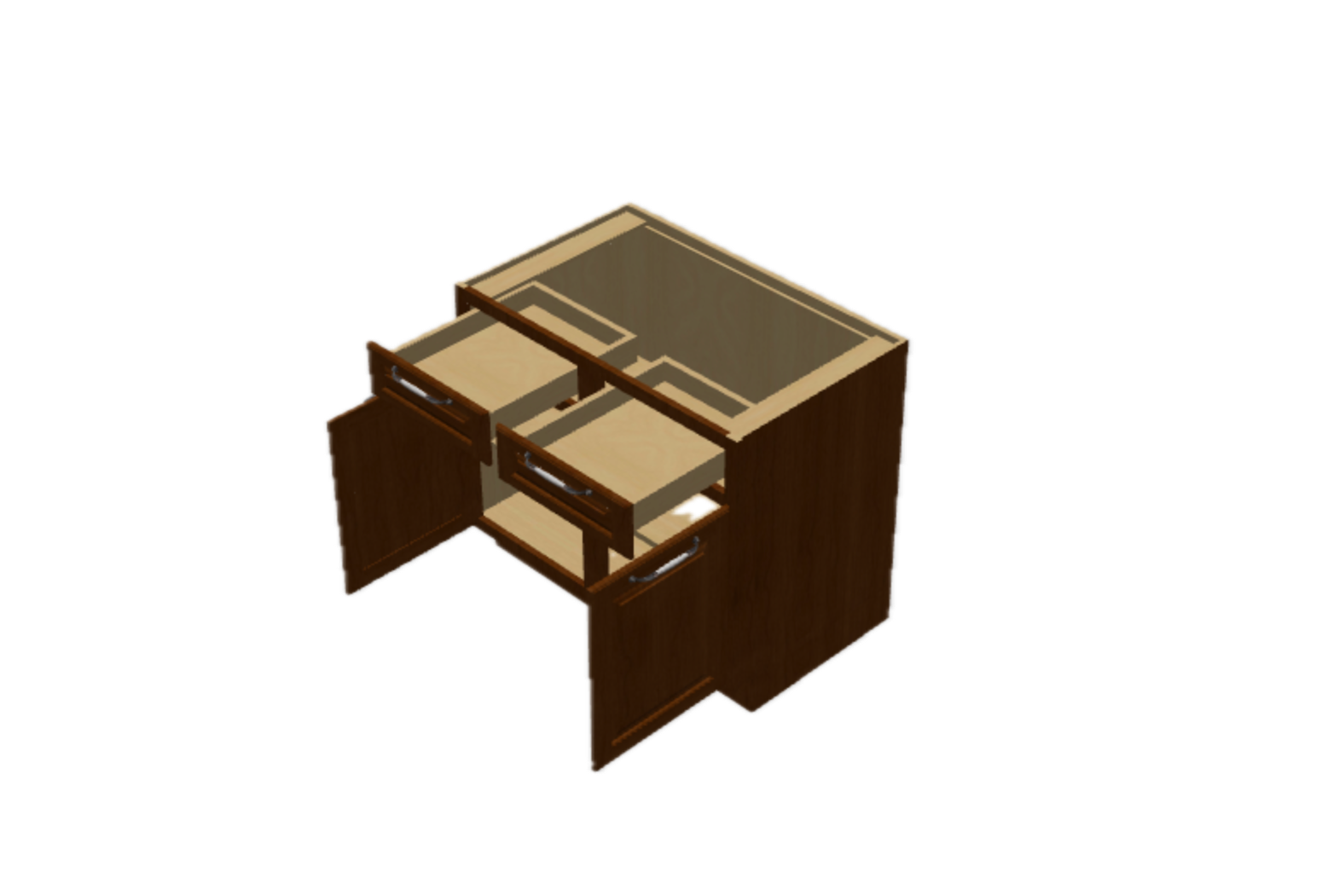} &
    \includegraphics[trim={10cm 2.8cm 12cm 8cm},clip,width=\instanceWidth]{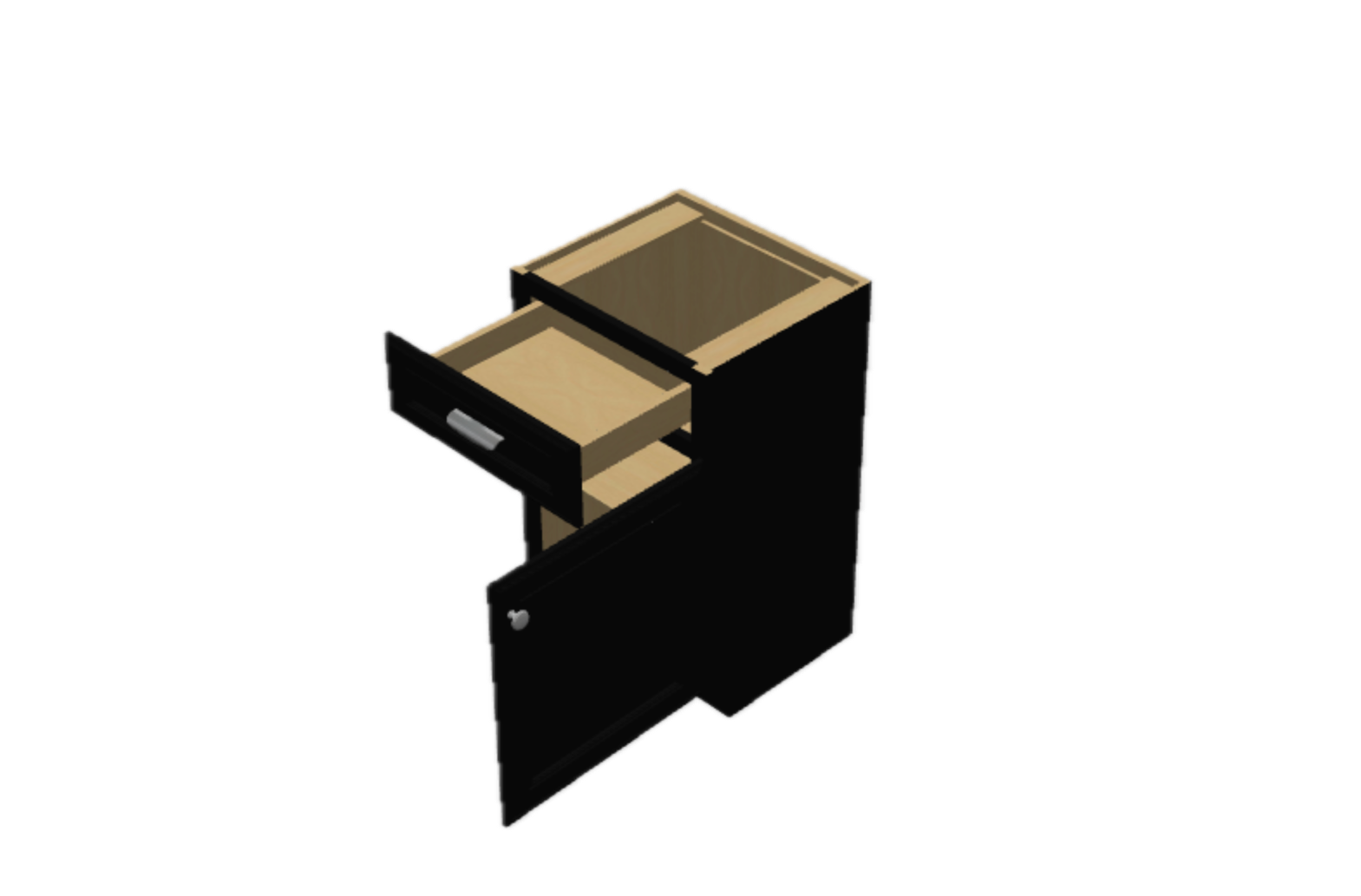} &
    \includegraphics[trim={5cm 2.8cm 5cm 2.8cm},clip,width=\instanceWidth]{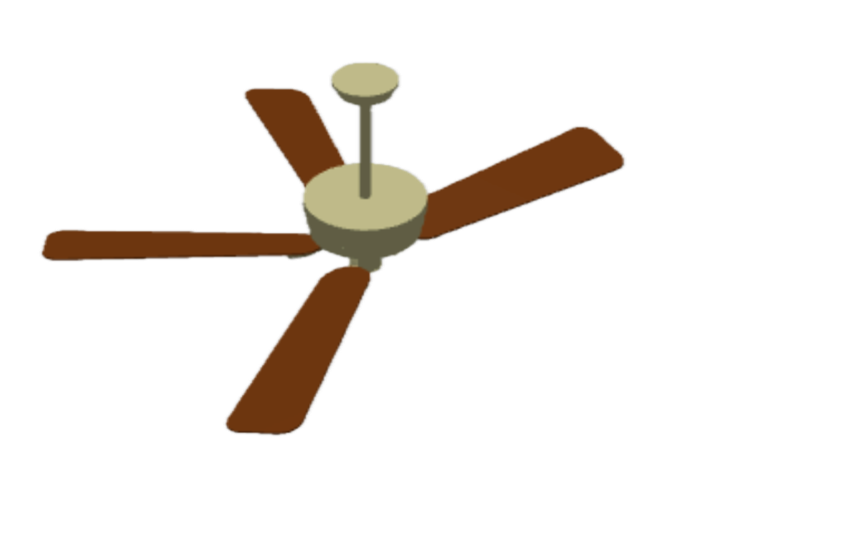} \\
    \small Storage 1 & 
    \small Storage 2 & 
    \small Storage 3 & 
    \small Storage 4 & 
    \small Storage 5 & 
    \small Storage 6 &
    \small Fan \\[4pt]
    \includegraphics[trim={11cm 8.5cm 9cm 0.9cm},clip,width=\instanceWidth]{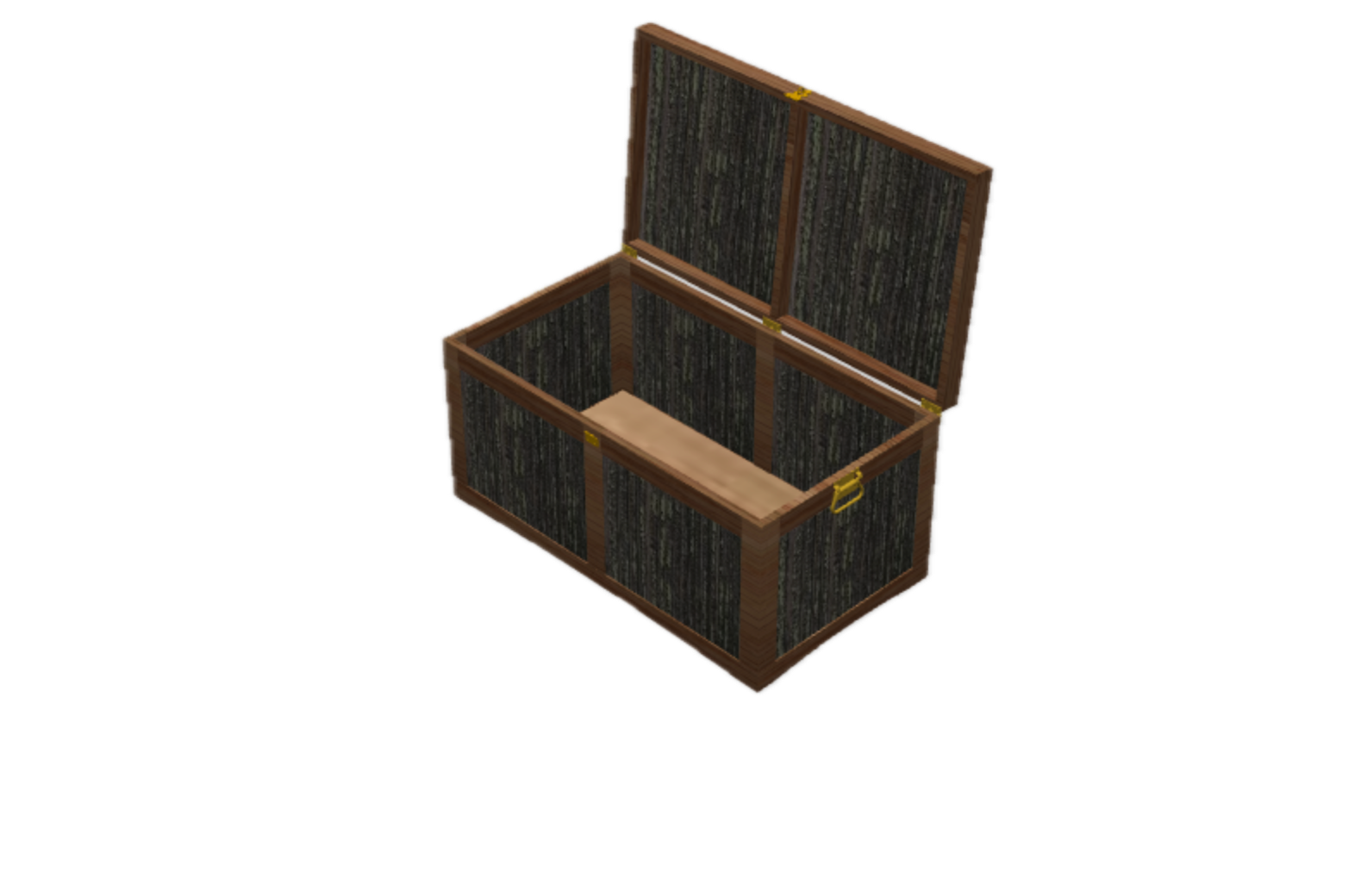} &
    \includegraphics[trim={18cm 10cm 15cm 9.5cm},clip,width=\instanceWidth]{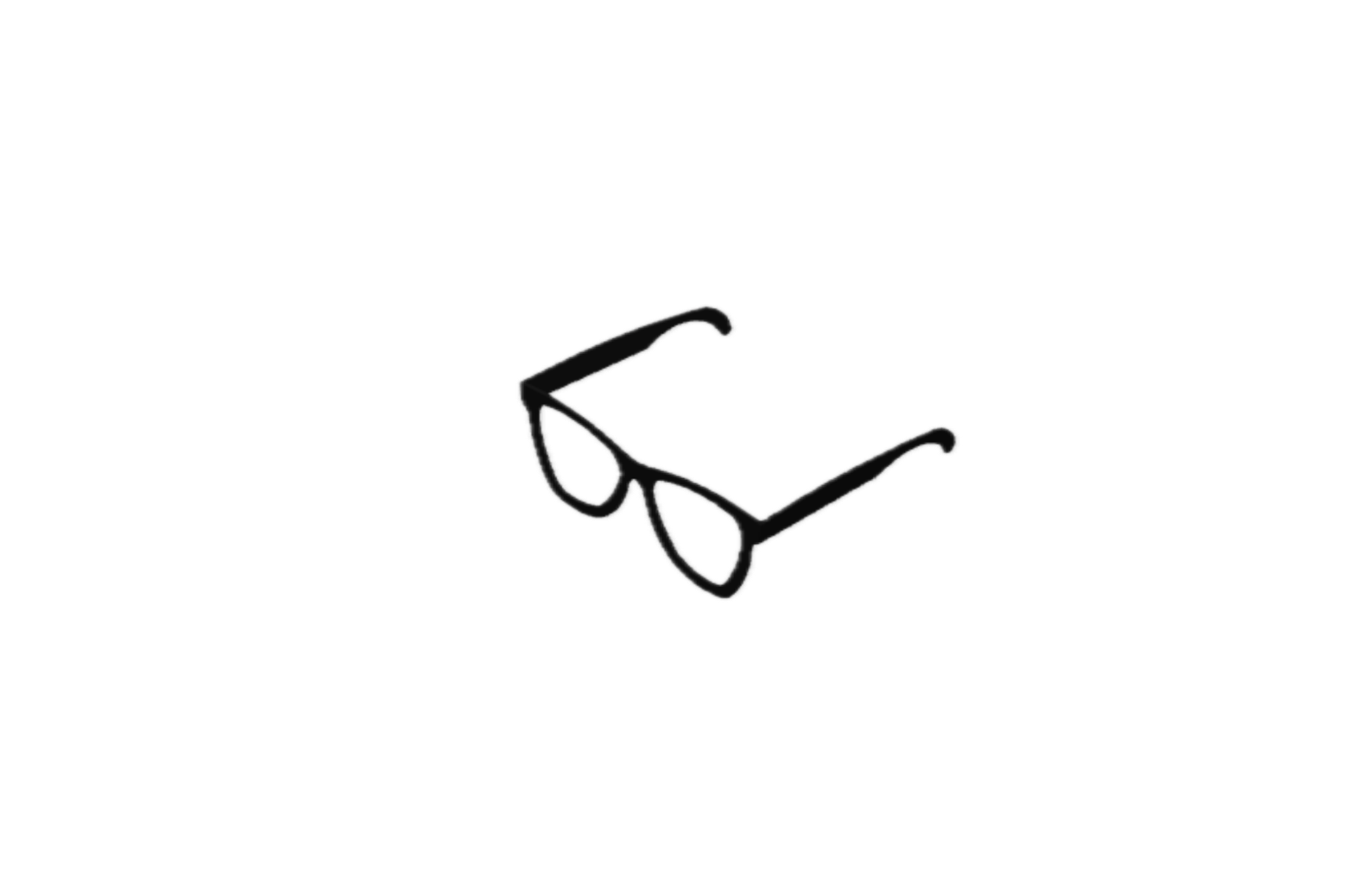} &
    \includegraphics[trim={18cm 13.5cm 17cm 7cm},clip,width=\instanceWidth]{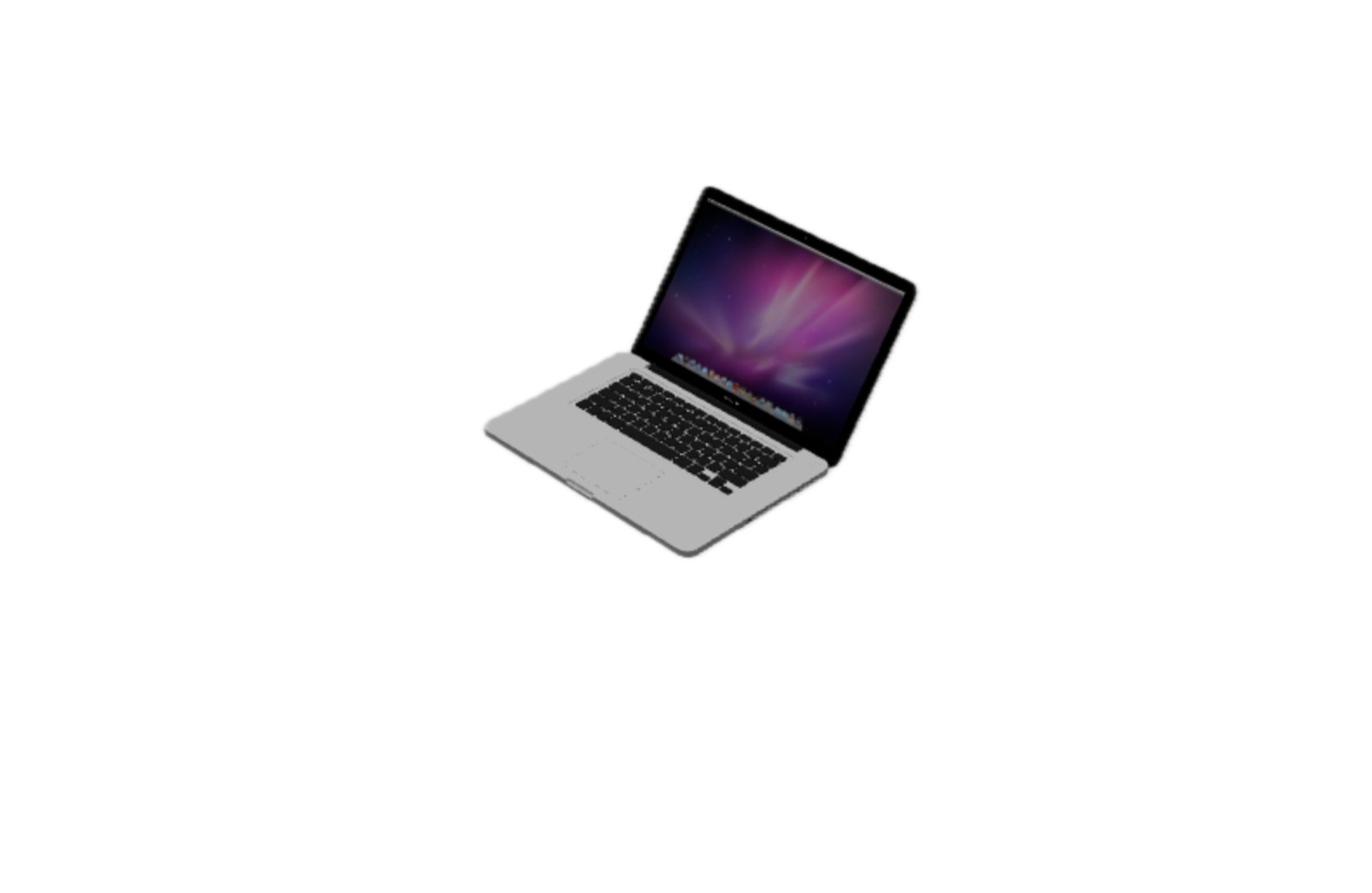} &
    \includegraphics[trim={11cm 7.5cm 11cm 3cm},clip,width=\instanceWidth]{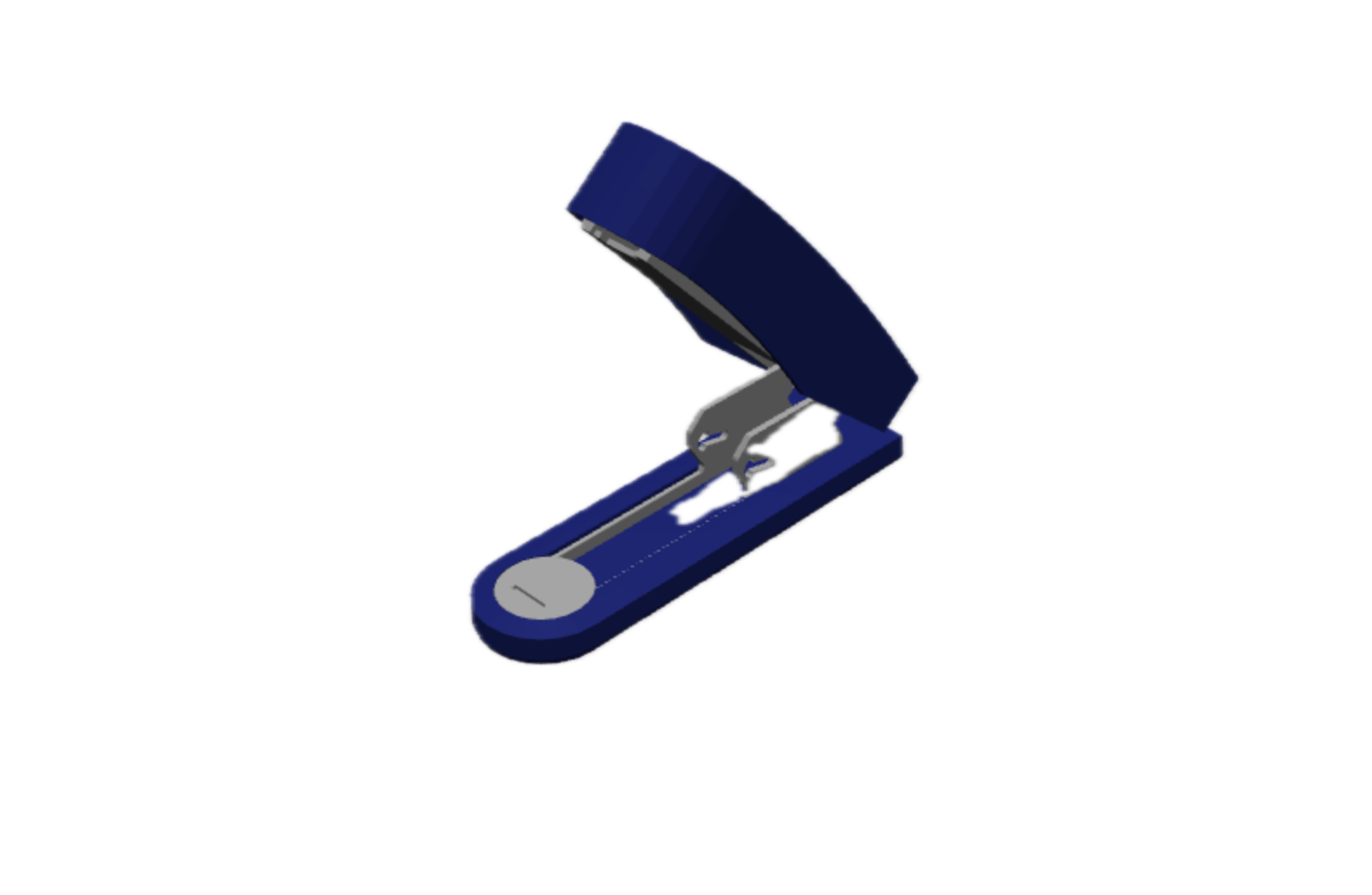} &
    \includegraphics[trim={11cm 3.5cm 11cm 7.5cm},clip,width=\instanceWidth]{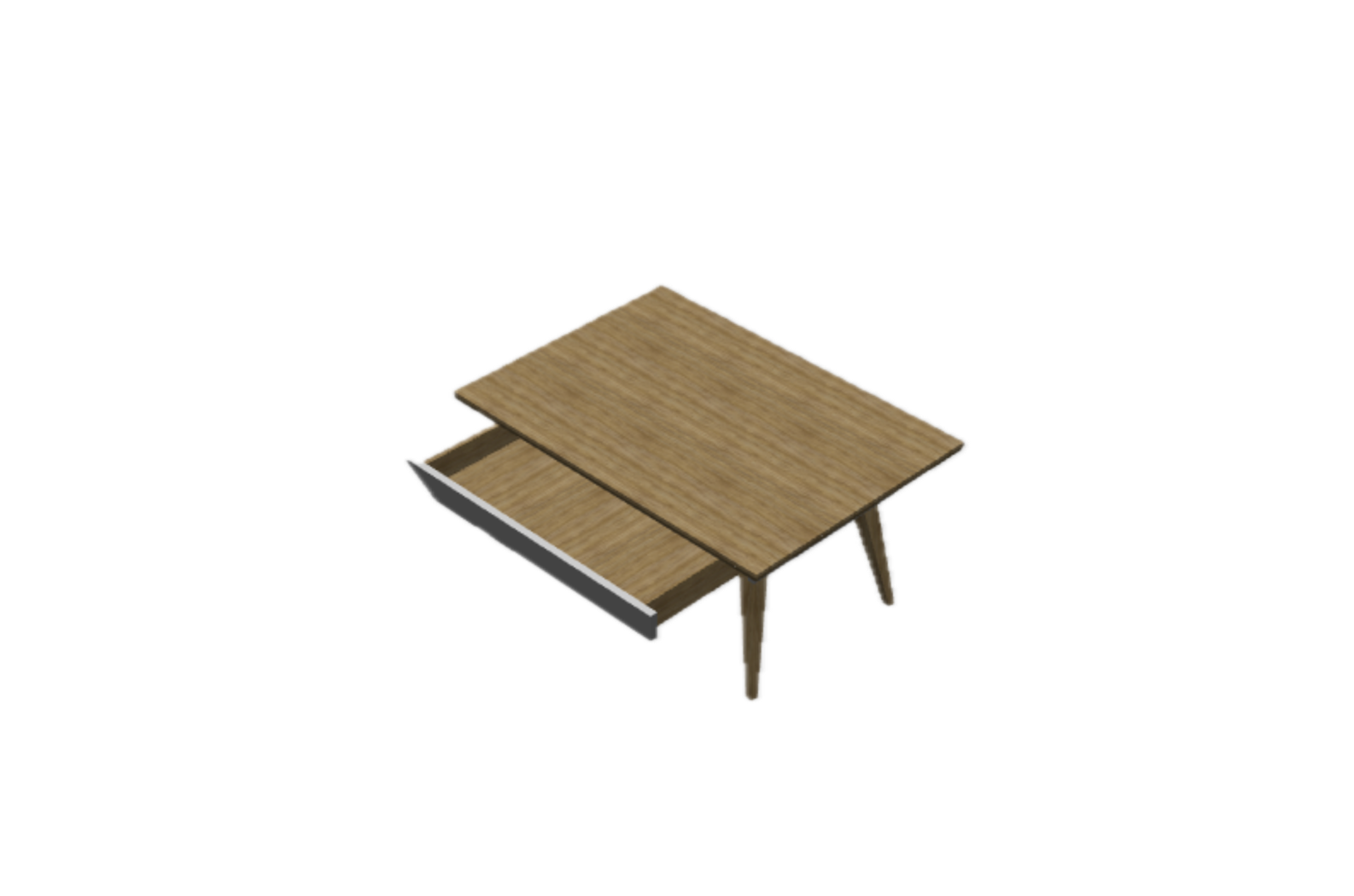} &
    \includegraphics[trim={15cm 6cm 15cm 11cm},clip,width=\instanceWidth]{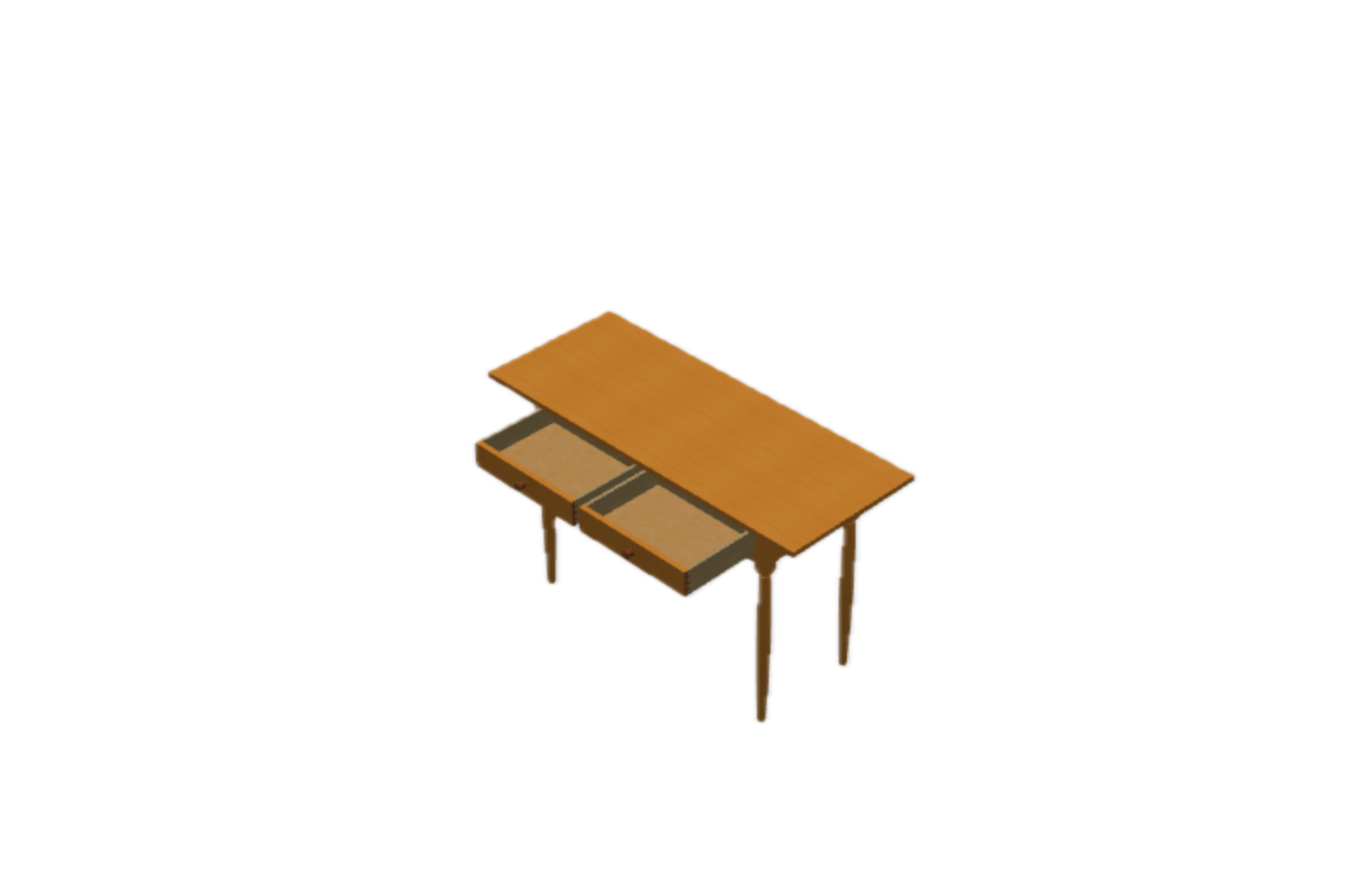} &
    \includegraphics[trim={3cm 5cm 3cm 5cm},clip,width=\instanceWidth]{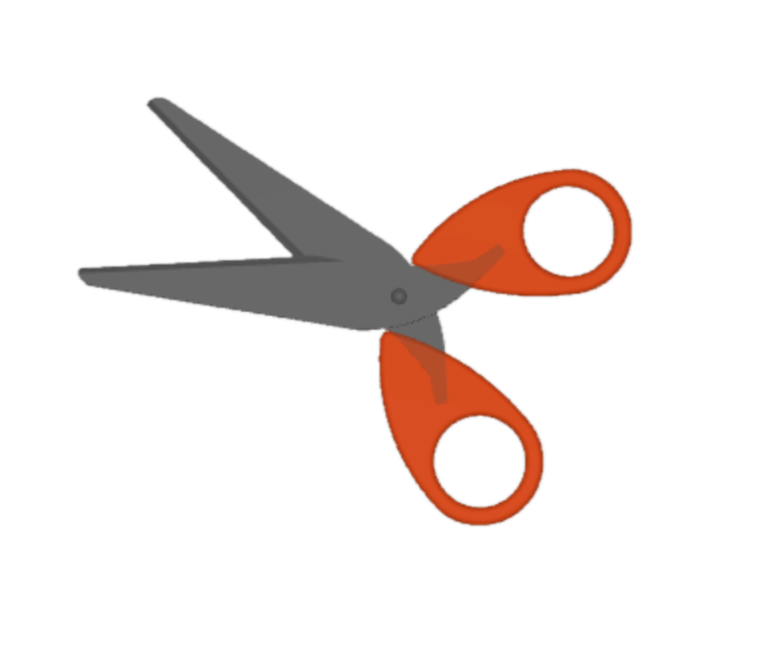} \\
    \small Box & 
    \small Eyeglasses & 
    \small Laptop & 
    \small Stapler & 
    \small Table 1 & 
    \small Table 2 &
    \small Scissors\\
  \end{tabular}
  \caption{\textbf{Representative instances from all the categories used in our dataset.}}
  \label{fig:category_fig}
\end{figure*}

Figure~\ref{fig:category_fig} shows representative instances from all the categories used in our dataset. Figure~\ref{fig:qual_fig_two_part_suppmat} and Figure~\ref{fig:qual_fig_multi_part_suppmat} present experimental results for the remaining categories from the 4art-synth dataset. As shown, our method clearly outperforms the competitors in both part segmentation and joint parameter prediction.

\begin{figure}
    \centering
    \begin{minipage}{0.8\linewidth}
        
        \includegraphics[width=\linewidth]{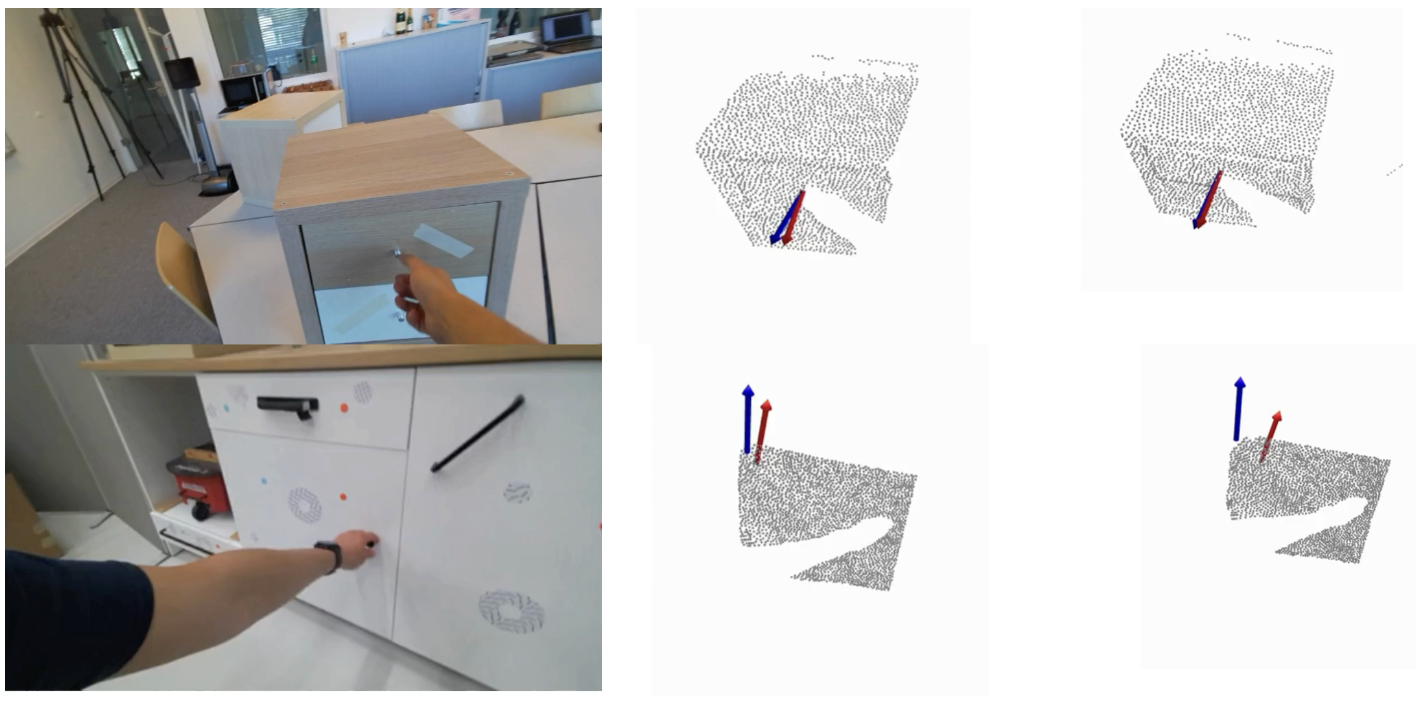}
        
        \vspace{0.1cm}
        
        \begin{minipage}[t]{0.333\linewidth}
            \centering \small
            Representative Frame
        \end{minipage}%
        \begin{minipage}[t]{0.5\linewidth}
            \centering \small
            sim2art (ours)
        \end{minipage}%
        \begin{minipage}[t]{0.15\linewidth}
            \centering \small
            Artipoint
        \end{minipage}
        
    \end{minipage} 
    
    \vspace{0.2cm}
    \caption{\textbf{Qualitative results on the Arti4D dataset.} Ground-truth and predicted joints are indicated by blue and red arrows, respectively. Our method achieves comparable or superior performance to Artipoint on the Arti4D dataset.}
    \label{fig:arti4d_comp} 
\end{figure}

\paragraph{Comparison to baseline methods.} \textbf{In the video supplementary material}, we compare our method against Reart, Articulate-Anything, Video2Articulation, FeatClust, and Artipoint. ReArt predicts part labels only for the canonical timestep. Therefore, we animate the canonical frame point cloud across other timesteps using the predicted joint parameters(e.g. axis, pivot), following the approach of the original authors. Articulate-Anything does not predict part labels; instead, it retrieves pre-existing part meshes from a database and assembles them. Their method is also limited to videos of single-part objects. Therefore, we show the selected object mesh with predicted joint parameters for single-part objects and mark others as $\times$. Additionally, we testded our method on Arti4D dataset provided by Artipoint~\cite{werby2025articulated} paper authors. Since, the dataset does not contain ground-truth part labels, we show only the joint prediction results on Table~\ref{tab:quant_artipoint_data_eval} and Figure~\ref{fig:arti4d_comp}. Our method achieves comparable or superior performance to Artipoint on the Arti4D dataset.

\sisetup{detect-all=true}
\begin{table}
\centering
\def\mywidth{0.8\columnwidth} 
\caption{\textbf{Quantitative results on arti4d (Artipoint) dataset.}} 
\setlength{\tabcolsep}{10pt}
\renewcommand{\arraystretch}{1.1}
\resizebox{\mywidth}{!}{
\sisetup{table-auto-round,table-format=.2}
\begin{tabular}{@{}l|cc}
\toprule

{\textbf{Method}} & {\textbf{Axis Ang ($^\circ$) $\downarrow$}} & {\textbf{Axis Pos (cm) $\downarrow$}} \\

\midrule
Artipoint~\cite{werby2025articulated} & 9.84 & 35.75  \\
sim2art (ours) & \textbf{9.15} & \textbf{11.23} \\
\bottomrule
\end{tabular}%
}
\label{tab:quant_artipoint_data_eval}
\end{table}

\paragraph{Dataset annotation.} We acquired ground-truth annotations for the 4art-real dataset through a guided process. Per-frame part segments were generated with the assistance of SAM2~\cite{ravi2024sam2}. The ground-truth joints were then computed by back-projecting the manually annotated 2D axes into 3D space and calculating the temporal average. As shown in our supplementary videos, this pipeline yields highly accurate ground-truth estimations.  

\paragraph{FeatClust baseline creation.} To rigorously evaluate the efficacy of our approach, we introduce a custom baseline denoted as FeatClust. This baseline relies on explicit feature clustering and classic rigid registration, utilizing the exact same preprocessing pipeline as our proposed framework (including depth maps, scene flow, and DINO features).

Formally, for each point $p_t^i \in \mathbb{R}^3$ in the point cloud sequence, we construct a unified feature representation $v_t^i$ by concatenating its 3D spatial coordinates, its projected DINO semantic feature $\phi_t^i \in \mathbb{R}^3$, and its associated scene flow vector $f_t^i \in \mathbb{R}^3$:

\begin{equation}
    v_t^i = [p_t^i, \phi_t^i, f_t^i]^\top \> ,
\label{eq:k_means}
\end{equation}

To achieve part segmentation, we apply the $k$-means clustering algorithm over the dense feature set $\{v_t^i\}_{i=1, t=1}^{N, T}$, assigning each point $p_t^i$ a discrete part label $l_t^i \in \{1, \dots, K\}$.

Following segmentation, we estimate the kinematics of each segmented part. First, the points belonging to the initial frame are forward-warped to the target timestep $t$ using the estimated scene flow:

\begin{equation}
    \hat{p}_t^i = p_0^i + f_0^i ,
\label{eq:k_means}
\end{equation}

For each predicted object part $m$, we compute the optimal rigid transformation—comprising a rotation matrix $R_m \in SO(3)$ and a translation vector $t_m \in \mathbb{R}^3$—that aligns the initial part geometry with its warped counterpart. This is solved analytically via the Kabsch~\cite{umeyama-pami91-leastsquaresestimation} algorithm by minimizing the squared deviation:

\begin{equation}
    R_m, t_m = \text{Kabsch}(\{p_t^i \mid l_t^i = m\}, \{\hat{p}_t^i \mid l_t^i = m\}) ,
\label{eq:k_means}
\end{equation}

Finally, the explicit kinematic joint parameters (e.g., rotation axis, pivot point, or translation axis) for the target part are analytically extracted from the recovered rigid transformation matrix $R_m$ and translation vector $t_m$.


\def\qualitWidth{0.20\linewidth}

\setlength{\tabcolsep}{0pt}
\begin{figure*}
  \centering
  
\begin{tabular}{c@{$\;$}c@{$\;\;$}cccccc}

\rotatebox{90}{\hspace{0.5cm}\vphantom{A}Ground} 
\rotatebox{90}{\hspace{0.6cm}\vphantom{A}Truth} &
\includegraphics[trim={5cm 5cm 5cm 5cm},clip,width=\qualitWidth]{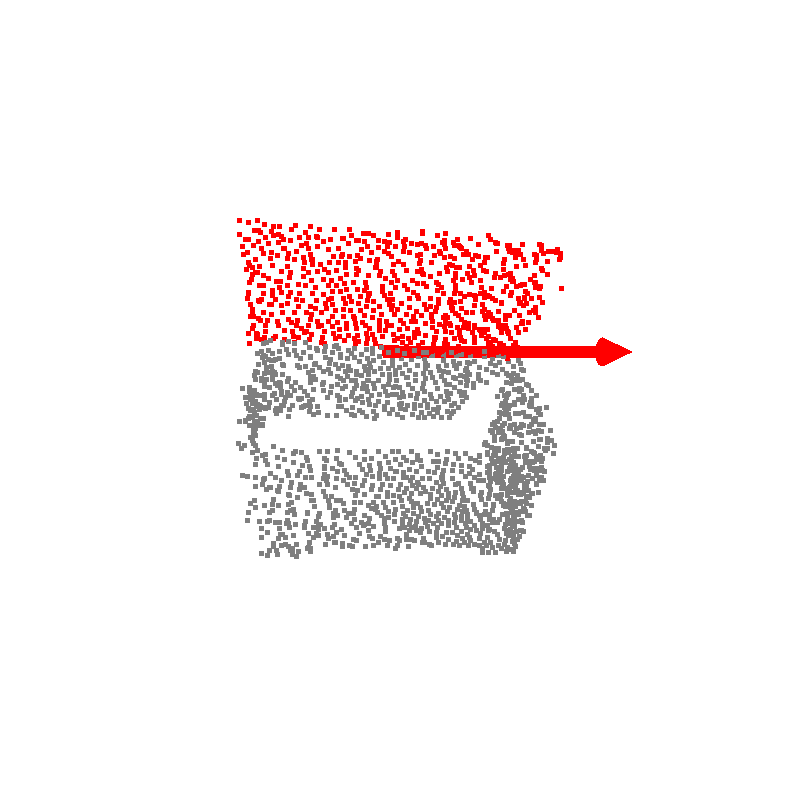} &
\includegraphics[trim={5cm 3cm 5cm 3cm},clip,width=\qualitWidth]{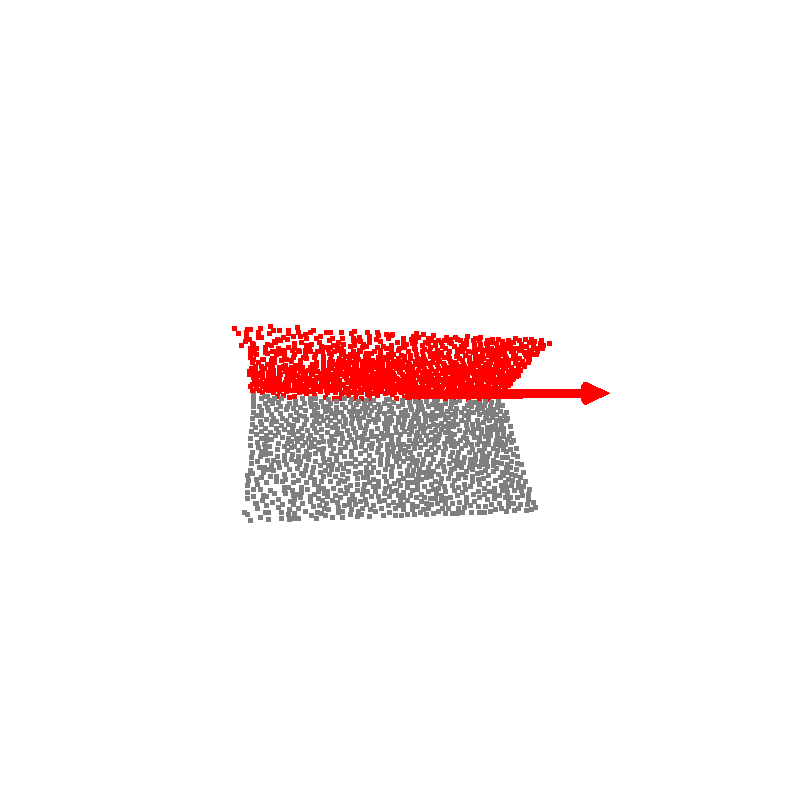} &
\includegraphics[trim={3cm 3cm 3cm 3cm},clip,width=\qualitWidth]{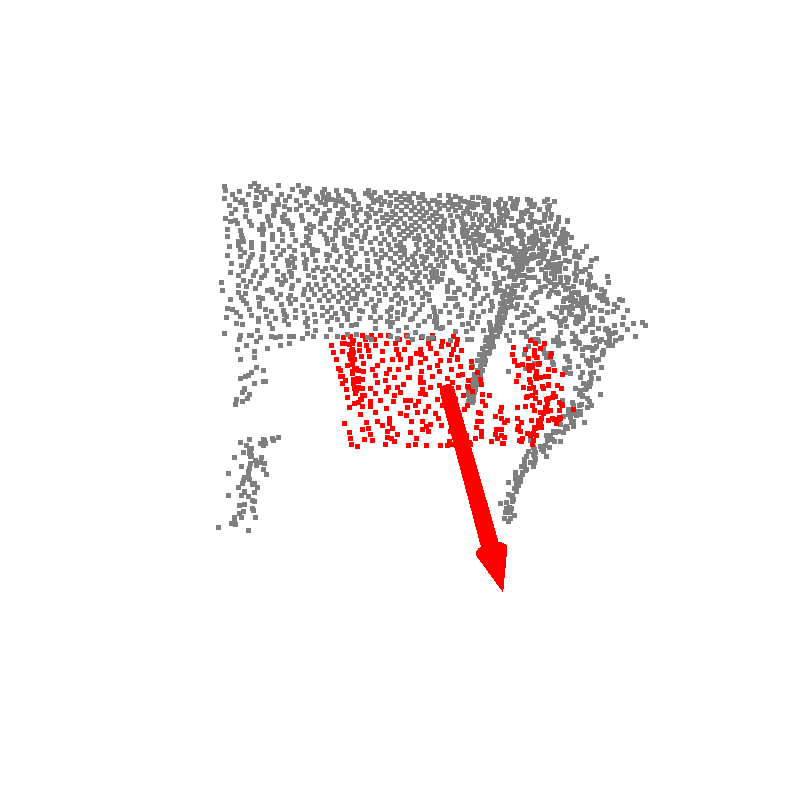} &
\includegraphics[trim={3cm 3cm 3cm 3cm},clip,width=\qualitWidth]{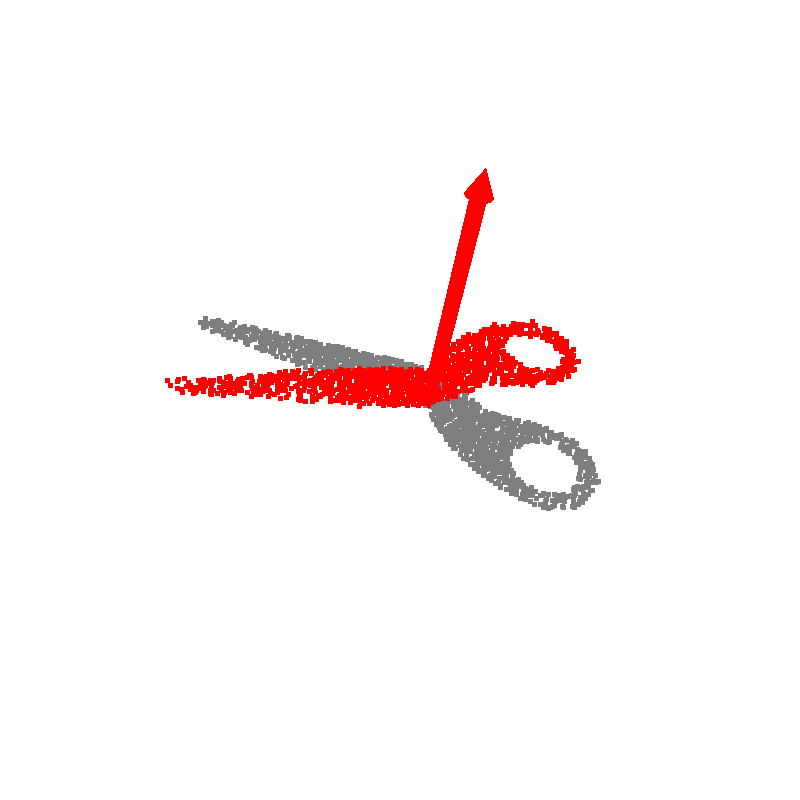} &
\includegraphics[trim={4cm 3cm 4cm 3cm},clip,width=\qualitWidth]{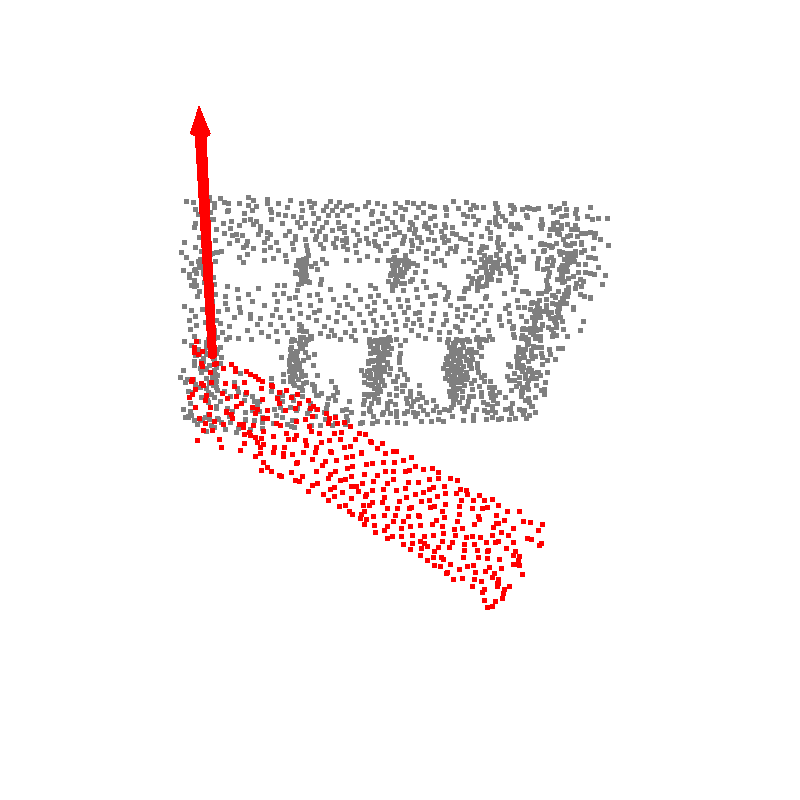}\\[-0.5cm]

\rotatebox{90}{\;\;\;\;\;sim2art} 
\rotatebox{90}{\;\;\;\;\;\;(ours)} &
\includegraphics[trim={5cm 5cm 5cm 5cm},clip,width=\qualitWidth]{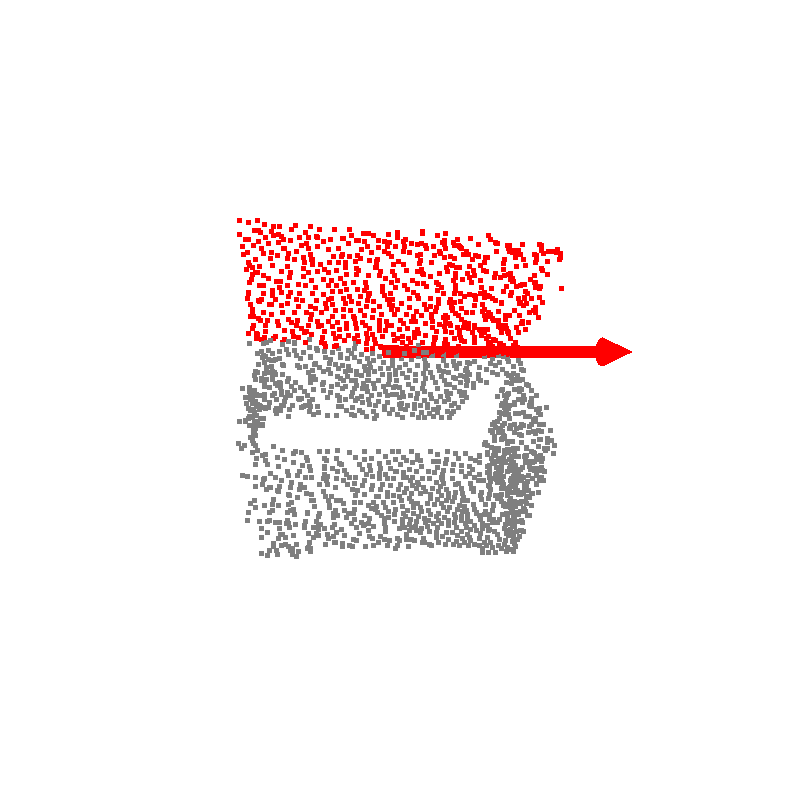} &
\includegraphics[trim={5cm 3cm 5cm 3cm},clip,width=\qualitWidth]{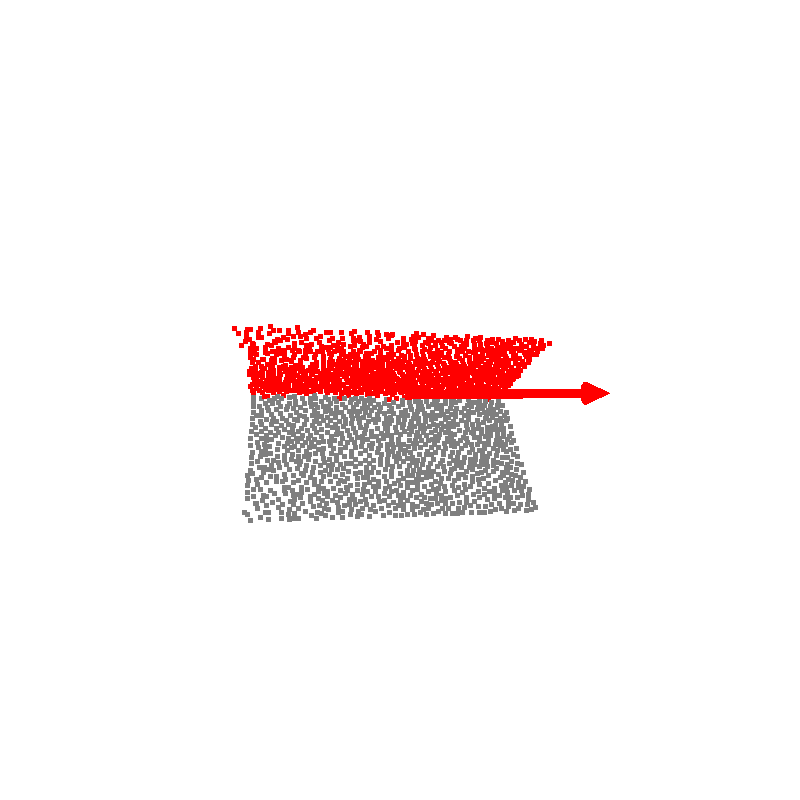} &
\includegraphics[trim={3cm 3cm 3cm 3cm},clip,width=\qualitWidth]{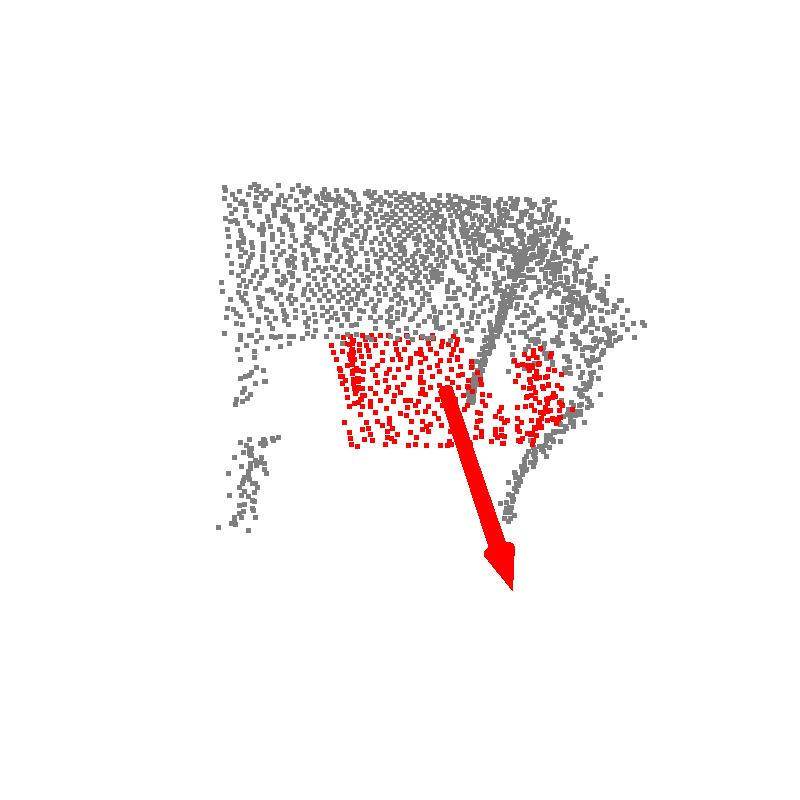} &
\includegraphics[trim={3cm 3cm 3cm 3cm},clip,width=\qualitWidth]{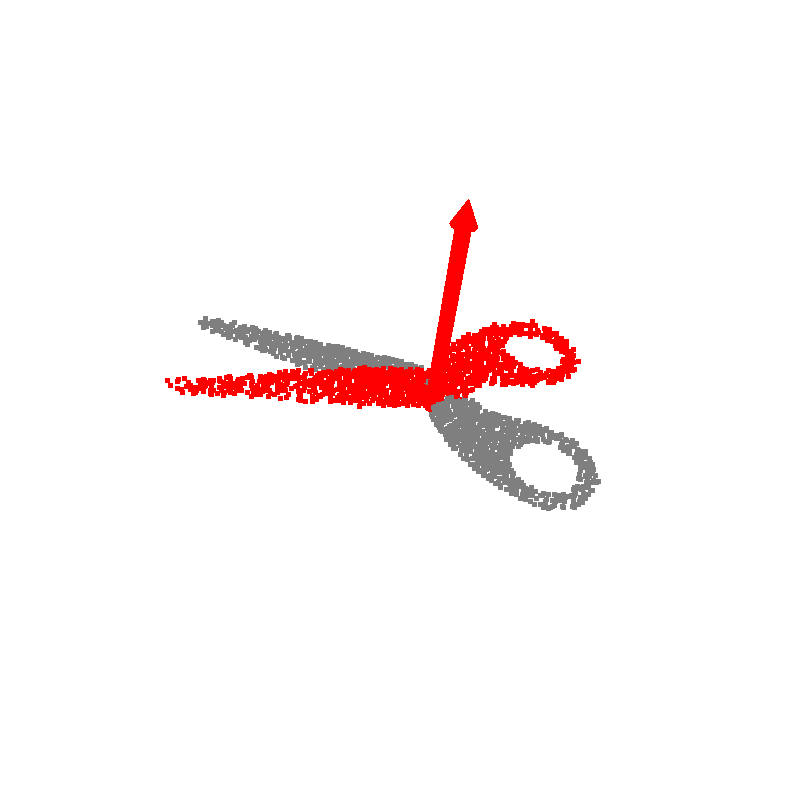} &
\includegraphics[trim={4cm 3cm 4cm 3cm},clip,width=\qualitWidth]{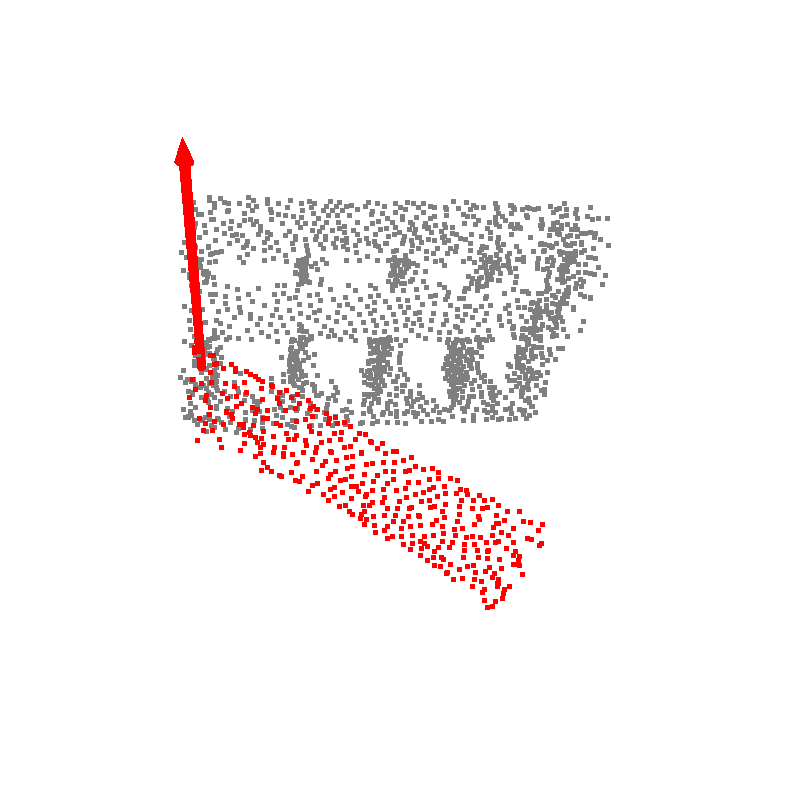}\\[-0.0cm]

\rotatebox{90}{\;\;\;\;\;Articulate-} 
\rotatebox{90}{\;\;\;\;\;Anything} &
\includegraphics[trim={5cm 5.5cm 5cm 5.5cm},clip,width=\qualitWidth]{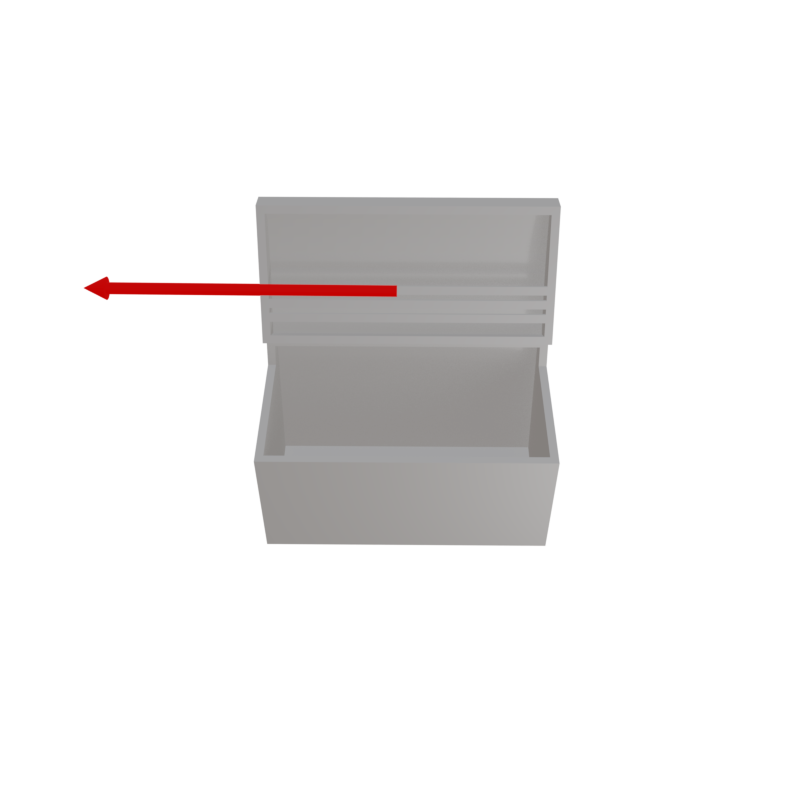} &
\includegraphics[trim={5cm 5cm 5cm 5cm},clip,width=\qualitWidth]{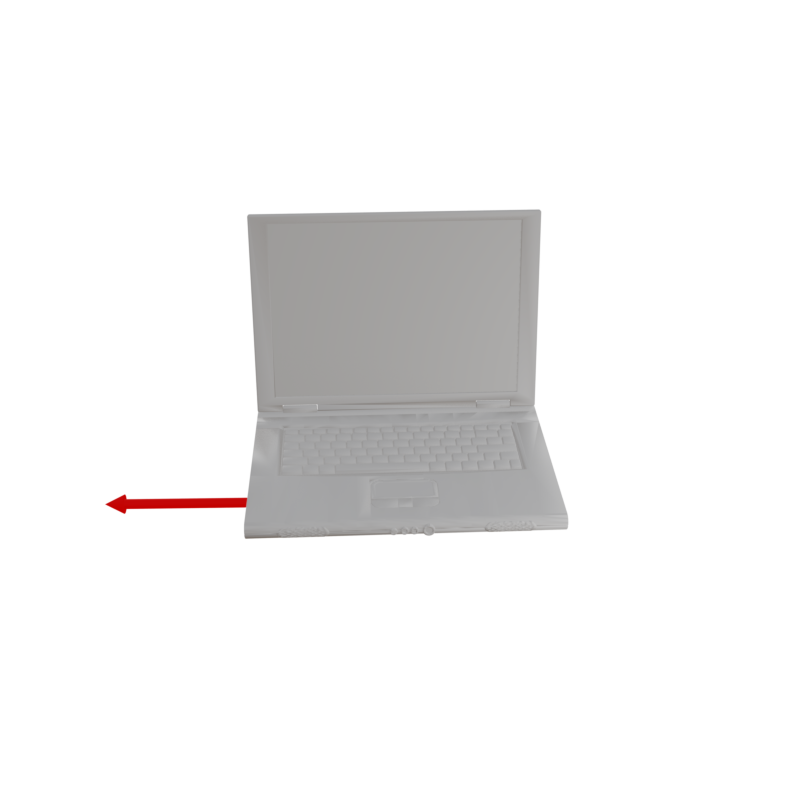} &
\includegraphics[trim={4cm 3cm 4cm 3cm},clip,width=\qualitWidth]{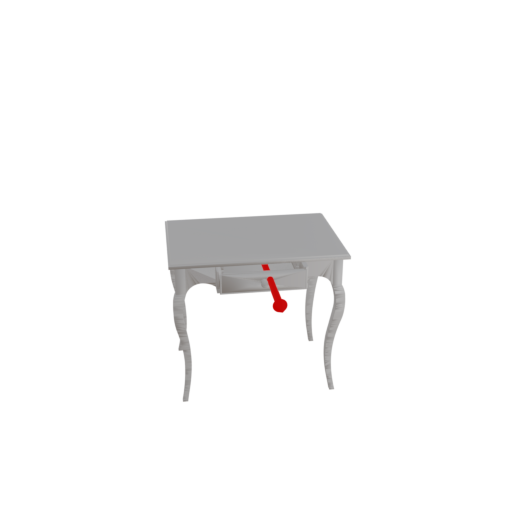} & 
\begin{minipage}{\qualitWidth}
    \centering 
    \raisebox{2.5cm}{(failed)} 
\end{minipage} &
\includegraphics[trim={4cm 3cm 4cm 3cm},clip,width=\qualitWidth]{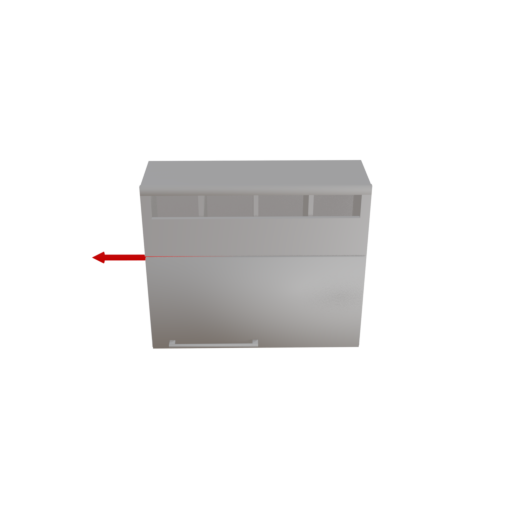}\\ [-0.8cm]


\rotatebox{90}{\hspace{1.0cm}\vphantom{A}Reart} &
\includegraphics[trim={3cm 3cm 3cm 3cm},clip,width=\qualitWidth]{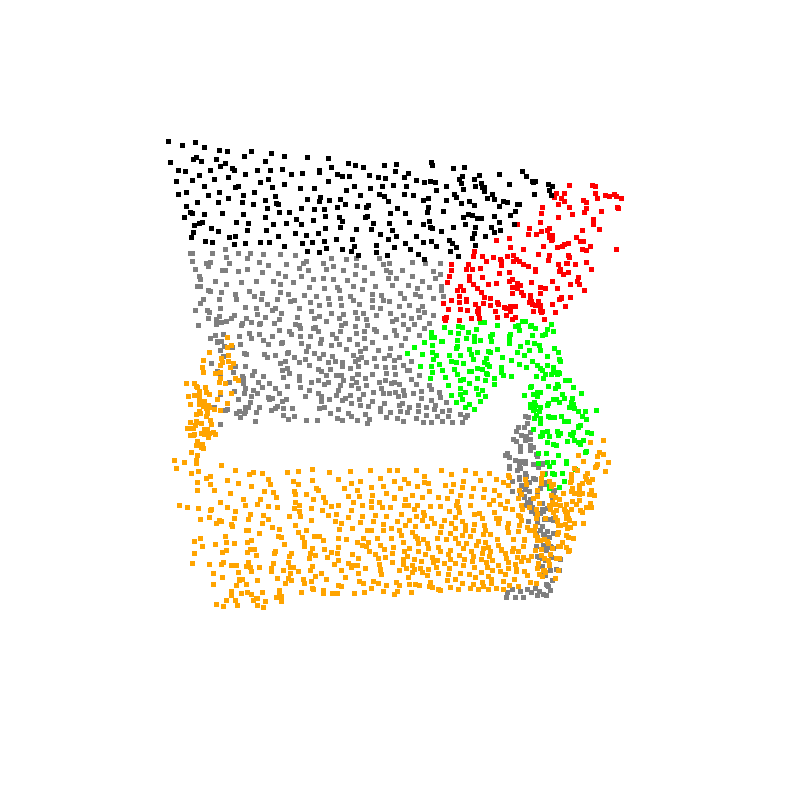} &
\includegraphics[trim={8cm 8cm 8cm 10cm},clip,width=\qualitWidth]{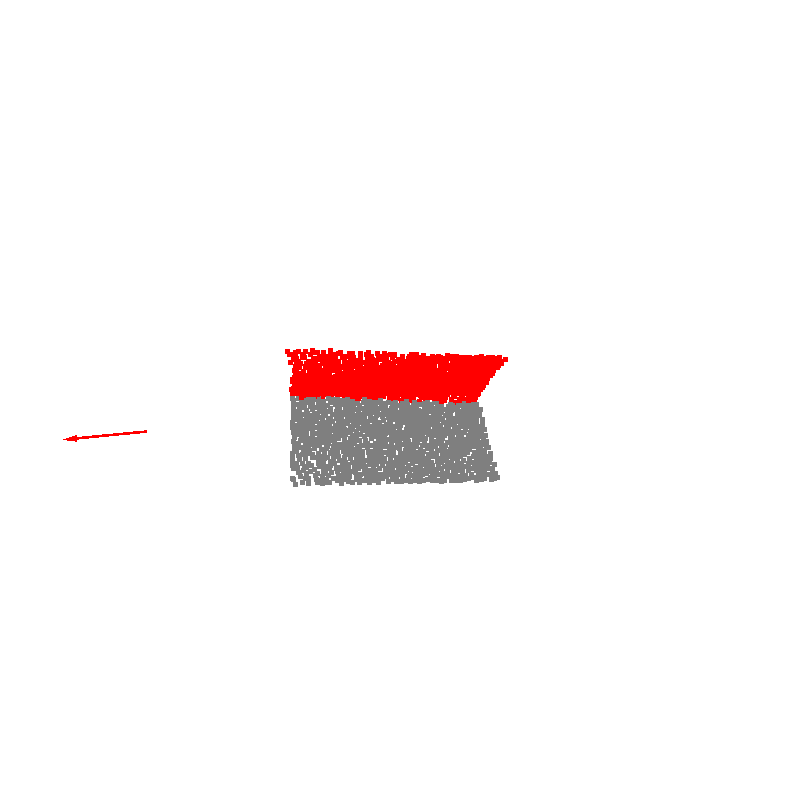} &
\includegraphics[trim={5cm 5cm 5cm 5cm},clip,width=\qualitWidth]{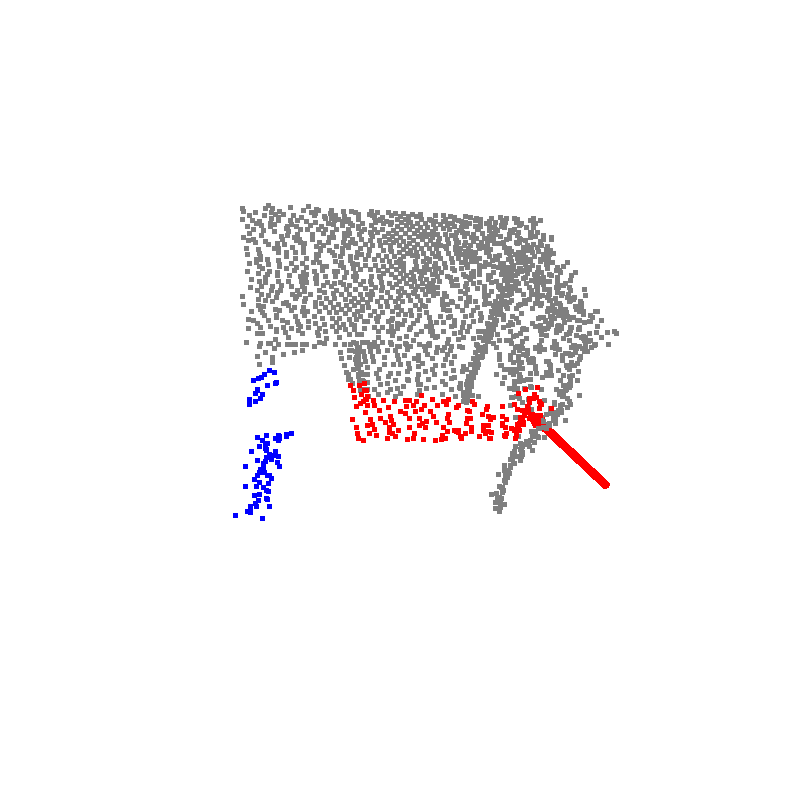} &
\includegraphics[trim={3cm 5cm 3cm 5cm},clip,width=\qualitWidth]{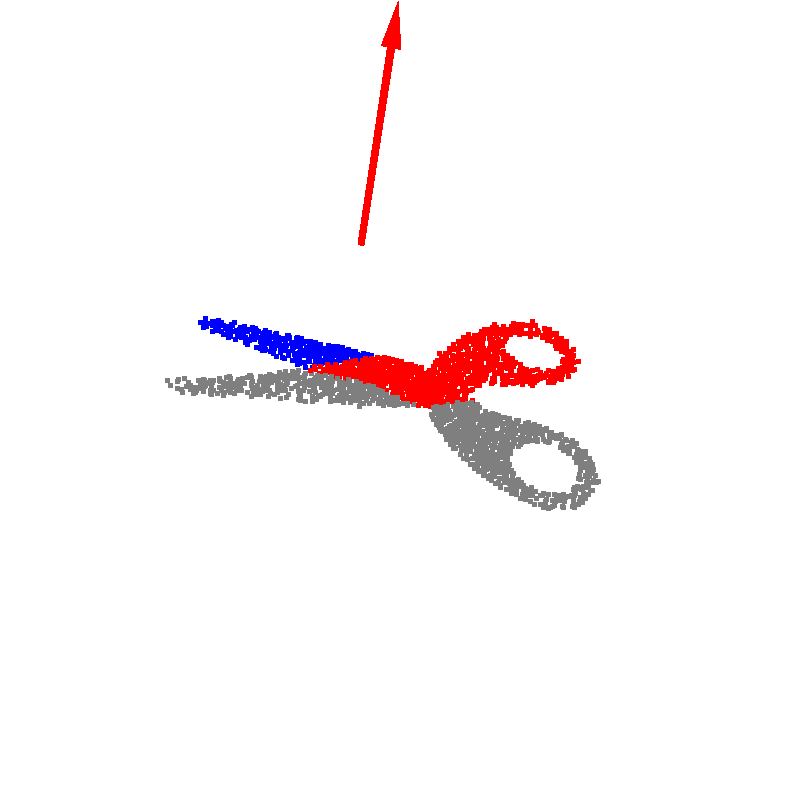} &
\includegraphics[trim={4cm 5cm 4cm 5cm},clip,width=\qualitWidth]{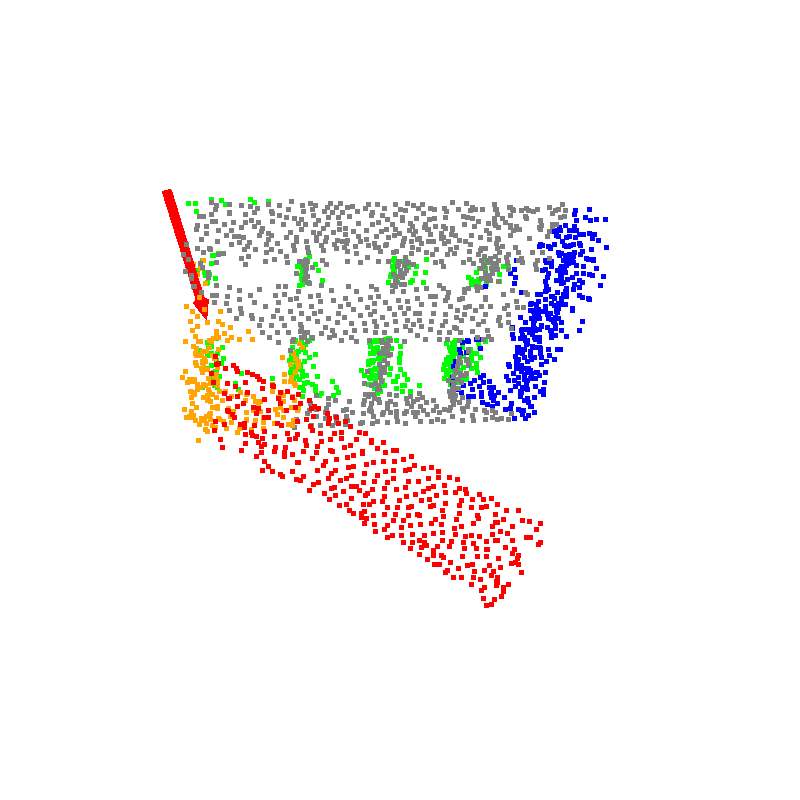}\\ [-0.4cm]

\rotatebox{90}{$\;\;\;\;$Video2} 
\rotatebox{90}{Articulation} &
\begin{minipage}{\qualitWidth}
    \centering 
    \raisebox{2cm}{(failed)} 
\end{minipage} &
\includegraphics[trim={6cm 7cm 6cm 7cm},clip,width=\qualitWidth]{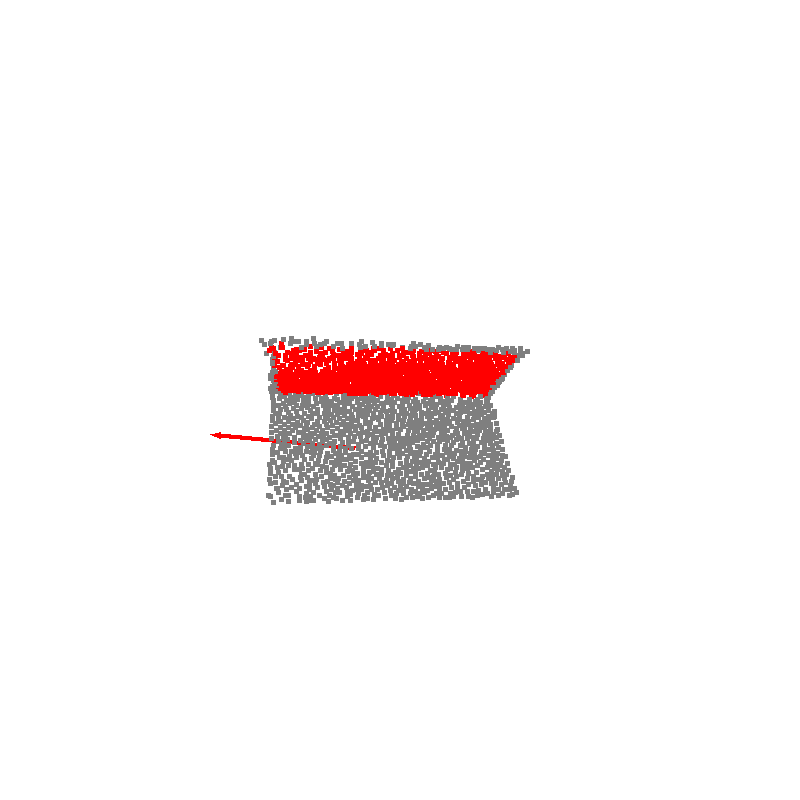} &
\includegraphics[trim={3cm 3cm 3cm 3cm},clip,width=\qualitWidth]{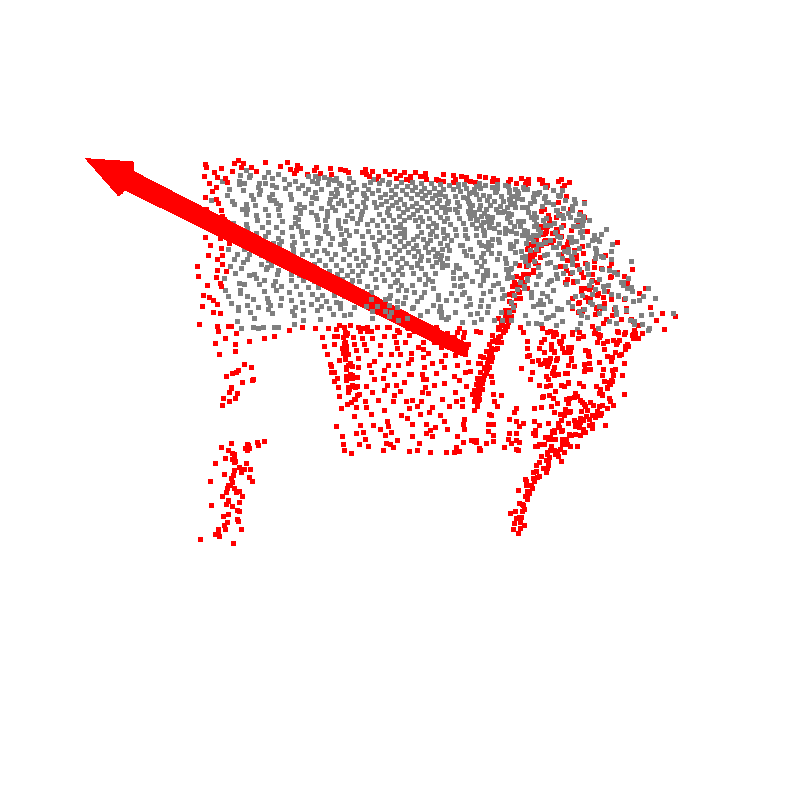} &
\includegraphics[trim={5cm 3cm 5cm 3cm},clip,width=\qualitWidth]{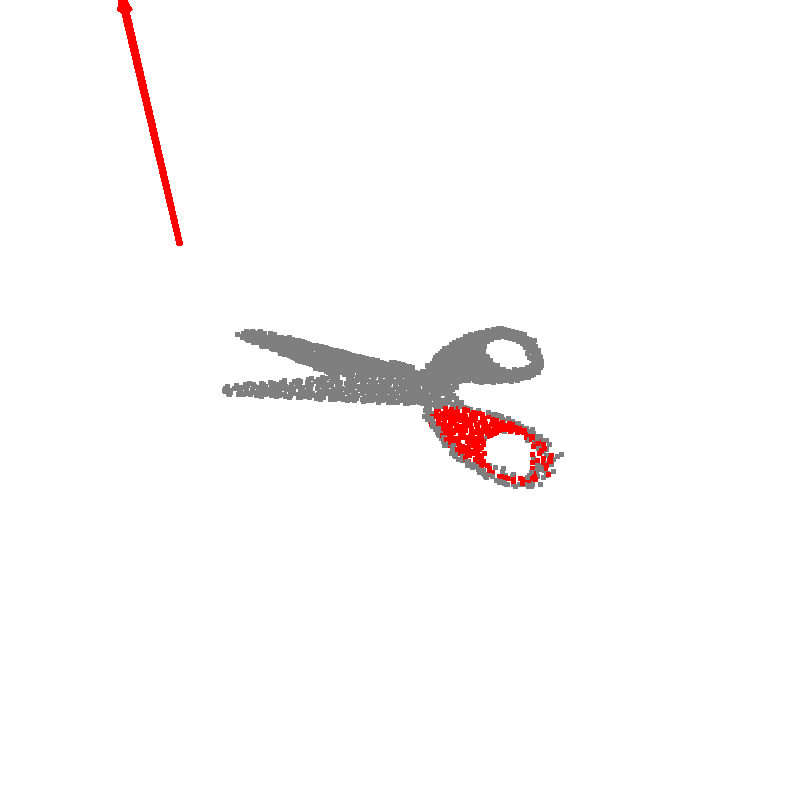} &
\includegraphics[trim={7cm 7cm 7cm 7cm},clip,width=\qualitWidth]{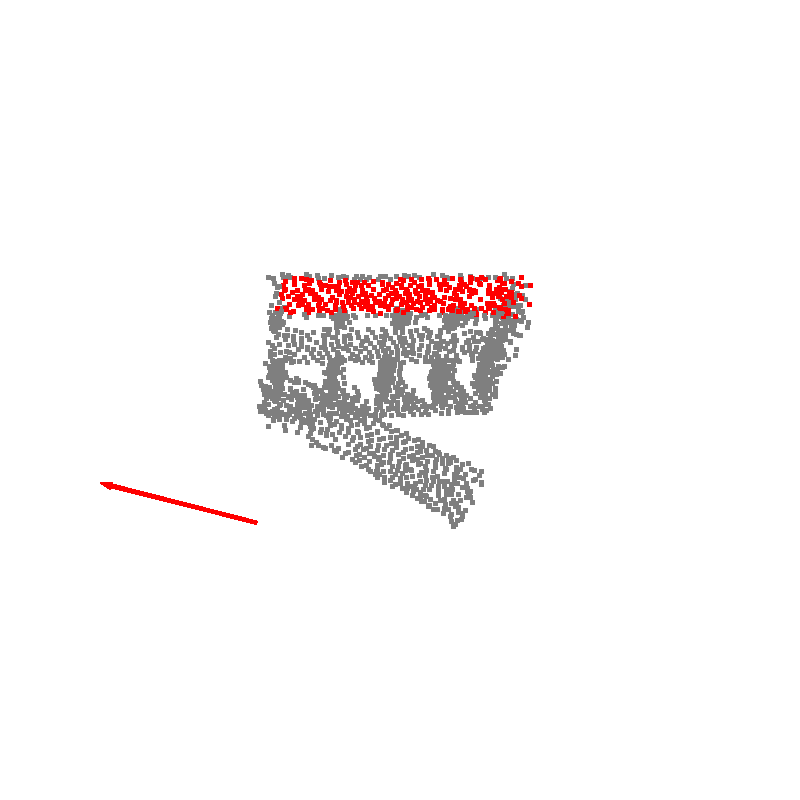}\\[-0.8cm]


\rotatebox{90}{FeatClust} &
\includegraphics[trim={5cm 7cm 5cm 7cm},clip,width=\qualitWidth]{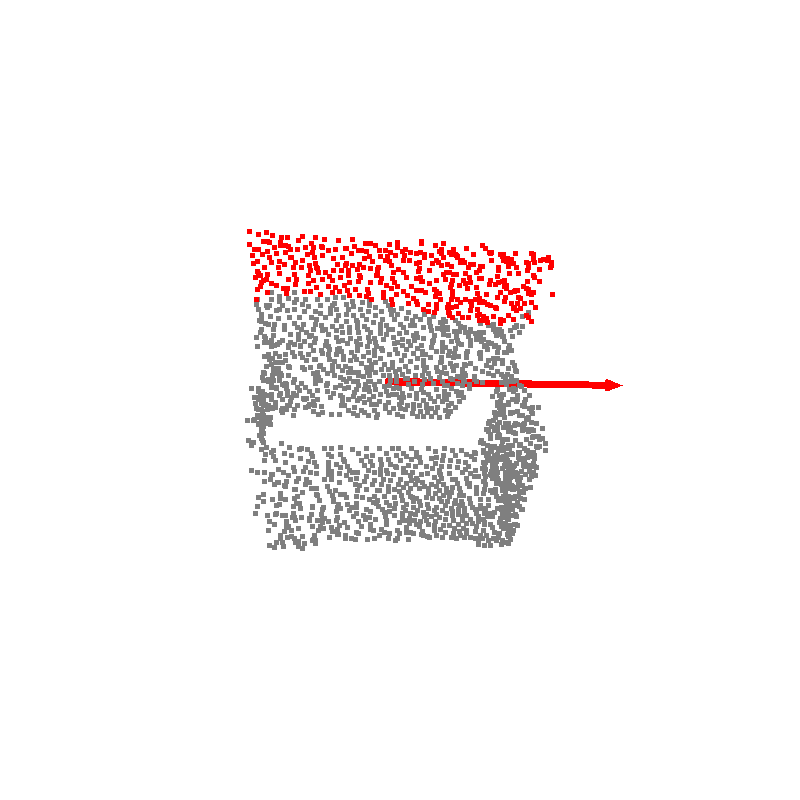} &
\includegraphics[trim={5cm 5cm 5cm 5cm},clip,width=\qualitWidth]{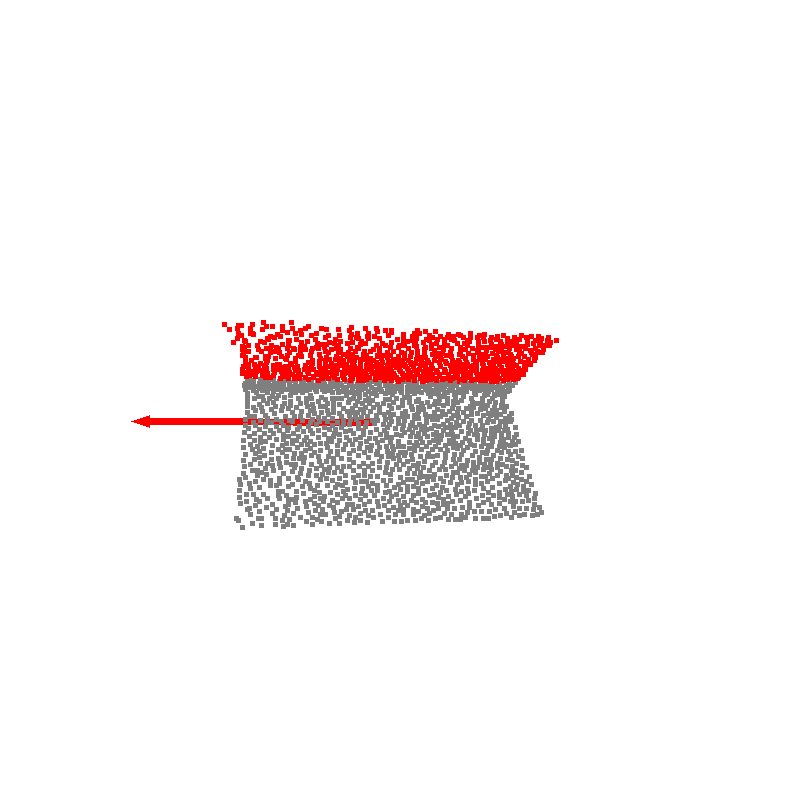} &
\includegraphics[trim={2cm 2cm 2cm 2cm},clip,width=\qualitWidth]{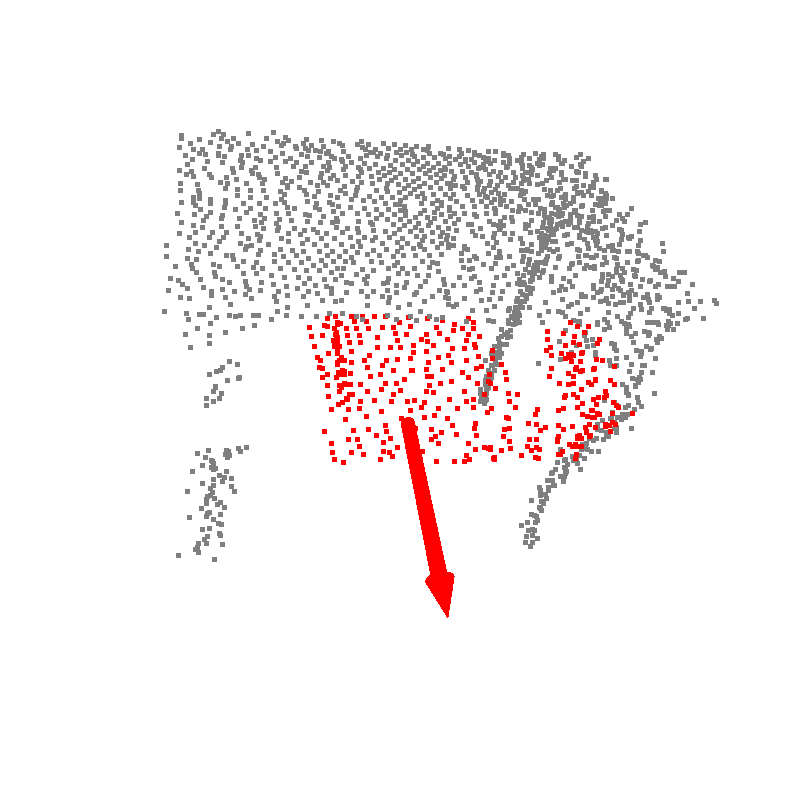} &
\includegraphics[trim={3cm 3cm 3cm 3cm},clip,width=\qualitWidth]{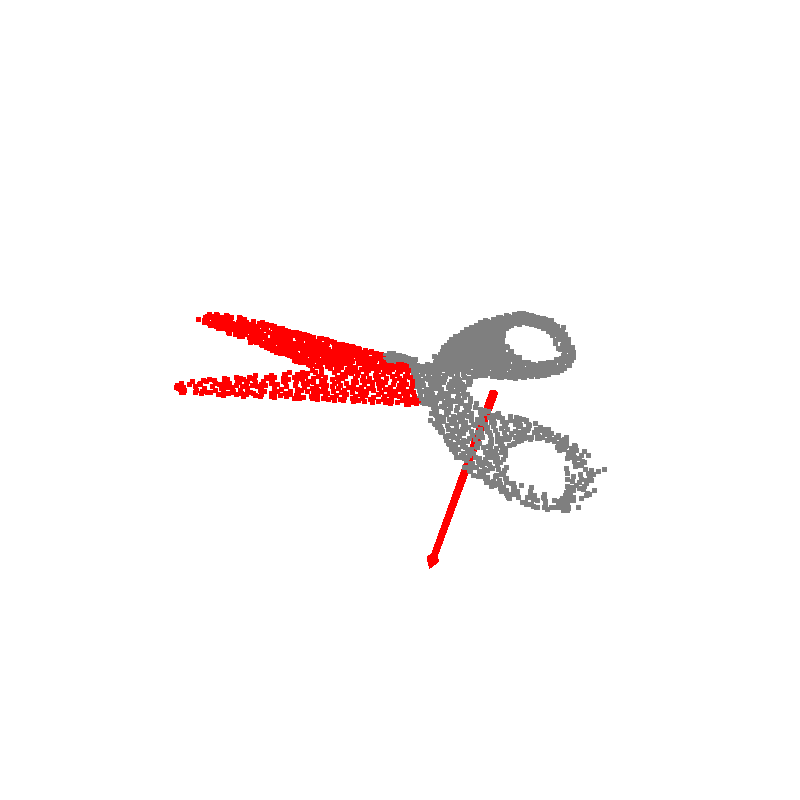} &
\includegraphics[trim={3cm 3cm 3cm 3cm},clip,width=\qualitWidth]{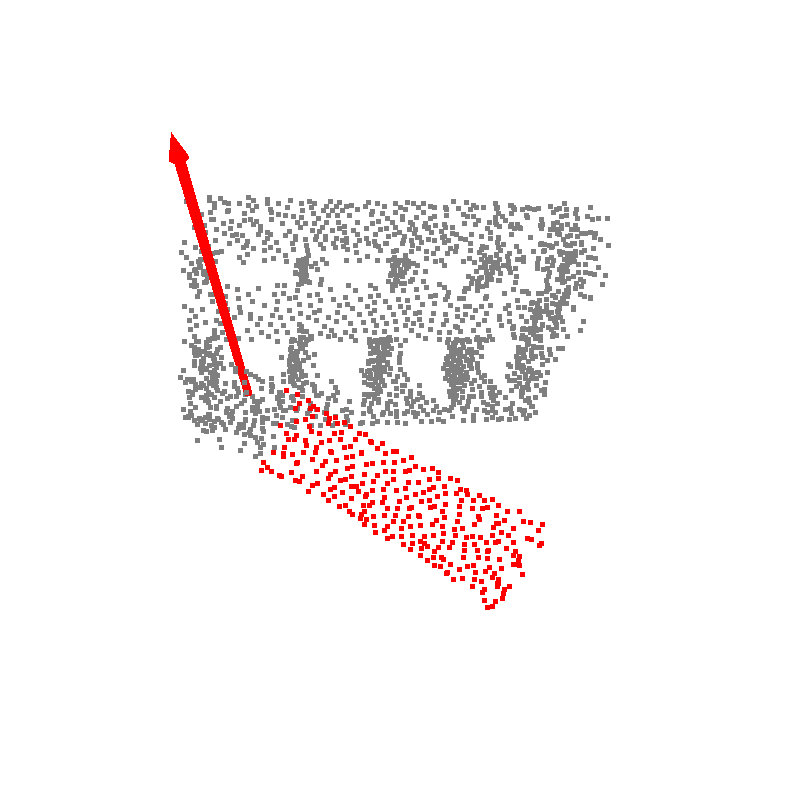}\\

&Box &Laptop &Table1 &Scissors &Storage3  \\

\end{tabular}
   \caption{\textbf{Qualitative results on the rest of the two-part samples from the 4art-synth dataset.} The red arrow represents predicted joint axis. Note how our method retrieves the part segmentations and the rotation axes much more accurately and robustly than all the other methods. \textit{Video results are provided in the video supplementary material. }}
   \label{fig:qual_fig_two_part_suppmat}
\end{figure*}


\def\qualitWidth{0.20\linewidth}

\setlength{\tabcolsep}{0pt}
\begin{figure*}
  \centering
  
\begin{tabular}{c@{$\;$}c@{$\;\;$}ccccc}

\rotatebox{90}{\hspace{0.5cm}\vphantom{A}Ground}
\rotatebox{90}{\hspace{0.7cm}\vphantom{A}Truth} &
\includegraphics[trim={5cm 5.5cm 5cm 5.5cm},clip,width=\qualitWidth]{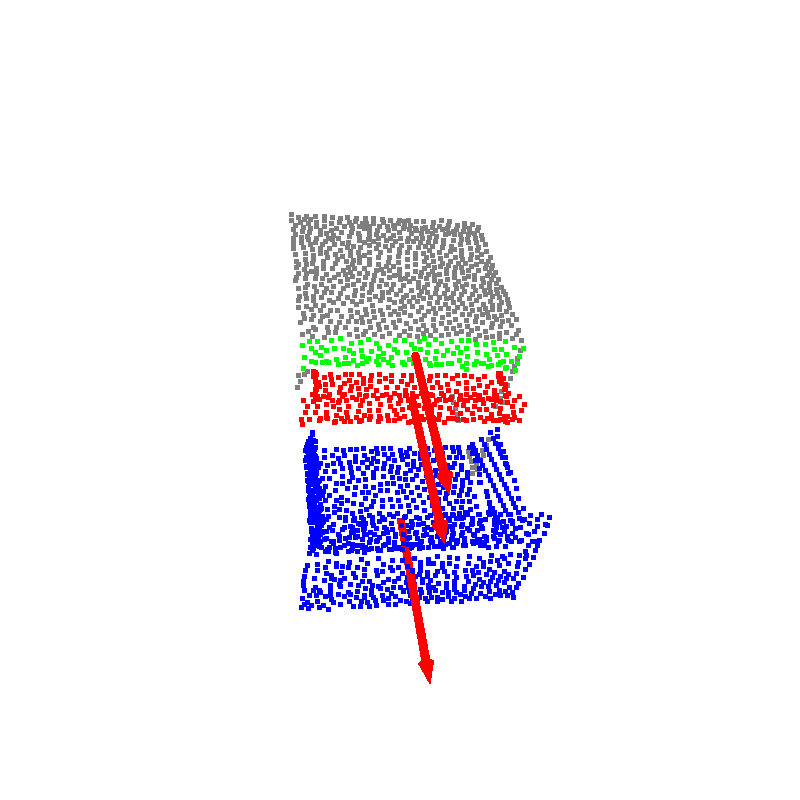} &
\includegraphics[trim={4cm 5.5cm 4cm 5.5cm},clip,width=\qualitWidth]{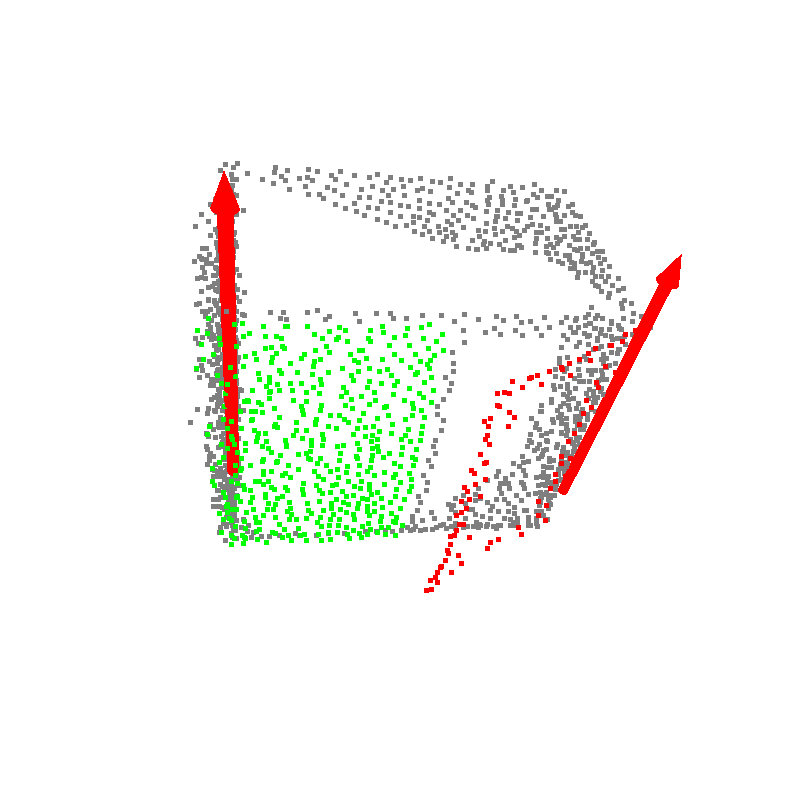} &
\includegraphics[trim={2cm 3cm 2cm 3cm},clip,width=\qualitWidth]{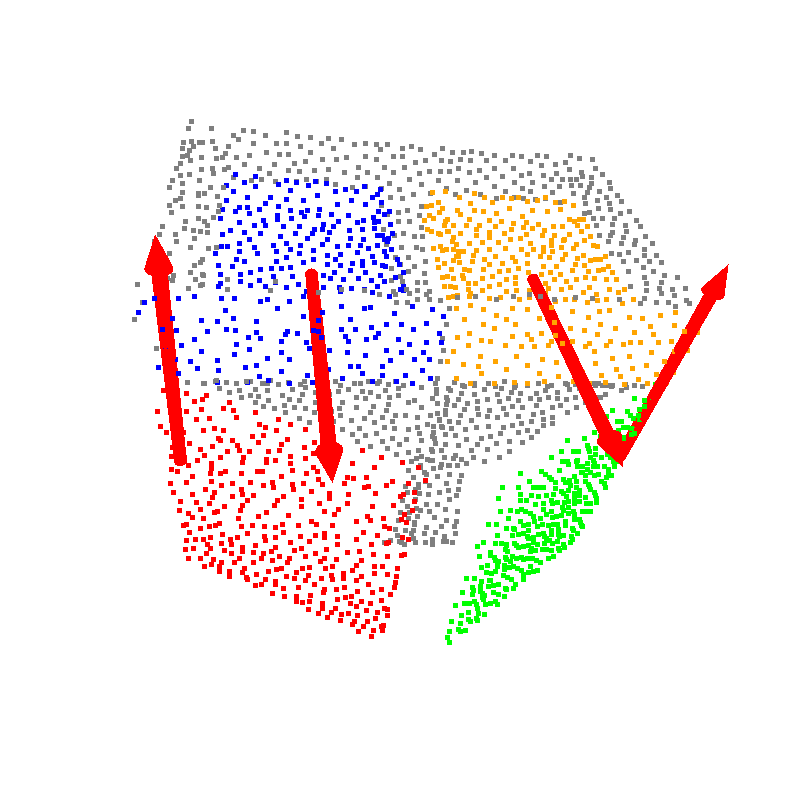} &
\includegraphics[trim={5cm 3cm 5cm 3cm},clip,width=\qualitWidth]{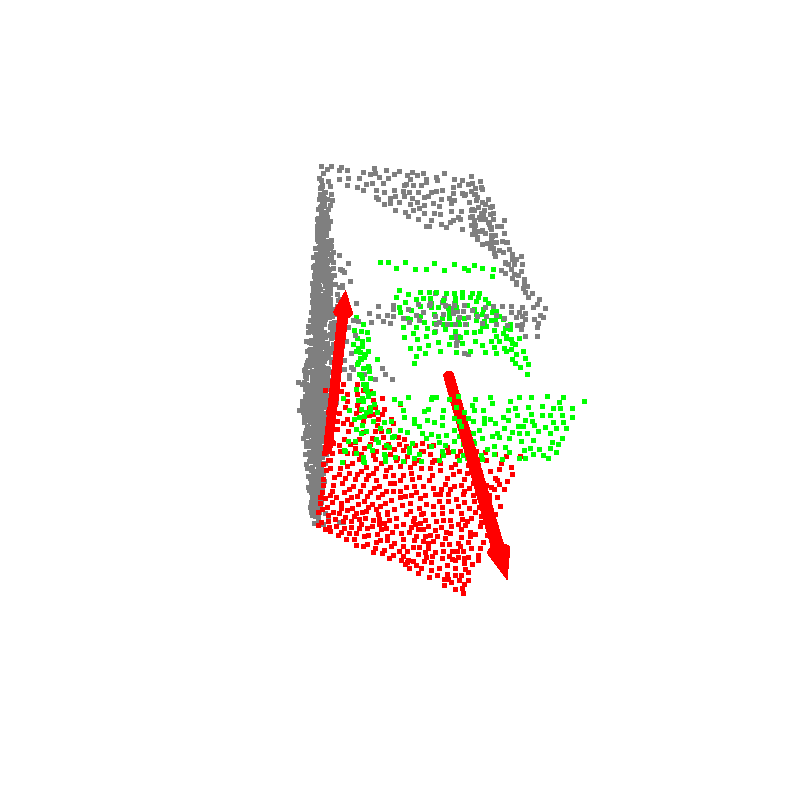} &
\includegraphics[trim={3cm 3cm 3cm 3cm},clip,width=\qualitWidth]{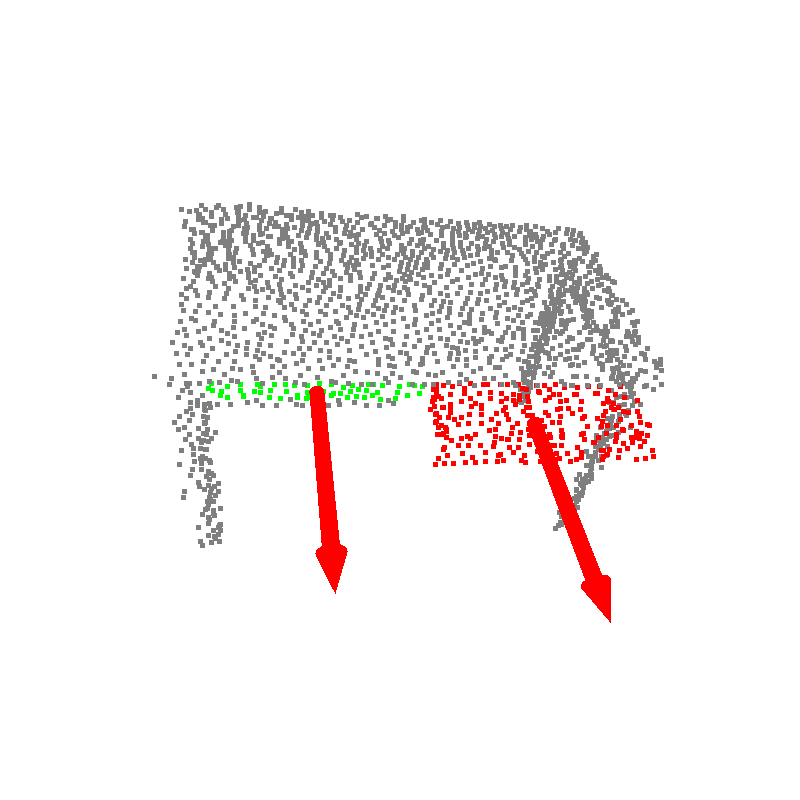} \\[-0.0cm]

\rotatebox{90}{\;\;\;\;\;sim2art} 
\rotatebox{90}{\;\;\;\;\;\;(ours)} &
\includegraphics[trim={5cm 5.5cm 5cm 5.5cm},clip,width=\qualitWidth]{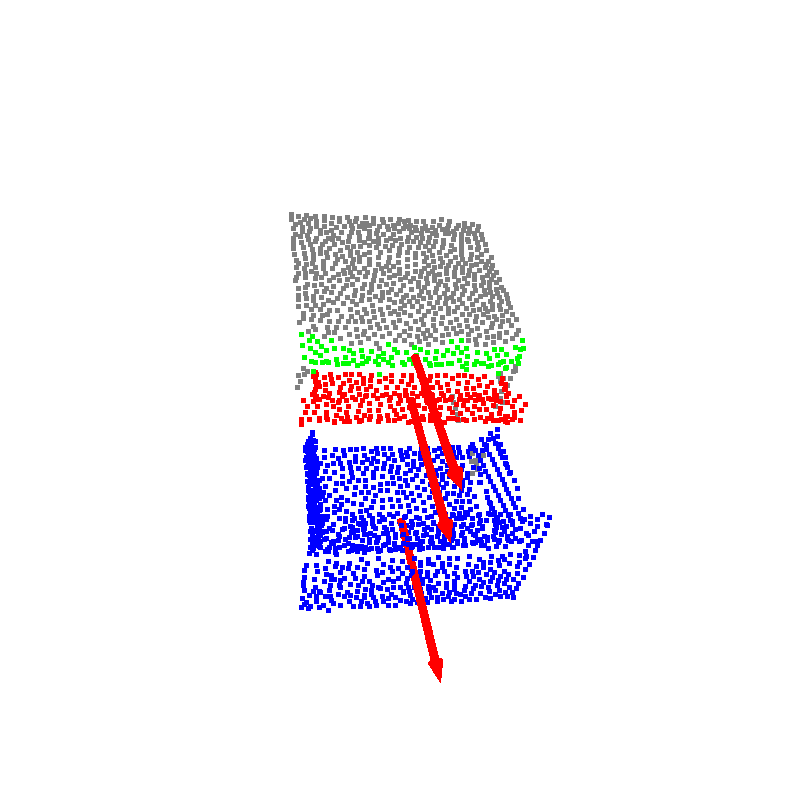} &
\includegraphics[trim={4cm 5.5cm 4cm 5.5cm},clip,width=\qualitWidth]{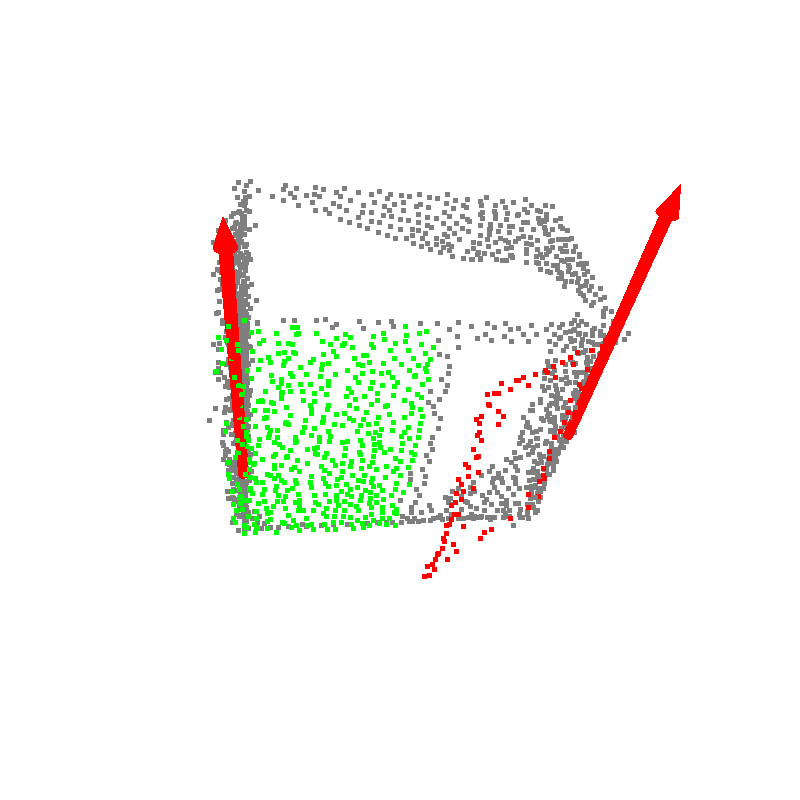} &
\includegraphics[trim={2cm 3cm 2cm 3cm},clip,width=\qualitWidth]{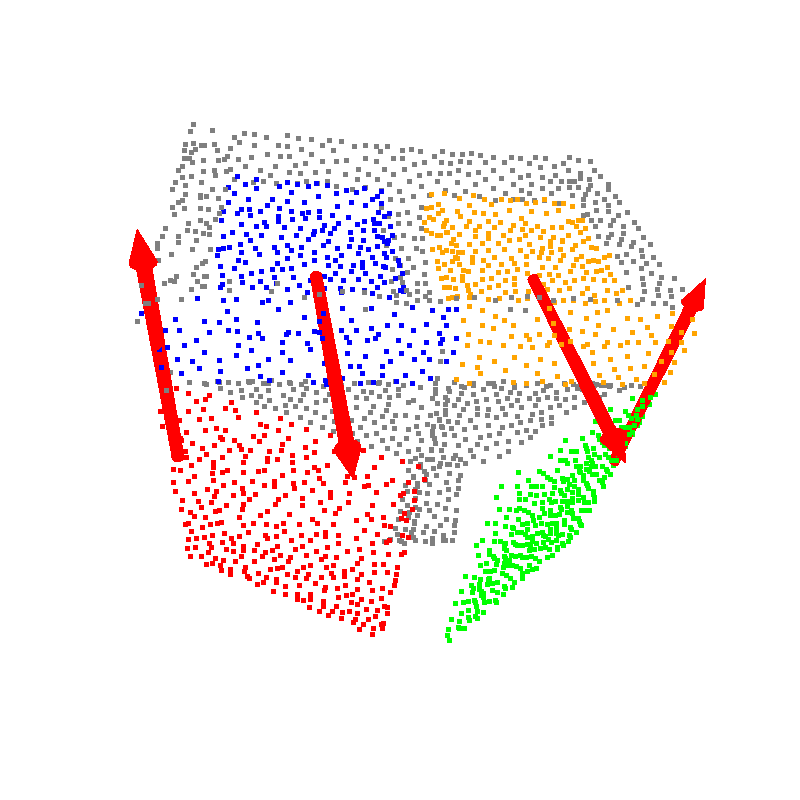} &
\includegraphics[trim={5cm 3cm 5cm 3cm},clip,width=\qualitWidth]{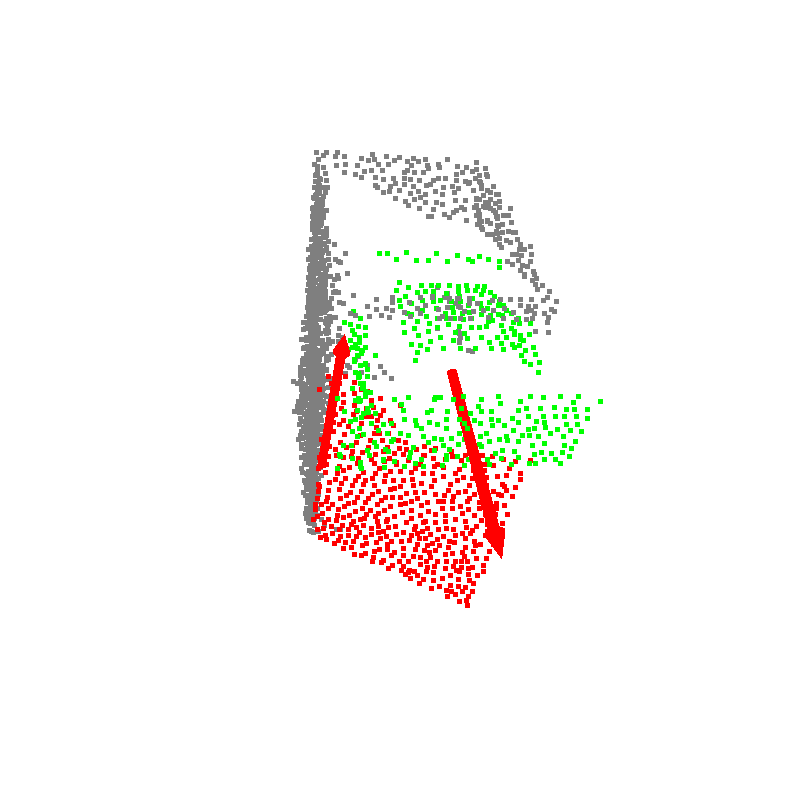} &
\includegraphics[trim={3cm 3cm 3cm 3cm},clip,width=\qualitWidth]{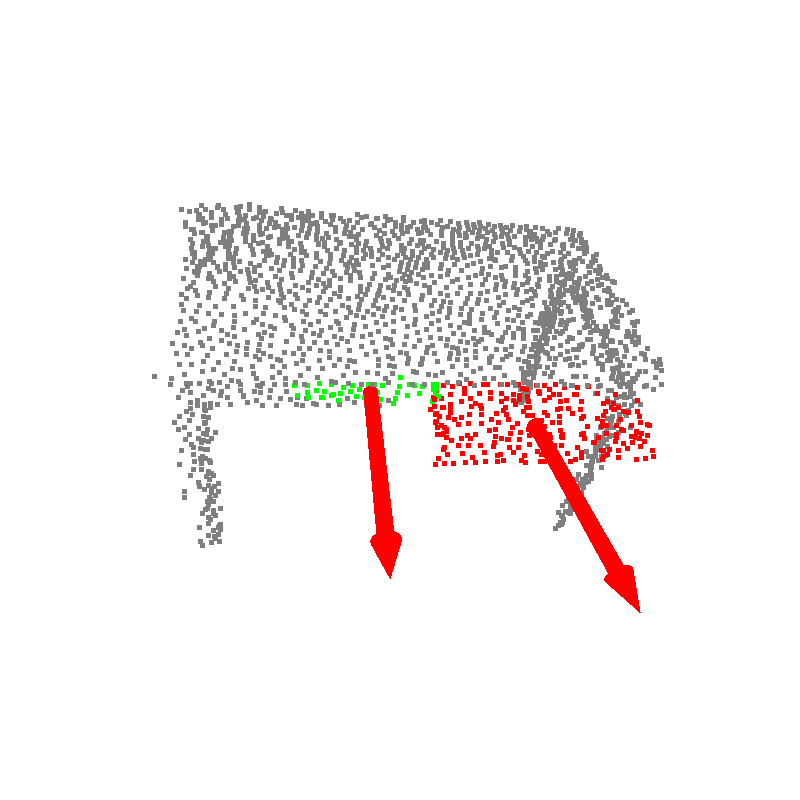}\\[-0.0cm]


\rotatebox{90}{\hspace{1.0cm}\vphantom{A}Reart} &
\includegraphics[trim={5cm 3cm 5cm 3cm},clip,width=\qualitWidth]{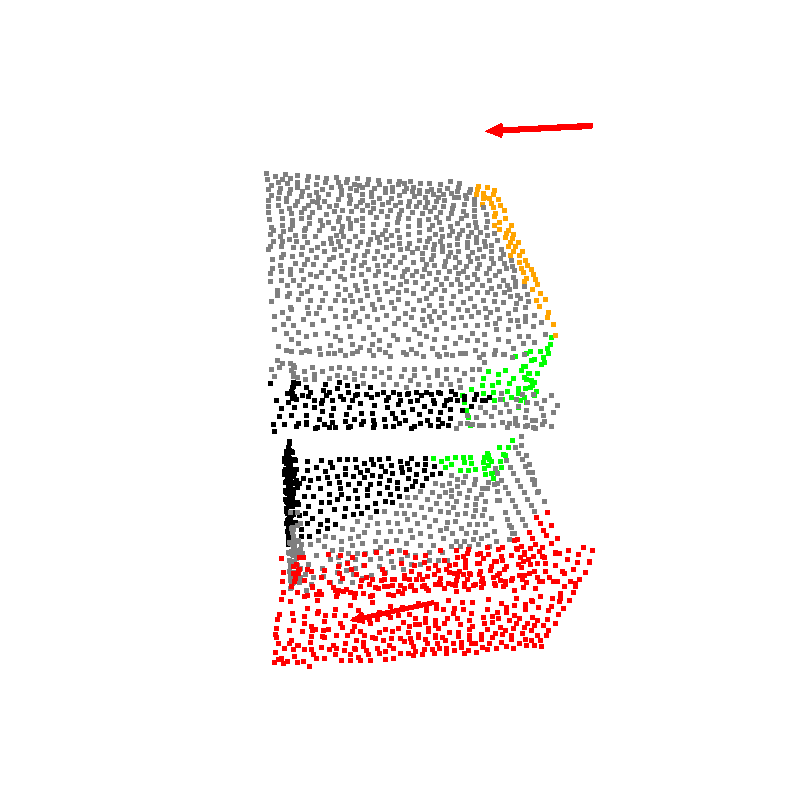} &
\includegraphics[trim={7cm 5cm 7cm 5cm},clip,width=\qualitWidth]{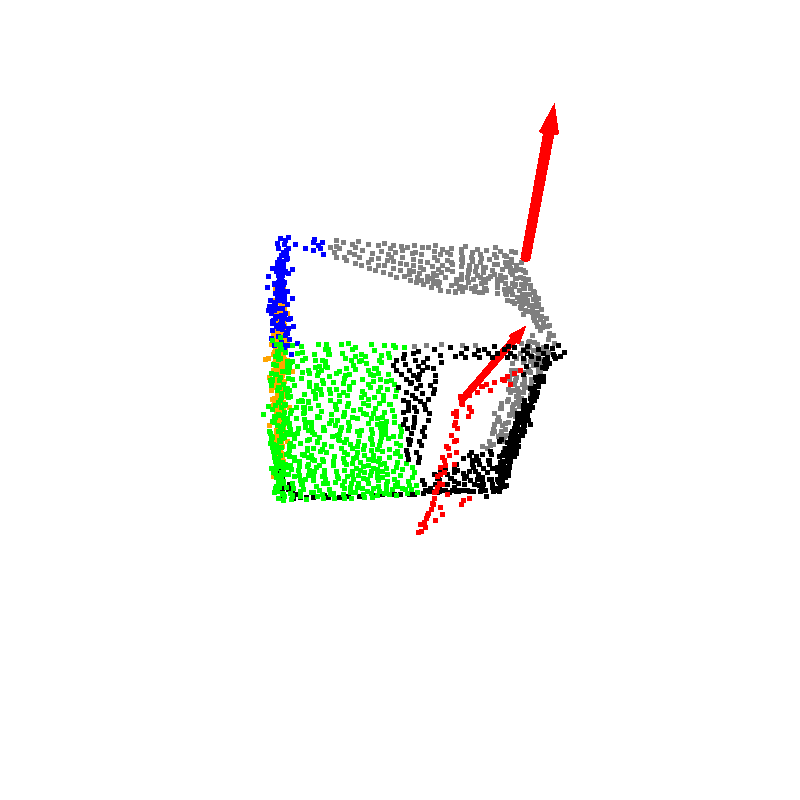} &
\includegraphics[trim={8cm 5cm 8cm 5cm},clip,width=\qualitWidth]{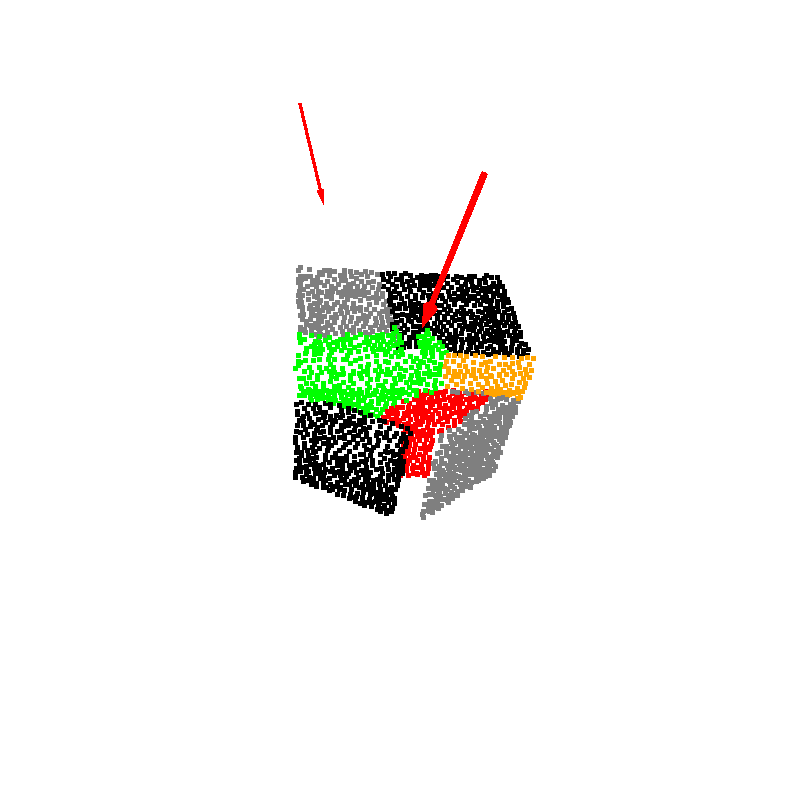} &
\includegraphics[trim={5cm 3cm 5cm 3cm},clip,width=\qualitWidth]{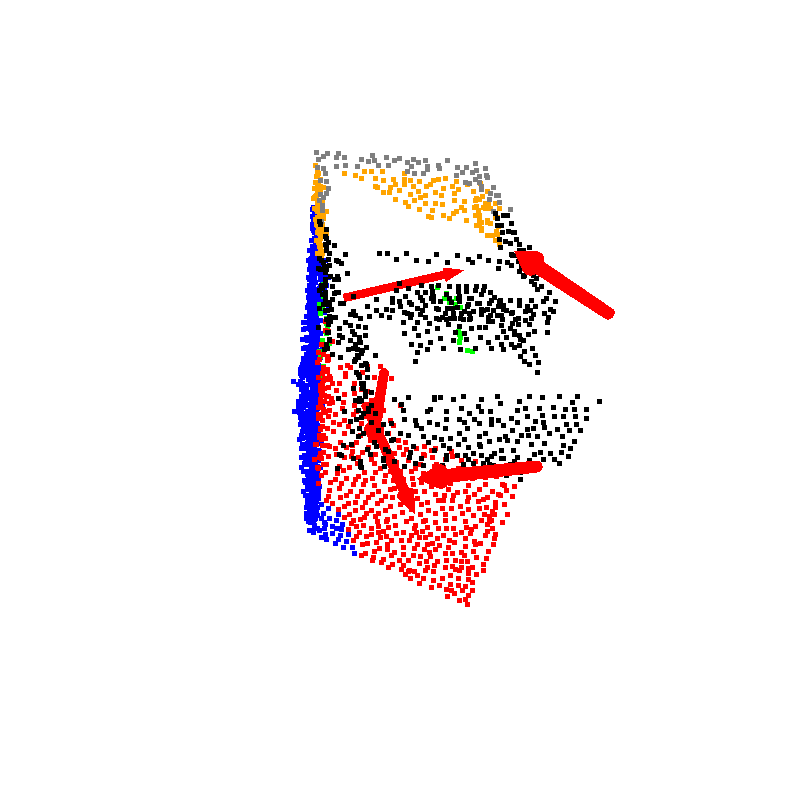} &
\includegraphics[trim={3cm 3cm 3cm 3cm},clip,width=\qualitWidth]{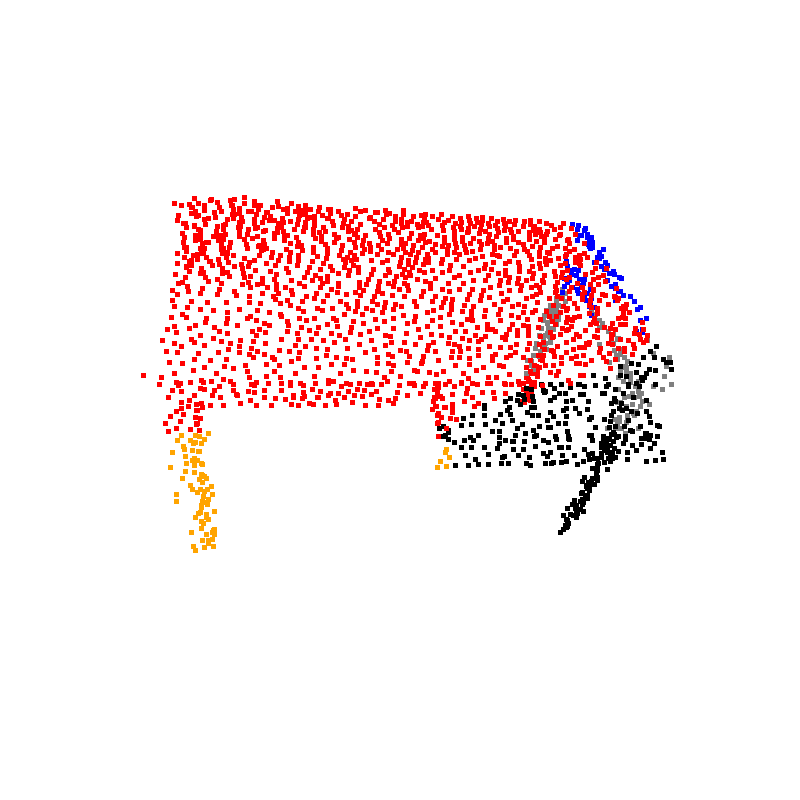}\\ [-0.0cm]


\rotatebox{90}{\hspace{0.5cm}FeatClust} &
\includegraphics[trim={7cm 7cm 7cm 7cm},clip,width=\qualitWidth]{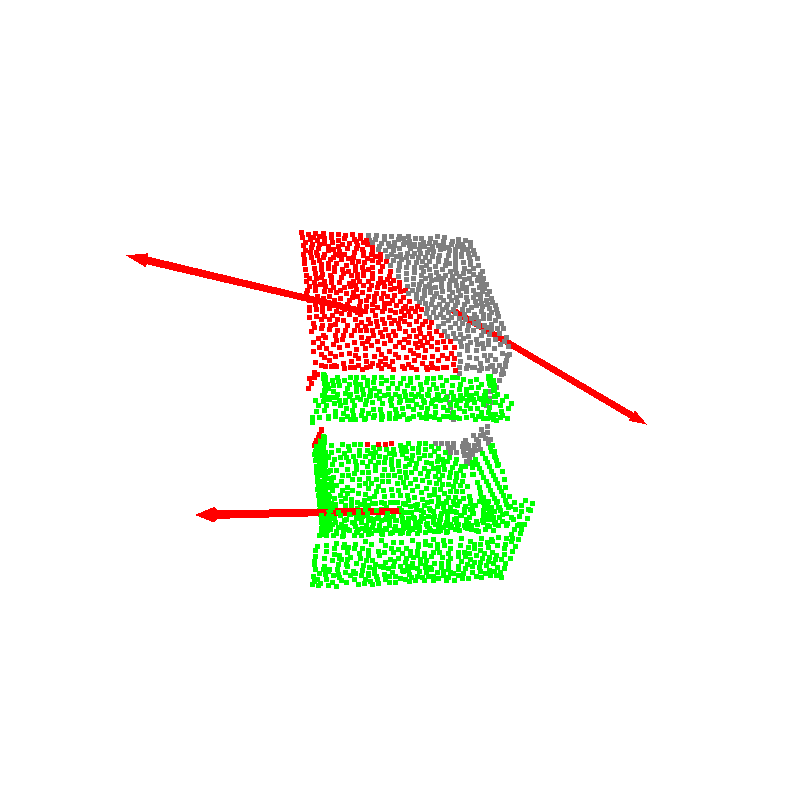} &
\includegraphics[trim={6cm 7cm 6cm 7cm},clip,width=\qualitWidth]{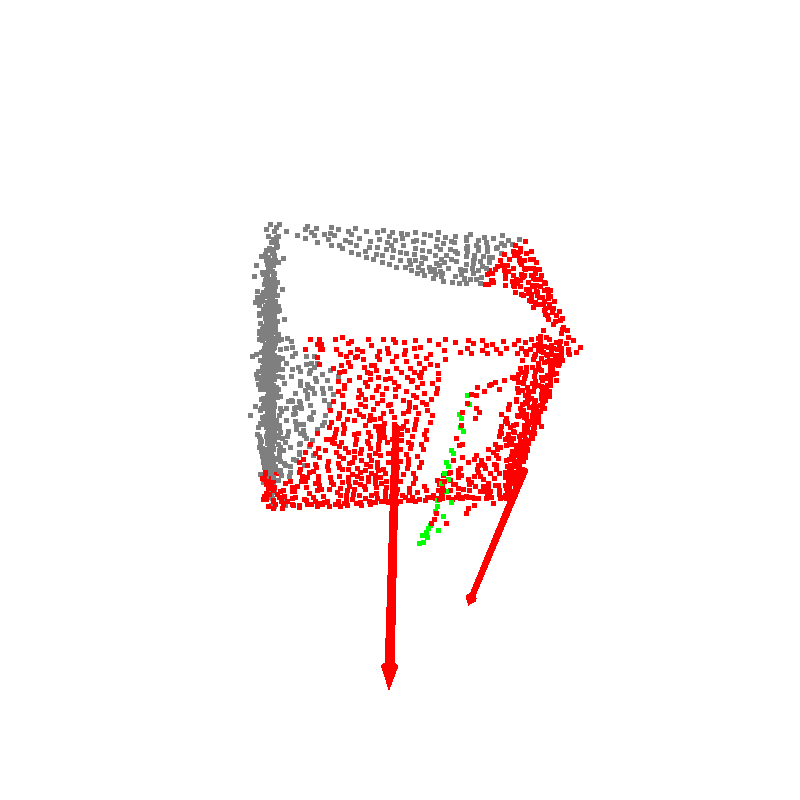} &
\includegraphics[trim={2.4cm 3cm 2.4cm 3cm},clip,width=\qualitWidth]{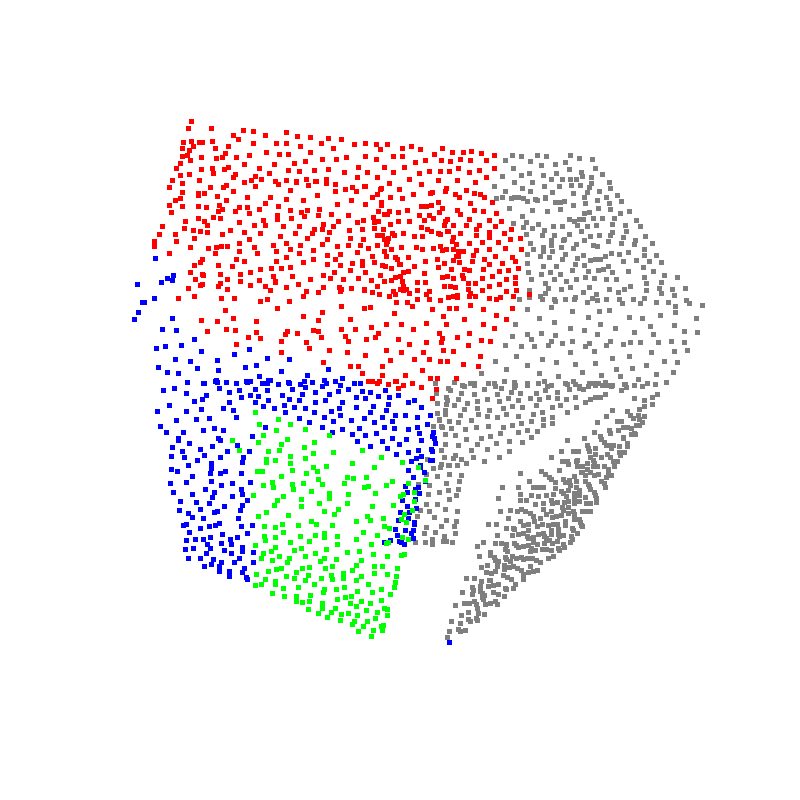} &
\includegraphics[trim={2cm 3cm 2cm 3cm},clip,width=\qualitWidth]{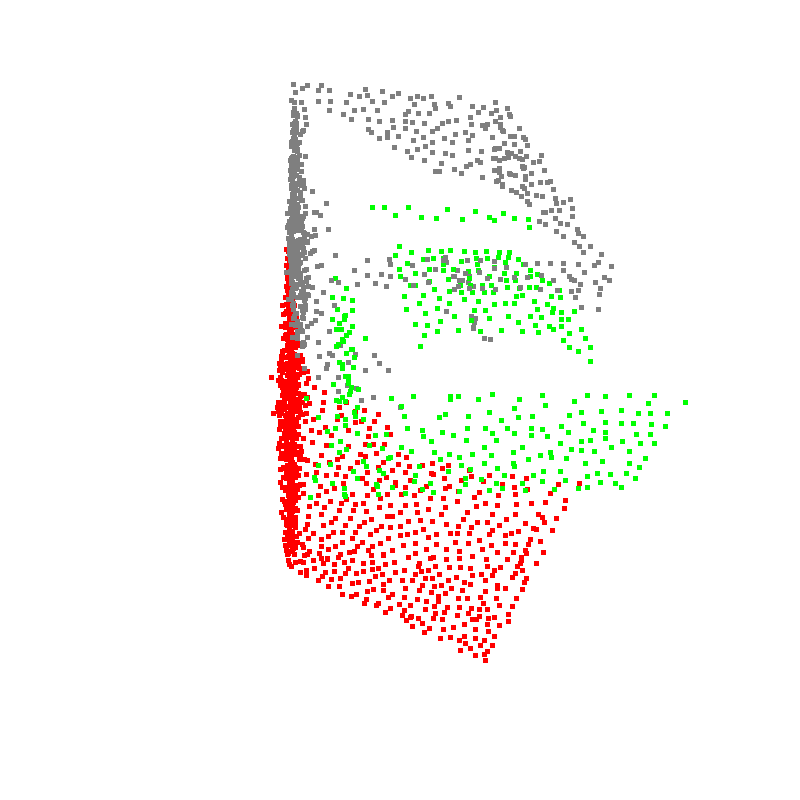} &
\includegraphics[trim={8cm 7cm 8cm 7cm},clip,width=\qualitWidth]{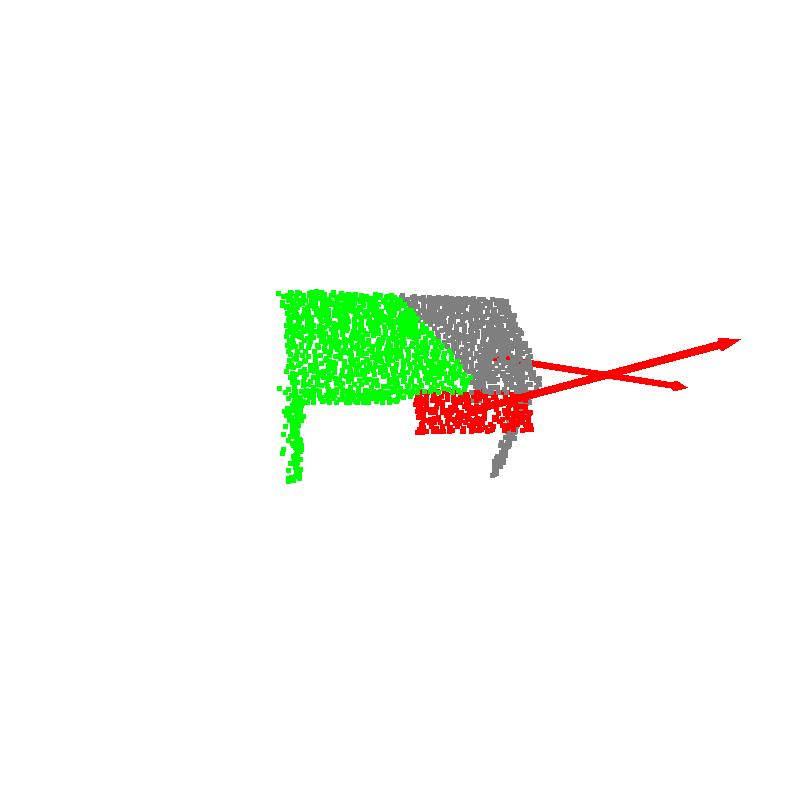}\\

&Storage1 &Storage2 &Storage5 &Storage6 &Table2\\

\end{tabular}
   \caption{\textbf{Qualitative results on the rest of the multi-part samples from the 4art-synth dataset.} The red arrow represents predicted joint axis. Note how our method retrieves the part segmentations and the rotation axes much more accurately and robustly than all the other methods. \textit{Video results are provided in the video supplementary material. }}
   \label{fig:qual_fig_multi_part_suppmat}
\end{figure*}

\paragraph{Downstream applications.} Given the part segmentation and joint parameters predicted by our method, it is possible to obtain a full digital twin of the object by training a 2D Gaussian Splatting model \cite{huang20242d} from a few observed RGB-D frames.

\begin{figure}
    \centering
    \includegraphics[width=\linewidth]{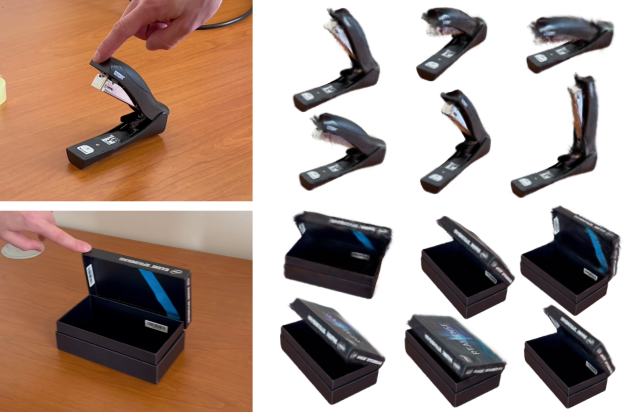}
    \caption{\textbf{2D Gaussian Splatting renderings.} We can train 2D Gaussian Splatting with the retrieved parameters from our method and obtain full digital twin of an object.}
    \label{fig:2d_gs} 
\end{figure}

More precisely, we attach one Gaussian to each point of the dynamic point cloud to ensure that all parts of the object are modeled. This results in $T$ sets of splats, each containing $N_p$ splats. The color of a given splat is initially set to the color of the corresponding pixel in the image, and its opacity to a small fixed value. For the orientation, we use the point normal obtained by taking the cross product of local tangent vectors estimated by finite differences in the depth map. The initial splat scale is isotropic and set to the average distance to its 3 nearest neighbors. Then, to represent the object in a specific joint configuration, we merge these $T$ sets into a single set containing $T \times N_p$ splats. To achieve this, we transform points using the predicted joints and segmentation: each splat corresponding to an articulated part is displaced according to the relative motion of its joint between the current and target configurations. We refine splat colors, opacities, and scales, as well as joint amounts, via gradient descent. We employ an $L_1$ RBG loss on the foreground, defined by the object masks, and an $L_1$ alpha loss on the background to suppress floating splats.
Figure~\ref{fig:2d_gs} shows renderings of the optimized 2D Gaussians from multiple camera viewpoints and various joint configurations.

\end{document}